\documentclass{article} % For LaTeX2e
\usepackage{iclr2024_conference,times}
\usepackage[utf8]{inputenc} % allow utf-8 input
\usepackage[T1]{fontenc}    % use 8-bit T1 fonts
\usepackage{hyperref}       % hyperlinks
\usepackage{url}            % simple URL typesetting
\usepackage{booktabs}       % professional-quality tables
\usepackage{amsfonts}       % blackboard math symbols
\usepackage{nicefrac}       % compact symbols for 1/2, etc.
\usepackage{microtype}      % microtypography
\usepackage{authblk}        % for author affiliation mark

\usepackage{xcolor}         % colors
\usepackage{amsmath}        % for align of equations
\usepackage{amssymb}        % for triangleq and other math symbols
\usepackage{algorithm}      % algorithms
\usepackage{algorithmic}    % algorithms
\usepackage{subfigure}
\usepackage{wrapfig}
\usepackage{multirow}
\usepackage{bm}
\usepackage{authblk}

\usepackage[capitalize]{cleveref}
\usepackage{hyperref}
\crefname{section}{Sec.}{Secs.}
\Crefname{section}{Section}{Sections}
\Crefname{table}{Table}{Tables}
\crefname{table}{Tab.}{Tabs.}

\usepackage[figuresright]{rotating}
\usepackage{makecell}

\usepackage[greek,english]{babel}
\usepackage{teubner}

\iclrfinalcopy % Uncomment for camera-ready version, but NOT for submission.

\title{VFLAIR: A Research Library and Benchmark for Vertical Federated Learning}

\author[1]{\textbf{Tianyuan Zou}}
\author[2]{\textbf{Zixuan Gu}}
\author[3]{\textbf{Yu He}}
\author[4]{\textbf{Hideaki Takahashi}}
\author[ \ 1,5]{\textbf{Yang Liu\thanks{Corresponds to Yang Liu (liuy03@air.tsinghua.edu.cn).}}} %
\author[1]{\textbf{Ya-Qin Zhang}} 

\affil[1]{\footnotesize Institute for AI Industry Research, Tsinghua University, Beijing, China}
\affil[2]{\footnotesize Weiyang College, Tsinghua University, Beijing, China}
\affil[3]{\footnotesize School of Computer Science, Fudan University, Shanghai, China}
\affil[4]{\footnotesize College of Arts and Sciences, The University of Tokyo, Tokyo, Japan}
\affil[5]{\footnotesize Shanghai Artificial Intelligence Laboratory, China}

\begin{document}

\maketitle

\begin{abstract}
  % {\color{red}{
  % Vertical Federated Learning (VFL) has emerged as a collaborative training paradigm that allows participants with different features of the same group of users to accomplish cooperative training without exposing their raw data or model parameters. VFL has gained significant attention for its research potential and real-world applications in recent years. Open-sourced industrial frameworks that support VFL like FATE have been widely deployed but are too heavy for researchers. Unfortunately, research oriented light-weight VFL frameworks is still absent. To facilitate the researches of VFL by making the implementation of new algorithms easier and faster, we introduce an extensible VFL framework \verb+VFLAIR+ (available at \url{https://github.com/FLAIR-THU/VFLAIR}) with standardized modules that provides basic VFL training flow supporting both neural network (NN) based VFL and tree-based VFL. $10$ datasets belonging to multiple modals are supported. $2$ kinds of model partition (aggVFL and splitVFL) and $2$ kinds of communication protocols (FedSGD and FedBCD) are also integrated, along with $11$ attack methods and $8$ defense methods considering the privacy issue of VFL. We also benchmark multiple attacks and defenses performance under different communication and model partition settings. We hope that \verb|VFLARI| can facilitate future VFL researches by providing a robust foundation of code and performance results of existing methods.}}
  Vertical Federated Learning (VFL) has emerged as a collaborative training paradigm that allows participants with different features of the same group of users to accomplish cooperative training without exposing their raw data or model parameters. VFL has gained significant attention for its research potential and real-world applications in recent years, but still faces substantial challenges, such as in defending various kinds of data inference and backdoor attacks. Moreover, most of existing VFL projects are industry-facing and not easily used for keeping track of the current research progress. To address this need, we present an extensible and lightweight VFL framework \verb+VFLAIR+ (available at 
  \url{https://github.com/FLAIR-THU/VFLAIR}), which supports VFL training with a variety of models, datasets and protocols, along with standardized modules for comprehensive evaluations of attacks and defense strategies. We also benchmark $11$ attacks and $8$ defenses performance under different communication and model partition settings and draw concrete insights and recommendations on the choice of defense strategies for different practical VFL deployment scenarios. 
  % We hope that \verb|VFLARI| can facilitate future VFL researches by providing a robust foundation of code and performance results of existing methods.
  
  %introduce a novel and unified privacy-utility trade-off evaluation metrics to asses the defense ability of a given defending method against different kinds of attacks. We then 
  % Furthermore, we perform extensive experiments to benchmark the ability of 8 existing defenses methods against a wide variety of attacks, including 6 types of label inference attacks, 2 types of feature reconstruction attacks and 3 types of backdoor attacks. %Each defense is evaluated with at least 4 defense rates covering a wide spectrum of defense ability. 
  % Each evaluation is also performed under multiple model partition (aggVFL and splitVFL) and communication protocols (FedSGD and FedBCD) with multiple datasets. Based on our comprehensive evaluations, we draw concrete insights and recommendations on the choice of defense strategies for different practical VFL deployment scenarios.% We hope that our benchmarks and framework could serve as a valuable guidance for future privacy-preserving VFL applications and facilitate the development of more robust and privacy-preserving VFL systems.
  %\tianyuan{@zixuan} \yang{code and documents need to be updated.}

\end{abstract}

\section{Introduction}

The concept of Federated Learning (FL) was first introduced by Google in 2016~\citep{McMahan2016fl} describing a cross-device scenario where millions of mobile users collaboratively train a shared model using their local private data
% with the help of a central server 
without centralizing these data. This scenario is regarded as Horizontal FL (HFL)~\citep{yang2019federatedbook} as data are partitioned by sample. In another type of FL, regarded as Vertical FL (VFL)~\citep{yang2019federatedbook}, data are partitioned by feature. VFL is often applied in industrial collaborative learning scenarios where each organization controls disjoint features of a common group of users. In VFL, local data and local model are kept private at each participant. Instead, local model outputs and their gradients are transmitted between parties.
% The industrial demand for VFL has grown rapidly in recent years with several real-word applications already exists~\citep{cai2020Bytedance,tencent2021}.%\yang{add reference, give at least one industry example} \tianyuan{DONE}\

VFL has drawn increasing attention from both academic and industry in recent years with hundreds of research papers published every year and a number of open-sourced projects released, including FATE~\citep{FATE,liu2021fate}, Fedlearner~\citep{Fedlearner}, PaddleFL~\citep{PaddleFL}, Pysyft~\citep{ryffel2018generic,romanini2021pyvertical}, FedTree~\citep{fedtree2022}, and FedML~\citep{he2020fedml}. Real-world industrial cases are also emerged in the field of advertising~\citep{cai2020Bytedance,tencent2021} and finance~\citep{cheng2020federated,Cheng2022DigitalEra} etc. However, mainstream VFL projects such as FATE are industrial grade and not designed for keeping up with research advances.  %, to name a few.

Meanwhile, research interests for VFL have been growing rapidly over the past years, focusing on improving various aspects of VFL protocols, such as communication efficiency~\citep{fu2021vf2boost,liu2022fedbcd,castiglia2022compressed,fu2022towards}, robustness to attacks~\citep{liu2021rvfr,cheng2021secureboost,li2022label,zou2022defending,zou2023mutual,sun2022label,yang2022differentially}, model utility~\citep{li2022semi,Yitao2022multiview,feng2020multi,feng2022semisupervised}, and fair incentive designs~\citep{liu2021achieving,qi2022fairvfl}. % are two main lines of reserach work considering VFL. 
For communication efficiency, methods like decrease communication rounds using multiple local updates between each round~\citep{liu2022fedbcd,fu2022towards} or compress information~\citep{castiglia2022compressed} have been proposed. As for data security and privacy, various attacks injected by one or multiple parties aiming to either steal other parties' private label~\citep{li2022label,fu2021label,zou2022defending}, private features~\citep{jin2021cafe,luo2021feature,li2022ressfl,jiang2022comprehensive,ye2022feature}, sensitive attributes~\citep{Song2020Overlearning} and sample relations~\citep{qiu2022your}, or negatively impact the model behavior~\citep{liu2021rvfr,zou2022defending} have been put forward. Multiple defending methods have also been proposed to tackle these threats, including adding noise~\citep{dwork2006DP,zou2022defending,li2022label}, sparsifying gradients~\citep{aji2017sparse,fu2021label,zou2022defending}, discreting gradients~\citep{fu2021label}, label differential privacy~\citep{ghazi2021deep,yang2022differentially}, adding distance correlation regularizor~\citep{sun2022label,vepakomma2019reducing}, disguising labels~\citep{zou2022defending}, adding mutual information regularizer~\citep{zou2023mutual}, adversarial training~\citep{sun2021defending,li2022ressfl} or performing robust feature recovery~\citep{liu2021rvfr}. However, each of these defenses are evaluated under specific tasks and settings, lacking of key insights and metrics on evaluating these defense strategies to defend all possible attacks in practical deployment.

To facilitate future research for VFL, we introduce a lightweight and comprehensive VFL framework, namely \verb|VFLAIR|, which includes not only basic VFL training and inference for a variety of models and settings but also efficiency enhancement techniques and multiple defense methods that mitigate potential threats. Moreover, we perform extensive experiments on combinations of the above settings using multiple datasets to provide different perspectives on VFL efficiency and safety. We believe \verb|VFLAIR| and these benchmark results will provide researchers with useful tools and guidance for their future work. Our contributions are summarized in the following:
% \begin{itemize}
%     \item We design \verb+VFLAIR+, a lightweight and extensible VFL framework that aims to facilitate research development of VFL (see \cref{fig:VFLAIR}). We design standardized pipelines for VFL training and validation, supporting $13$ datasets, $29$ different local model architectures including linear regression, tree and neural networks, $6$ different global models, $2$ model partition settings, $2$ communication protocols, $11$ attacks and $8$ defense methods, each implemented as a distinct module and can be easily extended. 
%     \item Extensive experiments are conducted to produce benchmark results. Performance of VFL models trained with standard VFL training, attack performance of $11$ attacks, defense capability of $8$ defense methods are evaluated under both aggVFL and splitVFL with communication protocol using both FedSGD and FedBCD.
% \end{itemize}
% \begin{itemize}

\begin{figure}[t]
\vspace{-1em}
    \centering
    \includegraphics[width=0.89\linewidth]{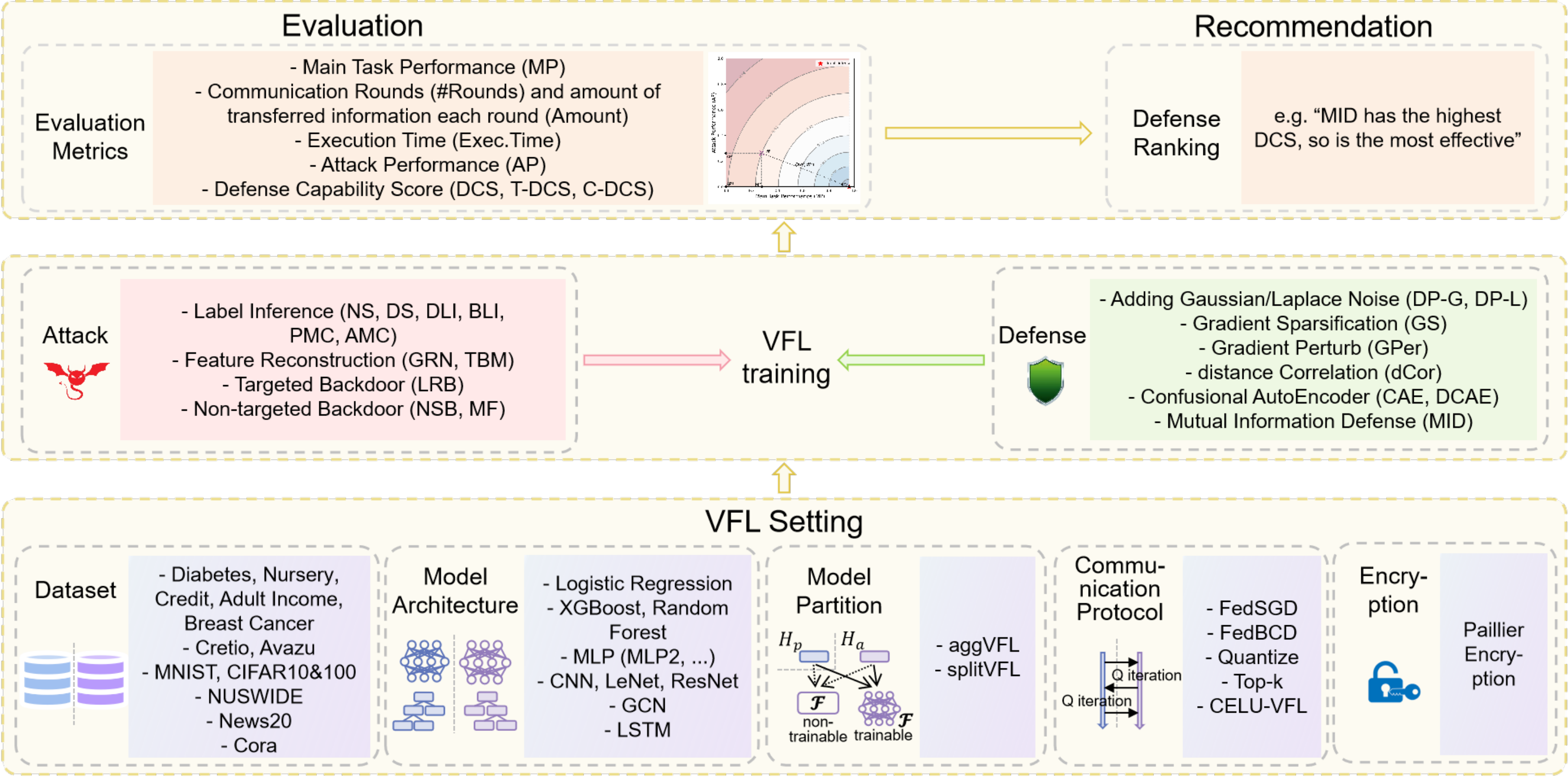}
    \caption{{An overview of the Components of VFLAIR.}} % After specifying the settings of VFL system, attacks and defense methods can be evaluated on top of the system. Overall evaluation metrics can be further exploit to get recommendation for selecting a proper defense.
\label{fig:VFLAIR}
\vspace{-1em}
\end{figure}

\textbf{(1).} We design \verb+VFLAIR+, a lightweight and extensible VFL framework that aims to facilitate research development of VFL (see \cref{fig:VFLAIR}). We design standardized pipelines for VFL training and validation, supporting $13$ datasets, $29$ different local model architectures including linear regression, tree and neural networks, $6$ different global models, $2$ model partition settings, $5$ communication protocols, $1$ encryption method, $11$ attacks and $8$ defense methods, each implemented as a distinct module and can be easily extended.

\textbf{(2).} We propose new evaluation metrics and modules, and perform extensive experiments to benchmark various perspectives of VFL, from which we draw key insights on VFL system design choice, in order to promote future development and practical deployment of VFL. % Performance of VFL models trained with standard VFL training, attack performance of $11$ attacks, defense capability of $8$ defense methods are evaluated under both aggVFL and splitVFL with communication protocol using both FedSGD and FedBCD.
% \end{itemize}

\section{Related Work}

A number of open-source FL projects have been developed supporting VFL. FATE~\citep{FATE,liu2021fate} is an industry-grade FL project which supports a variety of model architectures and secure computation protocols; Fedlearner~\citep{Fedlearner} is specialized in advertising scenarios; PaddleFL~\citep{PaddleFL} supports 2-party and 3-party VFL with MPC protection; Pysyft~\citep{ryffel2018generic,romanini2021pyvertical} introduces PyVertical, which focus on SplitNN-type of VFL settings; FedTree~\citep{fedtree2022} focuses on tree-based VFL only; FedML~\citep{he2020fedml} supports basic training of VFL with logistic regression models. Real-world industrial applications have been witnessed in domains such as advertising~\citep{cai2020Bytedance,tencent2021} and finance~\citep{cheng2020federated,Cheng2022DigitalEra}. These works demonstrate the widespread interest and the practical significance of VFL. 
However, these works are often relatively heavy-weight as they are designed for industrial deployment. 
On the other hand, most existing benchmarks on FL focus on HFL scenario~\citep{chai2020fedeval,lai2022fedscale,zhang2023fedaudio}. For VFL, ~\citep{kang2022framework} evaluates several defense strategies for data reconstruction attacks; SLPerf~\citep{zhou2023slperf} focuses on benchmarking and comparing various kinds of splitNN scenarios like splitVFL. No existing work provides a comprehensive evaluation covering a variety of key aspects of VFL settings, including model performance, communication efficiency and robustness to attacks. Due to space limitation, %{\color{red}{we only include previous works on VFL Framework and Benchmarks here}} but 
we discuss works on VFL definition and emerging fields of research interest in \cref{sec:appendix_related_work}. %\yang{did we also remove paragraphs from original manuscript to Appendix? If so, need to explain what are removed exactly.} \tianyuan{No, at least currently, no. But I am planning to remove this to the appendix if the space is not enough.}
% and only discuss VFL related platform and benchmark works here to highlight the meaning and value of our work.

% To the best of our knowledge, none of the existing benchmarks provide comprehensive evaluations on both defense \textit{depth} and \textit{breadth} in VFL considering a broad range of existing attacks and VFL settings. 

% \begin{table}[]
%     \centering
%     \begin{tabular}{c|c|c|c|c|c}
%     \toprule
%          & Efficiency & Encryption & Defenses & Supporting Model & Specialized Field & \\
%     \midrule
%          FATE~\citep{FATE,liu2021fate} & \\
%          Fedlearner~\citep{Fedlearner} & \\
%          PaddleFL~\citep{PaddleFL} & \\
%          Pysyft~\citep{ryffel2018generic,romanini2021pyvertical} & \\
%          FedTree~\citep{fedtree2022} & \checkmark & Encryption \& DP & Tree & - \\
%          FedML~\citep{he2020fedml} & \\
%          \verb|VFLAIR| (ours) & \checkmark & 11 attacks \& 8 defenses \& Encryption & NN \& Tree & - \\
%     \bottomrule
%     \end{tabular}
%     \caption{Caption}
%     \label{tab:my_label}
% \end{table}

\section{VFL Framework} \label{sec:vfl_setting}
%\subsection{Vertical Federated Learning (VFL)} 
In a typical VFL setting with $K$ parties, each party owns their local private feature $\{X_{k}\}_{k=1}^K$  and local model $\{G_{k}\}_{k=1}^K$ with parameters $\{\theta_{k}\}_{k=1}^K$ respectively. Only one party controls the private label information $Y$ and is referred to as \textit{active} party while other parties are referred to as \textit{passive} parties.
% In this work, we mainly consider a VFL setting with $1$ active party and $1$ passive party, denote as party $a,p$ respectively, with each party owning their local feature $X_a, X_p$ and local model $G_a, G_p$ with parameters $\theta_a,\theta_p$ respectively. \yang{this should be a general description with $K$ parties, consistent with algorithms.} The active party also controls the private label information $Y$. 
The active party also controls a global trainable model parameterized by $\varphi$ (splitVFL) or global non-trainable function $F$ (aggVFL) to aggregate each party's local model output. Note in tree-based VFL the global function is an aggregation function that identifies the optimal feature split based on feature splitting information received from all parties. %The classification task is also regarded as the main task since it is the purpose for the collaboration between parties and reflects the utility of the VFL model. 
Without loss of generality, we assume that the $K^{th}$ party is the active party while other $K-1$ parties are passive parties.

In the collaborative training process of NN-based VFL, each party computes its local feature embedding $H_k=G_k(X_k,\theta_k), k=1,\dots,K$. The active party collects $\{H_k\}_{k=1}^K$ and gets the final prediction $\hat{Y}=F(H_1,\dots,H_K, \varphi)$. %Based on whether $F$ is a trainable model or a non-trainable function, we follow~\citep{liu2022vertical} to categorize VFL systems into splitVFL and aggVFL separately. 
The loss $\mathcal{L}=\ell(Y,\hat{Y})$ is calculated at the active party. The gradient w.r.s. to $H_k$ as $g_k = \frac{\partial \mathcal{L}}{\partial H_k}, k=1,\dots,K$ are then calculated and transmitted back to each party by the active party. Using these gradients, each party performs local model updates by SGD using $\nabla_{\theta_k}\mathcal{L} = \frac{\partial \mathcal{L}}{\partial \theta_k}=\frac{\partial \mathcal{L}}{\partial H_k}\frac{\partial H_k}{\partial \theta_k}, k=1,\dots,K$. Also the active party performs model update with SGD on global model $F$ if it is trainable using $\nabla_{\varphi}\mathcal{L} = \frac{\partial \mathcal{L}}{\partial \varphi}$. In the inference procedure, the same is done but without the backward gradient descent to get the prediction of labels. If the exchange of $H_k$ and $g_k$ is performed each round, such VFL protocol is referred as FedSGD protocol. On the other hand, if communication is done every $Q>1$ steps of local updates, such protocol is referred to as FedBCD~\citep{liu2022fedbcd}. %{\color{red}{Also, communication protocols that quantifies $H_k$ (using only $2^{b}$ bits for each element), or sparsifies $H_k$ by preserving only Top-k useful elements (using only $0<r<1$ proportion of the total elements), before information transmission to reduce communication cost, are referred to as Quantize~\citep{castiglia2022compressed} and Top-k~\citep{castiglia2022compressed} respectively. Further, cached FedBCD communication protocol using different batches of data for local update steps are termed CELU-VFL~\citep{fu2022towards}.}}
The training procedure is shown in detail in \cref{alg:NN_VFL_setting} in \cref{sec:appendix_vfl_framework}. Training and inference procedures of tree-based VFL are included in \cref{alg:Tree_VFL_setting} in \cref{sec:appendix_vfl_framework}.

% In tree-based VFL, the active party first broadcasts the set of record indices for the current node. Next, each passive party calculates the percentiles for each feature based on those indices. The passive party then proceeds to create binary splits for each feature by comparing the feature value of each instance to the percentile values. After that, the passive party sends back the statistics of each split necessary for evaluation such as purity to the active party, and the active party selects the best split using a specific evaluation function. To be precise, Random Forest employs gini impurity for classification, while XGBoost utilizes its gain function, which is based on the gradient and hessian. Finally, the active party requests the owner of the best split to send the set of record indices for the children nodes generated by the best split. Tree-based VFL system continues these procedures recursively until certain stop conditions, like depth constraints, are satisfied. \cref{alg:Tree_VFL_setting}
% % (in the appendix) 
% demonstrates training details of tree-based VFL.

\section{Overview of VFLAIR}

%\yang{add the core features of VFLAIR here.} \tianyuan{DONE}
\textbf{Implemented Components.} An overview of the components of \verb|VFLAIR| is shown in \cref{fig:VFLAIR}.
% As shown in \cref{fig:VFLAIR}, 
\verb|VFLAIR| incorporates not only basic VFL training and testing process for both NN-based and tree-based VFL of various settings, but also multiple existing efficiency enhancement techniques, data leakage and model utility impairing attacks as well as defending methods that aim to mitigate potential threats. \verb|VFLAIR| provides support for both aggVFL and splitVFL with easily adjustable model architectures. %For tree-based VFL, \verb|VFLAIR| accommodates XGBoost and Random Forest for aggregation. 
Currently, \verb|VFLAIR| supports $5$ communication protocols %including FedSGD (vanilla protocol), FedBCD~\citep{liu2022fedbcd}, CELU-VFL~\citep{fu2022towards}, Quantize~\citep{castiglia2022compressed} and Top-k~\citep{castiglia2022compressed} 
to improve communication efficiency. Also, $11$ existing attacks and $8$ 
defenses are supported. Moreover, \verb|VFLAIR| supports the comprehensive assessment of defense performance using carefully designed metrics (see \cref{sec:metrics_definition}), based on which defense strategy recommendations can be provided. Paillier Encryption~\citep{cheng2021secureboost} is also supported to further protect transmitted results.
% (See \cref{alg:encrypted_NN_VFL_setting})
In total, $13$ datasets from  a diverse range of industrial domains, including but not limited to medical, financial, and recommendation
% systems 
are supported. Datasets are partitioned into either balanced or unbalanced subsets to mimic real-world participants based on human knowledge (see \cref{tab:datasets} in \cref{subsec:appendix_model_dataset} for a full list of dataset and partition methods).
%{\color{red}{Last but not least, real world datasets are also supported in \verb|VFLAIR|, like Criteo and Avazu for click through prediction in advertising, Adult Income for finance, Cora for citation network and News20 of real world news data. }}

\textbf{How to use and extend.} \verb|VFLAIR| is a light-weight and comprehensive VFL framework that can be launched on a single GPU or CPU (see \cref{tab:hardware} for its system requirement compared to FATE). \verb|VFLAIR| facilitates the easy integration of different datasets for model training and inference through simple dataset loading and partitioning functions. New attacks and defenses can be quickly incorporated into the framework thanks to the modular structure.  % VFL model utility and its susceptibility to attacks under various defense strategies by using $5$ carefully designed evaluation metrics. Furthermore,
Step-by-step guidance for using and extending VFLAIR are included in \cref{section:appendix_user_guidance}. The workflow of \verb|VFLAIR| is also shown in \cref{section:appendix_vfl_workflow}.

%\yang{Add one example here would be better.} \tianyuan{I am afraid there is not space left for us to include an example.}

\section{VFL Benchmark} 

\subsection{VFL Settings, Models and Datasets} 
% \label{subsec:vanilla_vfl}

% % Please add the following required packages to your document preamble:
% % \usepackage{booktabs}
% \begin{table}[!th]
% \centering
% \begin{tabular}{@{}c|c|c|c@{}}
% \toprule
% Dataset & \#Classes & \#Samples   & Local Models         \\ \midrule
% Credit~\citep{Dua:2019}  & 2         & 20000:10000 & RandomForest/XGBoost \\
% Nursery~\citep{Dua:2019} & 5         & 8640:4320   & RandomForest/XGBoost \\ \bottomrule
% \end{tabular}
% \vspace{0.5em}
% \caption{Summary of evaluated datasets under tree-based VFL. In "\#Samples" column, the values denote the number of training and testing samples separately.}
% \label{tab:datasets}
% \end{table}

Using \verb|VFLAIR|, We benchmark the VFL main task performance using $13$ datasets including MNIST~\citep{MNISTdataset}, CIFAR10~\citep{krizhevsky2009learning}, CIFAR100~\citep{krizhevsky2009learning}, NUSWIDE~\citep{NUSWIDEdataset}, Breast Cancer~\citep{street1993nuclear}, Diabetes~\citep{Diabetes1999dataset}, Adult Income~\citep{AdultIncome1996dataset}, Criteo~\citep{guo2017deepfm_Criteo}, Avazu~\citep{qu2018product_Avazu}, Cora~\citep{mccallum2000automating_Cora}, News20~\citep{lang1995News20},Credit~\citep{Dua:2019} and Nursery~\citep{Dua:2019}.
Detailed data partition strategies are included in \cref{subsec:appendix_model_dataset}.
% \yang{how is the data partitioned? How many parties in the VFL setting? This is the place to define terms mentioned in the following, such as $H_a$, $H_p$} \tianyuan{Added.}
We explore $2$ distinct architectures, namely aggVFL and splitVFL, and comprehensively benchmark their performance. The local models used for both settings are detailed in \cref{tab:datasets} in \cref{subsec:appendix_model_dataset}. For global model $F$, a global softmax function is applied under aggVFL setting while a 1-layer fully-connected model serves as the global model for splitVFL setting (except for Cora dataset, for which a 1-layer graph convolution layer is applied). 
% Besides, FedBCD is also applied so as to benchmark the main task performance of VFL systems that leverage efficiency enhancement techniques. 
Additionally, we investigate the impact of different communication protocols by comparing FedBCD~\citep{liu2022fedbcd} ($Q=5$) and CELU-VFL~\citep{fu2022towards} ($Q=5,W=5$), as well as compression mechanisms Quantize ($b=16$)~\citep{castiglia2022compressed} and Top-k ($r=0.9$)~\citep{castiglia2022compressed} to the conventional FedSGD, as discussed in \cref{sec:vfl_setting} and further provide insights into the communication cost reduction achieved by communication efficient protocols,
% compared to FedSGD 
as well as the impact of FedBCD when various attacks and defenses are deployed. % demonstrating its effectiveness in enhancing efficiency.
We also evaluate the impact of the number of participating parties as well as the type of local model (logistic regression, tree, NN) on the main task performance of VFL. For tee-based VFL, we further benchmark both Random Forrest and XGBoost algorithms. Moreover, for 
% both NN-based and 
tree-based VFL, we employ Paillier Encryption~\citep{cheng2021secureboost} to protect transmitted information and measure its impact on computation efficiency. %such as labels and gradients. {\color{red}{We measure the execution time along side the model performance across various tree-based VFL aggregation algorithms with and without the application of Paillier Encryption, shedding light on the execution overhead incurred by encryption.}} 
Details of corresponding datasets and models for tree-based VFL can be found in \cref{tab:datasets} and \cref{subsec:appendix_model_dataset} in the appendix.
% {\color{red}{Furthermore, we benchmark $8$ attacks with $11$ defense methods on MNIST~\citep{MNISTdataset}, CIFAR10~\citep{krizhevsky2009learning}, NUSWIDE~\citep{NUSWIDEdataset} and Breast Cancer~\citep{street1993nuclear} in the context of NN-based VFL.}}
% \yang{to make the evaluation stronger, add experiments on more parties.e.g. 4-8., models: LR, tree, NN, more communication protocols other than FedBCD}

% The details of these datasets and the corresponding models used for NN-based VFL can be found in \cref{tab:datasets} and those for tree-based VFL can be found in \cref{tab:datasets}.
% More details on datasets are provided in \cref{subsec:appendix_model_dataset}.
% \tianyuan{@zixuan, please add an introduction of the dataset settings (what is includes, how much sample, data partition, shuffle or not) in the appendix.}

\subsection{Attacks and Defenses} \label{subsec:attacks}
%We classify the existing attacks into three types: \textbf{Label Inference (LI) Attacks}, \textbf{Feature Reconstruction(FR) Attacks} and \textbf{Backdoor Attacks}. Moreover, \textbf{Backdoor Attacks} can be further categorized into \textbf{Targeted Backdoor (TB) Attacks} and \textbf{Non-targeted Backdoor (NTB) Attacks}.We use $\mathcal{A}=\{\text{LI}, \text{FR}, \text{TB}, \text{NTB}\}$ to denote the types of attacks.
 
\begin{table}[!tb]
\vspace{-1em}
\caption{Summary of attacks for NN-based VFL}
\label{tab:attacks}
\resizebox{0.998\linewidth}{!}{
    % \small
    \begin{tabular}{c|c|c|c|c}
    \toprule
    Attack Type & Attack & \shortstack{Requirements\\/ Limitations} & \shortstack{Attacker\\Party} & Attack Performance (AP) \\
    \midrule
    \multirow{7}{*}{\shortstack{Label\\Inference\\(LI)}} & \shortstack{Norm-based Scoring (NS)~\citep{li2022label}} & \multirow{2}{*}{\shortstack{binary classification,\\sample-level}} & \multirow{7}{*}{passive} & \multirow{2}{*}{\shortstack{AUC of\\inferred labels}}\\
    \cline{2-2}
    \\[-1em]
     & \shortstack{Direction-based Scoring (DS)~\citep{li2022label}} &  &  & \\
    \cline{2-3}
    \cline{5-5}
    % \cline{2-5}
    \\[-1em]
     % & Spectal Attack (SA)~\citep{sun2022label} &  &  & \\
    % \cline{2-3}
    % \\[-1em]
     & \shortstack{Direct Label Inference (DLI)~\citep{li2022label,zou2022defending}} & sample-level &  & \multirow{5}{*}{\shortstack{ratio of correctly\\inferred labels\\ \,}}\\
    \cline{2-3}
    \\[-1em]
    & Batch-level Label Inference (BLI)~\citep{zou2022defending} & - &  & \\
    \cline{2-3}
    \\[-1em]
    & Passive Model Completion (PMC)~\citep{fu2021label} & \multirow{2}{*}{\shortstack{auxiliary labeled\\data for each class}} &  & \\
    \cline{2-2}
    \\[-1em]
    & Active Model Completion (AMC)~\citep{fu2021label} &  &  & \\
    \cline{1-5}
    \\[-1em]
    
    % \multirow{3}{*}{\shortstack{Feature\\Reconstruction\\(FR)}} & CAFE~\citep{jin2021cafe} & \shortstack{black-box (victim's\\model architecture)} & active & \multirow{3}{*}{\shortstack{reduction in MSE\\between recovered\\and real image\\compared to\\random guess}} & \multirow{3}{*}{0.0} \\
    % \cline{2-4}
    % \\[-1em]
    %  & \shortstack{Generative Regression Network\\(GRN)~\citep{luo2021feature}} & \shortstack{black-box (query\\victim's model)} & \multirow{2}{*}{\shortstack{active /\\ passive}} & \\
    % \cline{2-3}
    % \\[-1em]
    %  & \shortstack{Training-based Back Mapping by\\model inversion (TBM)~\citep{li2022ressfl}} & \shortstack{auxiliary\\i.i.d. data} &  & \\
    
    \multirow{3}{*}{\shortstack{Feature\\Reconstruction\\(FR)}} & \shortstack{Generative Regression Network (GRN)~\citep{luo2021feature}} & \shortstack{black-box} & \multirow{3}{*}{\shortstack{active}} & \multirow{3}{*}{\shortstack{$1-\text{MSE}(U_0,U_{rec})$ }}\\
    \cline{2-3}
    \\[-1em]
     & \shortstack{Training-based Back Mapping by\\model inversion (TBM)~\citep{li2022ressfl}} & \shortstack{white-box,\\auxiliary i.i.d. data} &  & \\
    \cline{1-5}
    \\[-1em]
    \shortstack{Targeted\\Backdoor (TB)\\ \,} & \shortstack{Label Replacement\\Backdoor (LRB)~\citep{zou2022defending}} & \shortstack{$\geq1$ sample of\\target class} & \shortstack{passive\\ \,\\ \,} & \shortstack{ratio of triggered samples\\inferred as target class}\\
    \cline{1-5}
    \\[-1em]
    \multirow{2}{*}{\shortstack{Non-targeted\\Backdoor (NTB)}} & Noisy-sample Backdoor (NSB)~\citep{zou2023mutual} & - & \multirow{2}{*}{passive} & \multirow{2}{*}{\shortstack{MP difference between total\\and noisy/missing samples}}\\
    % \cline{2-3}
    % \\[-1em]
     % & {\color{red}{Adversarial-Sample Backdoor (ASB)~\citep{??}}} & {\color{red}{???}} &  &  &  \\
    \cline{2-3}
    \\[-1em]
     & Missing Feature (MF)~\citep{liu2021rvfr} & - &  & \\
    \bottomrule
    \end{tabular}
}
\vspace{-1.5em}
\end{table}

% \subsubsection{Label Inference (LI) Attacks}
We benchmark the performance of $11$ attacks with $8$ defenses on $3$ datasets including MNIST~\citep{MNISTdataset}, CIFAR10~\citep{krizhevsky2009learning} and NUSWIDE~\citep{NUSWIDEdataset}.
% and Breast Cancer~\citep{street1993nuclear}.
For these evaluations, we mainly consider a VFL setting with $1$ active party and $1$ passive party following original works~\citep{li2022label,luo2021feature,li2022ressfl}, denoted as party $a,p$ respectively, with each party owning their local feature $X_a, X_p$ and local model $G_a, G_p$.
% with parameters $\theta_a,\theta_p$. 
The local model output of the active and passive party are denoted as $H_a, H_p$ respectively.
We summarized the evaluated attacks in \cref{tab:attacks}. Note in \cref{tab:attacks}, NS and DS attacks can only be applied to binary classification scenarios; "sample-level" indicates that the attack requires gradient information for each sample, whereas "batch-level" means only batch-level gradients information are available; "black-box" indicates that the model is kept private at the party under attack, but can be queried by the attacker and honestly return the output to the attacker, whereas "white-box" means the attacker has access to the model; $\text{MSE}(U_0,U_{rec})=\mathbb{E}[(u_0^{(f)}-u_{rec}^{(f)})^2]$ where $u_0^{(f)},u_{rec}^{(f)}$ are the $f^{th}$ feature of original input $U_0$ and recovered input $U_{rec}$ respectively. LI, FR and NTB attacks are inference time attacks that are launched separately from VFL training procedure while only TB attacks are training time attacks. 
Detailed descriptions and hyper-parameters of all the evaluated attacks are provided in \cref{sec:appendix_attacks,subsec:appendix_attack_parameters}, mainly following the original papers. Defense methods are summarized in \cref{tab:defense_param} with respective hyper-parameters. 
Detailed descriptions of defense methodologies and implementations are included in \cref{sec:appendix_defenses,subsec:appendix_defense_parameters}. 

\begin{table}[!tb]
\vspace{-1em}
\caption{Summary of defense methods and tested hyper-parameter values for NN-based VFL.} % \tianyuan{parameter double check.}
\label{tab:defense_param}
%%%%%%%%%%%%%%%%%% tabel without methodology %%%%%%%%%%%%%%%%%%
\resizebox{0.998\linewidth}{!}{
    \centering
    \begin{tabular}{c|c|c|c}
    \toprule
        Defense & Methodology & Hyper-parameter & Hyper-parameter Values \\
    \midrule
        G-DP~\citep{dwork2006DP,fu2021label,zou2022defending} & add noise to gradients or local prediction & DP Strength & $0.0001,0.001,0.01,0.1$\\
        L-DP~\citep{dwork2006DP,fu2021label,zou2022defending} & add noise to gradients or local prediction & DP Strength & $0.0001,0.001,0.01,0.1$\\
        GS~\citep{aji2017sparse,fu2021label,zou2022defending} & drop gradient elements close to $0$ & Sparsification Rate & $95.0\%,97.0\%,99.0\%,99.5\%$\\ %
        GPer~\citep{yang2022differentially} & perturb gradient with that of other class & Perturbation Strength & $0.0001,0.001,0.01,0.1$\\
        dCor~\citep{sun2022label,vepakomma2019reducing} & distance correlation regularization & Regularizer Strength & $0.0001,0.01,0.1,0.3$\\
        CAE~\citep{zou2022defending} & disguise label & Confusion Strength $\lambda$ & $0.0,0.1,0.5,1.0$\\
        DCAE~\citep{zou2022defending} & discrete gradient in addition to CAE & Confusion Strength $\lambda$ & $0.0,0.1,0.5,1.0$\\
    MID~\citep{zou2023mutual} & mutual information (MI) regularization & Regularizer Strength $\lambda$ & $0.0,1e^{-8},1e^{-6},1e^{-4},0.01,0.1,1.0,1e^{2},1e^{4}$ \\
    \bottomrule
    \end{tabular}
% \vspace{-2em}
}
%%%%%%%%%%%%%%%%%% tabel with methodology %%%%%%%%%%%%%%%%%%
% %%%%%%%%%%%%%%%%%% tabel without methodology %%%%%%%%%%%%%%%%%%
%     \centering
%     \begin{tabular}{c|c|c}
%     \toprule
%         Defense & Hyper-parameter & Hyper-parameter Values \\
%     \midrule
%         G-DP~\citep{dwork2006DP,fu2021label,zou2022defending} & DP Strength & $0.0001,0.001,0.01,0.1$\\
%         L-DP~\citep{dwork2006DP,fu2021label,zou2022defending} & DP Strength & $0.0001,0.001,0.01,0.1$\\
%         GS~\citep{aji2017sparse,fu2021label,zou2022defending} & Sparsification Rate & $95.0\%,97.0\%,99.0\%,99.5\%,100.0\%$\\ 
%         GPer~\citep{yang2022differentially} & Perturbation Strength & $0.0001,0.001,0.01,0.1$\\
%         dCor~\citep{sun2022label,vepakomma2019reducing} & Correlation Strength & $0.0001,0.001,0.01,0.1$\\
%         CAE~\citep{zou2022defending} & Confusion Strength $\lambda$ & $0.0,0.1,0.5,1.0$\\
%         DCAE~\citep{zou2022defending} & Confusion Strength $\lambda$ & $0.0,0.1,0.5,1.0$\\
%         MID~\citep{zou2023mutual} & $\lambda$ & $0.0,1e^{-8},1e^{-6},1e^{-4},0.01,0.1,0.5,1.0$ \\
%     \bottomrule
%     \end{tabular}
% %%%%%%%%%%%%%%%%%% tabel without methodology %%%%%%%%%%%%%%%%%%
\vspace{-1.5em}
\end{table}

\subsection{Evaluation Metrics}\label{sec:metrics_definition}

% To fairly evaluate attacks and defenses in a VFL system, we identify two crucial metrics:  \textit{\textbf{Attack Performance (AP)}} and \textit{\textbf{Main Task Performance (MP)}}.

\textbf{Main Task Performance (MP).}
%For the main task performance, as the task we used are all classification tasks in out experiments, we select \textit{the correctly classified samples}, also the \textit{main task accuracy} as the evaluation metrics. A detailed definition is: 
MP is defined as the final model prediction accuracy on the test dataset, which reveals the utility of the VFL system.
% MP is an important metric to evaluate defenses because a good defense should preserve the performance of the main task of VFL as much as possible. 

\textbf{Communication and Computation Efficiency.} Number of communication rounds (\#Rounds) and the amount of data transferred for each round (Amount) are used for measuring communication efficiency. Execution Time (Exec.Time) is used to measure computation efficiency. 

\textbf{Attack Performance (AP).} %As we also include various kinds of attacks and defense in our platform and benchmark experiments, we define AP so as to evaluate the success rate of a given attack which also reflects the vulnerability of a VFL system to that given attack. 
The definition of AP varies with respect to the type of the attack and is summarized in \cref{tab:attacks}. Definition details can be seen in \cref{sec:appendix_ap}. 

\textbf{Defense Capability Score (DCS).} %To evaluate the defense methods, 
Intuitively, an ideal defense should not compromise the utility of the original main task and should thwart the attack completely. Therefore, considering that both AP and MP are key metrics to evaluate defenses. We further propose \textbf{Defense Capability Score (DCS)}, to directly compare all the defenses under one unified metric. Let $df=(\text{AP},\text{MP})$ represents the performance of a defense on an AP-MP graph, then we define its defense capability score (DCS) based on the distance between $df$ to an ideal defense $df^{*}=(\text{AP}^{*}, \text{MP}^{*})$.  MP$^{*}$ is the MP of VFL without defense and AP$^{*}$ is set to $0.0$ representing the performance of a completely incapable attacker. Then, we formulate the definition of DCS as:
\begin{equation} \label{eq:dcs}
 \setlength\abovedisplayskip{0.05cm}
 \setlength\belowdisplayskip{0.05cm}
    % D(df,df^{*}) &=  \\
    \text{DCS} = \frac{1}{1+D(df,df^{*})} = \frac{1}{1+\sqrt{(1-\beta)(\text{AP}-\text{AP}^{*})^2+\beta(\text{MP}-\text{MP}^{*})^2}},
\end{equation}
%\begin{equation} \label{eq:t-dds}
%    DSC = D[(\hat{m}_{j},\hat{a}_{j}), %%(\hat{m}_{j}^{*},\hat{a}_{j}^{*})] = \sqrt{(\hat{m}_{j}-%\hat{m}_{j}^{*})^2+\beta(\hat{a}_{j}-\hat{a}_{j}^{*})^2)},
%\end{equation}

\begin{wrapfigure}{l}{0.36\textwidth}
\vspace{-2em}
% \begin{figure}[H]
    \centering
    \includegraphics[width=0.99\linewidth]{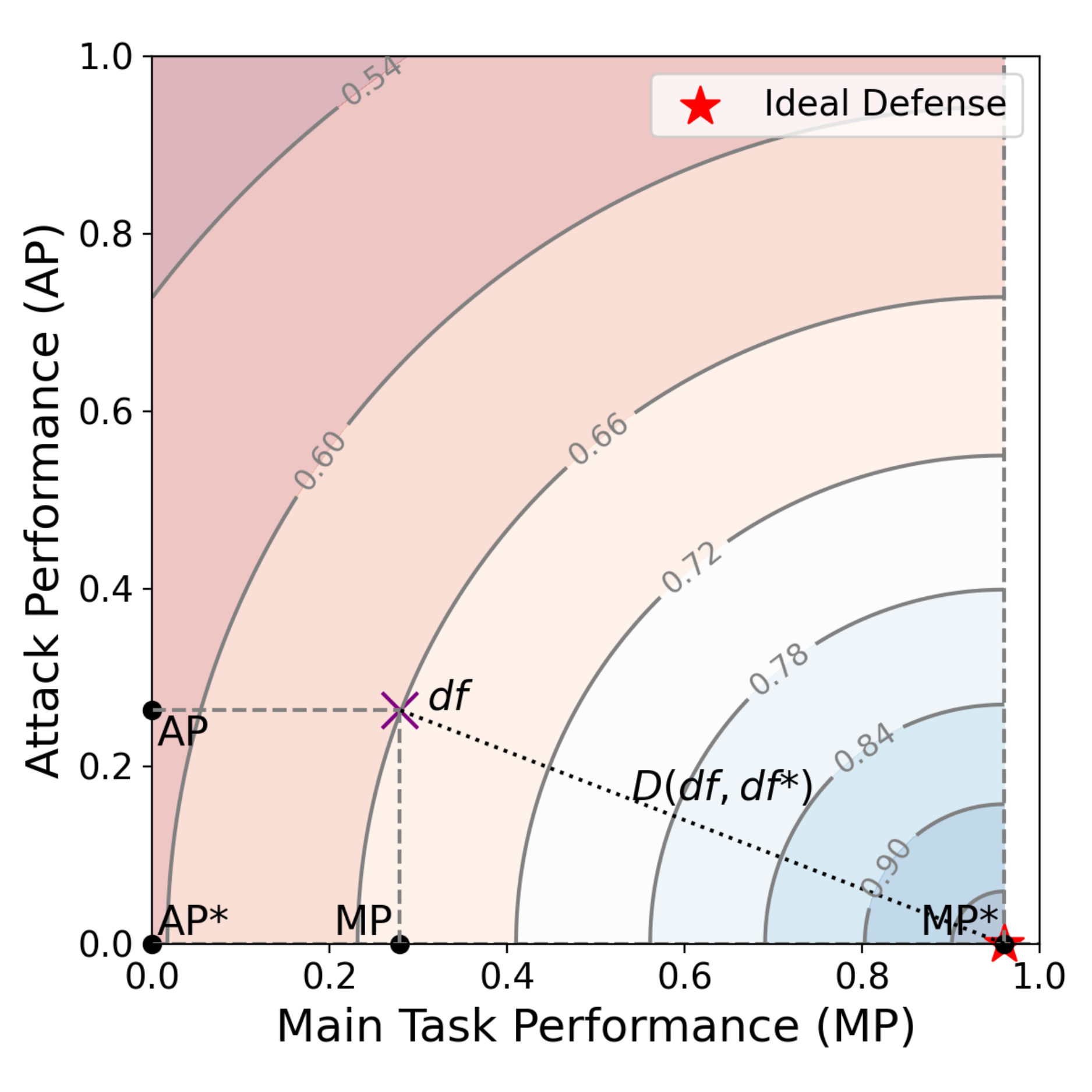}
    \vspace{-2em}
    \caption{A visual illustration example of DCS. The numbers on the contour lines are DCSs calculated with $\beta=0.5$.} % \yang{make the points larger and font bigger}
    \label{fig:dcs}
% \end{figure}
\vspace{-3em}
\end{wrapfigure}

where $D(\cdot)$ is a user-defined distance function. Here we use Euclidean distance with an adjustable trade-off weighting parameter $\beta$. A visualization of DCS on an AP-MP graph with $\beta=0.5$ can be seen in \cref{fig:dcs}. %\yang{add a sketch figure to illustrate the distance}\tianyuan{DONE, but this figure takes up too much space.}). 
A point closer to the bottom-right corner of an AP-MP graph has a higher DCS score indicating a better defense capability, consistent with intuition. $\beta=0.5$ is used in our experiments. %Also, $\hat{m}_{j}^{*}$ denotes the \textit{ideal MP} defined as the MP of a vanilla VFL without defense, representing no accuracy loss on the main task and $\hat{a}_{j}^{*}$ denotes the \textit{ideal AP} defined as the worst AP for each attack (each attack type) as defined above in \cref{subsubsec:ap_mp} and \cref{tab:attack_performance_definition}.

\textbf{Type-level Defense Capability Score (T-DCS).} T-DCS is the DCS score averaged by attack type. %proposed to evaluate the capability of a defense method on a certain type of attack. %we considered in \cref{subsec:attacks} rather than on a single attack. 
Treating all $I_j$ attacks of the same attack type $j$ as equally important, we average DCS for each attack $i$ to get T-DCS for attack type $j$:% $j\in \mathcal{A}$:
\begin{equation} \label{eq:t-dcs}
 \setlength\abovedisplayskip{0.05cm}
 \setlength\belowdisplayskip{0.05cm}
    \text{T-DCS}_j = \frac{1}{I_j} \sum_{i=1}^{I_j} \text{DCS}_i .
\end{equation}

\textbf{Comprehensive Defense Capability Score (C-DCS).} C-DCS is a comprehensive assessment of the capability of a defense strategy with respect to all kinds of attacks and is a weighted average of T-DCS as shown in \cref{eq:c-dcs}:
\begin{equation} \label{eq:c-dcs}
 \setlength\abovedisplayskip{0.05cm}
 \setlength\belowdisplayskip{0.05cm}
    \text{C-DCS} = \sum_{j\in \mathcal{A}} w_j \text{T-DCS}_j, \,\, \text{with} \sum_{j \in \mathcal{A}} w_j = 1.0 \, .
\end{equation}
Weights $\{w_j\}_{j \in \mathcal{A}}$ can be tailored to user preference. In our experiments, we simply use an unbiased weight $w_j=\frac{1}{|\mathcal{A}|}$ for each attack type $j \in \mathcal{A}=\{\text{LI}, \text{FR}, \text{TB}, \text{NTB}\}$.

\section{{Evaluation Results}}

\subsection{VFL Main Task Performance} \label{subsec:MP}
We first comprehensively evaluate the impact of various settings on the performance of VFL. Details on training model and training hyper-parameters for the following experiments are included in \cref{tab:datasets} (in \cref{subsec:appendix_model_dataset}) and \cref{subsec:appendix_main_task_parameters} respectively.

\textbf{Model Partition.}
The splitVFL setting yields a comparable or slightly higher MP  compared to aggVFL on most datasets, due to the additional trainable layer serving as global model, evidenced by results from \cref{tab:NN_MP,tab:4party_MP,tab:real_world_dataset_MP}.
% \yang{Since splitVFL has additional trainable parameters, it would be reasonable that splitVFL demonstrates higher MP, as evidenced in the Table for most dataset.}\tianyuan{We have introduced it in Sec.5.1. And from \cref{tab:NN_MP}, I don't think we can conclude that splitVFL is better than aggVFL.}\yang{then we need to explain why splitVFL is not better than aggVFL given that it has an extra FCC? or it could be just the experiments are not right.} \tianyuan{The theorem is that, "a simple two-layer neural network with 2n+d parameters is capable of perfectly fitting any dataset of n samples of dimension d"~\citep{zhang2021understanding}. So for CIFAR100 and Breast Cancer, it make sense that the MP under splitVFL is not higher than that under aggVFL. For NUSWIDE, we will tune the training hyper-parameter for better results.} 
% indicating that when the resources permits, splitVFL is a slightly better choice 
%indicating that these $2$ settings can both be selected for real-world application considering their similar main task utility {\color{red}{although splitVFL may perform a little better}}.

\textbf{Communication Protocols.} %We compare $5$ different kinds of communication protocols in VFL. 
As shown in \cref{tab:NN_MP} and \cref{tab:communication_MP}, compared to FedSGD, FedBCD and CELU-VFL exhibit comparable MP across all datasets with fewer communication rounds, supporting their efficacy in reducing communication overhead. Quantize and Top-k compress the transmitted data and successfully reduce the communication cost per round, but may result in an increase in communication rounds. %CELU-VFL also effectively reduces the \#Rounds compared to FedSGD, and is slightly less than that of FedBCD
% As shown in \cref{tab:NN_MP}, FedSGD and FedBCD exhibit comparable MP across all datasets, while FedBCD achieves the specified MP with fewer communication rounds than FedSGD. These together support the efficacy of the FedBCD algorithm in achieving global convergence while reducing communication overhead. 

\textbf{Encryption.} %{\color{red}{We compare the MP and execution time with and without HE protection using Paillier Encryption in 
% \cref{tab:encryption_MP} shows that implementing Paillier Encryption significantly elongated the execution time. %due to the encryption and decryption process. %Note that, our current implementation on Paillier Encryption supports only linear calculation like shown in \cref{alg:encrypted_NN_VFL_setting}, which means a global averaging function instead of softmax function is adopted resulting in a MP slightly affected.}} \yang{not sure how this Encryption is implemented. add a full algorithm in Appendix.}
%{\color{red}{Currently, encryption is supported in \verb|VFLAIR| only for tree-based VFL setting.}}
For tree-based VFL, we consider two models with and without Paillier Encryption using 512-bit key size in \cref{tab:tree_MP}. Note that XGBoost with Paillier Encryption is equivalent to SecureBoost~\citep{cheng2021secureboost}. Although MP values are consistent regardless of encryption, the execution time experiences a notable increase of 3 to 20 times when encryption is applied due to the additional encryption and decryption process. 

\textbf{Number of Participants.}
Impact of number of participants are shown in \cref{tab:4party_MP}. 
% Compared to 2-party VFL system, a 
A slightly lower MP is achieved using fewer communication rounds as the number of participants increases, demonstrating the increasing challenges brought by multi-party collaboration.

%\textit{\textbf{Tree-based VFL and its comparison with NN-based VFL.}}
\textbf{Model Architectures.}
\cref{tab:different_local_models_MP,tab:tree_MP} (in \cref{subsec:appendix_MP}) compare the MP of different model architectures and show that different model architectures result in slightly different MP for each dataset, highlighting the importance of selecting best performing model for each dataset.% {\color{red}{tree-based VFL performs better on Credit dataset, while NN-based VFL performs better on Nursery dataset.
% these tabular datasets, while local neural network model and linear regression model each performs better with different datasets.
%}}

\textbf{Real world datasets.}
Additional results on Criteo~\citep{guo2017deepfm_Criteo}, Avazu~\citep{qu2018product_Avazu}, Cora~\citep{mccallum2000automating_Cora} and News20~\citep{lang1995News20} datasets using domain specific models (e.g. Wide\&Deep Model~\citep{cheng2016wide} for Criteo and Avazu, GNN for Cora) are provided in \cref{tab:real_world_dataset_MP}
% in \cref{subsec:appendix_MP}
, as they are considered for typical VFL applications, such as in recommendation problems. %click through rate prediction in advertising, node classification in citation network and classification of news data

\begin{table}[!tb]
\vspace{-1em}
\caption{MP under $4$ different settings of NN-based VFL. $Q=5$ when FedBCD is applied. In "\#Rounds" column, the first and second numbers are the communication rounds needed to reach the specified MP for FedSGD and FedBCD respectively.}
\label{tab:NN_MP}
\resizebox{0.998\linewidth}{!}{
    \centering
    \begin{tabular}{c||c|c|c||c|c|c}
    \toprule
    Dataset & aggVFL, FedSGD & aggVFL, FedBCD & \#Rounds & splitVFL, FedSGD & splitVFL, FedBCD & \#Rounds \\
    \midrule
    MNIST & 0.972$\pm$0.001 & 0.971 $\pm$0.001 & 150 / 113 & 0.973$\pm$0.001 & \textbf{0.974$\pm$0.001} & 180 / 143 \\
    % CIFAR10 & 0.790$\pm$0.003 & 0.762$\pm$0.004 & ?? / ?? & \textbf{0.798$\pm$0.010} & 0.771$\pm$0.013 & ?? / ?? \\
    % CIFAR100 & 0.454$\pm$0.006 & \textbf{0.487$\pm$ 0.005} & ?? / ?? & 0.423$\pm$0.005 & 0.443$\pm$0.004 & ?? / ?? \\
    NUSWIDE & 0.887$\pm$0.001 & 0.882$\pm$0.001 & 60 / 26 & \textbf{0.888$\pm$0.001} & 0.884$\pm$0.001 & 60 / 29 \\ 
    Breast Cancer & 0.914$\pm$0.033 & 0.919$\pm$0.029 & 5 / 3 & \textbf{0.925$\pm$0.028} & 0.907$\pm$0.045 & 5 / 4 \\
    Diabetes & 0.755$\pm$0.043 & 0.736$\pm$0.021 & 15 / 13 & \textbf{0.766$\pm$0.024} & 0.746$\pm$0.039 & 15 / 11 \\
    Adult Income & 0.839$\pm$0.006 & 0.841$\pm$0.005 & 17 / 15 & 0.842$\pm$0.004 & \textbf{0.842$\pm$0.005} & 30 / 13 \\
    \bottomrule
    \end{tabular}
}
\vspace{-1em}
\end{table}

\begin{table}[!tb]
\caption{MP under 2-party VFL verses MP under 4-party VFL  under $4$ different settings of NN-based VFL using FedSGD communication protocol. "\#Rounds" has the same meaning as in \cref{tab:NN_MP}.}
\label{tab:4party_MP}
\resizebox{0.998\linewidth}{!}{
    \centering
    \begin{tabular}{c|c||c|c||c|c}
    \toprule
    Dataset  & ~ & aggVFL, 2-party & aggVFL, 4-party & splitVFL, 2-party & splitVFL, 4-party\\
    \midrule
    \multirow{2}{*}{CIFAR10} & MP & 0.790$\pm$0.003 & 0.747$\pm$0.003 & \textbf{0.798$\pm$0.010} & 0.762$\pm$0.003 \\
    ~ & \#Rounds & 244$\pm$16 & 205$\pm$12 & 238$\pm$14 & 173$\pm$3 \\
    \midrule
    \multirow{2}{*}{CIFAR100} & MP & \textbf{0.454$\pm$0.006} & 0.417$\pm$0.008 & 0.423$\pm$0.005 & 0.382$\pm$0.004 \\
    ~ & \#Rounds & 130$\pm$11 & 124$\pm$2 & 125$\pm$2 & 100$\pm$1 \\
    \bottomrule
    \end{tabular}
}
\vspace{-1em}
\end{table}

% \begin{table}[!tb]
% \caption{Comparison MP using different local models in NN-based VFL with FedSGD communication protocol. We mainly follow~\citep{ye2022feature} for the selection of MLP model architecture.}
% \label{tab:different_local_models_MP}
% \resizebox{0.998\linewidth}{!}{
%     \centering
%     \begin{tabular}{c|c||c|c||c|c}
%     \toprule
%     \multirow{2}{*}{} & \multirow{2}{*}{} & \multicolumn{2}{c||}{Credit} & \multicolumn{2}{c}{Nursery} \\
%     \cline{3-6}
%     \\[-1em]
%     ~ & ~ & Linear Regression & Neural Network (MLP-4) & Linear Regression & Neural Network (MLP-3) \\
%     \midrule
%     \multirow{2}{*}{aggVFL} & MP & 0.806$\pm$0.001 & 0.772$\pm$0.006 & 0.938$\pm$0.001 & 0.999$\pm$0.001  \\
%     ~ & Exec.Time [s] & 286.30$\pm$29.51 & 14.73$\pm$4.27 & 1.51$\pm$1.23 & 1.35$\pm$0.76 \\
%     \midrule
%     \multirow{2}{*}{splitVFL} & MP & 0.804$\pm$0.001 & 0.779$\pm$0.001 & 0.931$\pm$0.005 & 0.999$\pm$0.001 \\
%     ~ & Exec.Time [s] & 252.06$\pm$39.29 & 3.85$\pm$2.40 & 1.50$\pm$0.75 & 3.69$\pm$2.57 \\
%     \bottomrule
%     \end{tabular}
% }
% \vspace{-1em}
% \end{table}

\begin{table}[!tb]
\caption{MP and execution time under 2 different types of tree-based VFL.} %\yang{too many digits, round to fewer, especially for time}
\label{tab:tree_MP}
\resizebox{0.99\linewidth}{!}{
\centering
\begin{tabular}{@{}c|c|c|c|c|c@{}}
\toprule
Dataset &
   &
  \begin{tabular}[c]{@{}c@{}}Random Forest \\ w/o Encryption\end{tabular} &
  \begin{tabular}[c]{@{}c@{}}XGBoost\\ w/o Encryption\end{tabular} &
  \begin{tabular}[c]{@{}c@{}}Random Forest \\ w/ Encryption\end{tabular} &
  \begin{tabular}[c]{@{}c@{}}XGBoost\\ w/ Encryption\\ (a.k.a. SecureBoost)\end{tabular} \\ \midrule
\multirow{2}{*}{Credit} &
  % MP &
  % 0.8163$\pm$0.0027 &
  % 0.8161$\pm$0.0029 &
  % 0.8163$\pm$0.0027 &
  % 0.8161$\pm$0.0029 \\
  MP &
  0.816$\pm$0.005 &
  0.816$\pm$0.004 &
  0.816$\pm$0.005 &
  0.816$\pm$0.004 \\
 &
  % \multicolumn{1}{l|}{Exec.Time [s]} &
  % \multicolumn{1}{l|}{24.9240$\pm$2.0013} &
  % \multicolumn{1}{l|}{52.9372$\pm$3.5586} &
  % \multicolumn{1}{l|}{225.0062$\pm$9.2994} &
  % \multicolumn{1}{l}{1012.4125$\pm$18.0930} \\ 
  \multicolumn{1}{c|}{Exec.Time [s]} &
  \multicolumn{1}{c|}{138$\pm$4} &
  \multicolumn{1}{c|}{366$\pm$16} &
  \multicolumn{1}{c|}{410$\pm$10} &
  \multicolumn{1}{c}{881$\pm$6} \\ 
% \midrule
\hline
\\[-1em]
\multirow{2}{*}{Nursery} &
  % MP &
  % 0.8889$\pm$0.0136 &
  % 0.8928$\pm$0.0074 &
  % 0.8889$\pm$0.0136 &
  % 0.8928$\pm$0.0074 \\
  MP &
  0.884$\pm$0.010 &
  0.890$\pm$0.011 &
  0.884$\pm$0.010 &
  0.890$\pm$0.011 \\
 &
  % \multicolumn{1}{l|}{Exec.Time [s]} &
  % \multicolumn{1}{l|}{108.3264$\pm4.4093$} &
  % \multicolumn{1}{l|}{262.6055$\pm$7.2585} &
  % \multicolumn{1}{l|}{338.9600$\pm$9.4467} &
  % \multicolumn{1}{l}{705.6566$\pm$18.1272} \\ 
  \multicolumn{1}{c|}{Exec.Time [s]} &
  \multicolumn{1}{c|}{29$\pm$2} &
  \multicolumn{1}{c|}{69$\pm$4} &
  \multicolumn{1}{c|}{243$\pm$5} &
  \multicolumn{1}{c}{1194$\pm$21} \\ 
\bottomrule
\end{tabular}
}
\vspace{-1em}
\end{table}

\begin{table}[!tb]
% \vspace{-1em}
\caption{
MP, communication rounds (\#Rounds), amount of information exchanged per round (Amount) under different communication protocols of NN-based VFL under aggVFL setting. %$Q=5$ when FedBCD and CELU-VFL are used, otherwise $Q=1$. For Quantize, $b=16$ while for Top-k, top $90\%$ of elements are kept in forward local model prediction.
'Total' column is the total amount that equals to \#Rounds$\times$Amount.}
\label{tab:communication_MP}
\resizebox{0.99\linewidth}{!}{
    \centering
    \begin{tabular}{c|cccc|cccc}
    \toprule
        ~ & \multicolumn{4}{c|}{MNIST} & \multicolumn{4}{c}{NUSWIDE} \\
        \cline{2-9}
        \\[-1em]
        ~ & MP & \#Rounds & Amount (MB) & Total (MB) & MP & \#Rounds & Amount (MB) & Total (MB)\\ 
    \midrule
        FedSGD & \textbf{0.972$\pm$0.001} & 150 & 0.156 & 23.438 & \textbf{0.887$\pm$0.001} & 60 & 0.039 & 2.344 \\
        FedBCD & 0.971$\pm$0.001 & 113 & 0.156 & 17.656 & 0.882$\pm$0.001 & 26 & 0.039 & 1.016 \\
        % Quantize(FedBCD) & 0.961$\pm$0.003 & {\color{red}{203}} & 0.117 & {\color{red}{23.789}} & 0.874$\pm$0.001 & 29 & 0.029 & 0.850 \\
        % Top-k(FedBCD) & 0.967$\pm$0.003 & {\color{red}{163}} & 0.148 & {\color{red}{24.195}} & 0.876$\pm$0.001 & 29 & 0.037 & 1.076 \\
        Quantize & 0.959$\pm$0.006 & 161 & 0.117 & 18.867 & 0.881$\pm$0.002 & 94 & 0.029 & 2.754 \\
        Top-k & 0.968$\pm$0.001 & 150 & 0.148 & 22.266 & 0.887$\pm$0.001 & 60 & 0.037 & 2.227 \\
        CELU-VFL & 0.971$\pm$0.002 & 105 & 0.156 & 16.406 & 0.880$\pm$0.001 & 25 & 0.039 & 0.977 \\
    \bottomrule
    \end{tabular}
}
\vspace{-1em}
\end{table}

\begin{table}[!tb]
% \vspace{-1em}
\caption{Comparison of aggVFL and splitVFL on MP, \#Rounds, Amount, total communication cost, Exec.Time for reaching specified MP with $4$ real-world datasets of NN-based VFL with FedSGD communication protocol.}
\label{tab:real_world_dataset_MP}
\resizebox{0.99\linewidth}{!}{
    \centering
    \begin{tabular}{c|ccccc|ccccc}
    \toprule
        \multirow{2}{*}{Dataset} & \multicolumn{5}{c|}{aggVFL} & \multicolumn{5}{c}{splitVFL} \\
        \cline{2-11}
        \\[-1em]
        ~ & MP & \#Rounds & \shortstack{Amount\\(MB)} & \shortstack{Total\\(MB)} & Exec.Time [s] & MP & \#Rounds & \shortstack{Amount\\(MB)} & \shortstack{Total\\(MB)} & Exec.Time [s]\\
        % ~ & MP & \#Rounds & \shortstack{Amount\\(MB)} & \shortstack{Total\\(MB)} & \shortstack{Exec.Time\\[s]} & MP & \#Rounds & \shortstack{Amount\\(MB)} & \shortstack{Total\\(MB)} & \shortstack{Exec.Time\\[s]} \\
    \midrule
        Criteo & 0.715$\pm$0.053& 2 & 0.125 & 0.250 & 0.190$\pm$0.132 & 0.744$\pm$0.001 & 3 & 0.125 & 0.375 & 0.234$\pm$0.126\\
        Avazu & 0.832$\pm$0.001 & 5 & 0.125 & 0.625 & 0.517$\pm$0.185 & 0.832$\pm$0.001 & 9 & 0.125 & 1.125 & 1.203$\pm$1.516\\ 
        Cora & 0.721$\pm$0.004 & 11 & 0.145 & 1.591 & 0.205$\pm$0.085 & 0.724$\pm$0.012 & 13 & 0.145 & 1.880 & 0.270$\pm$0.082 \\
        News20-S5 & 0.882$\pm$0.014 & 57 & 0.005 & 0.278 & 0.430$\pm$0.076 & 0.893$\pm$0.013 & 61 & 0.005 & 0.298 & 0.613$\pm$0.269 \\
    \bottomrule
    \end{tabular}
}
% \vspace{-1em}
\end{table}

% \begin{table}[!tb]
% \caption{MP and communication time with Homomorphic Encryption protected encryption communication in NN-based VFL with aggVFL setting and  FedSGD communication protocol.}
% \label{tab:encryption_MP}
% \resizebox{0.99\linewidth}{!}{
%     \centering
%     \begin{tabular}{c||c|c|c|c|c|c}
%     \toprule
%         ~ & ~ & MNIST & \multicolumn{2}{c|}{Credit} & \multicolumn{2}{c}{Nursery}\\
%         \cline{3-7}
%         \\[-1em]
%         ~ & ~ & \shortstack{Neural\\Network (MLP-2)} & \shortstack{Logistic\\Regression} & \shortstack{Neural\\Network (MLP-4)} & \shortstack{Logistic\\Regression} & \shortstack{Neural\\Network (MLP-3)} \\
%     \midrule
%         \multirow{2}{*}{\shortstack{w/o\\Encryption}} & MP & 0.972$\pm$0.001 & 0.806$\pm$0.001 & 0.772$\pm$0.006 & 0.938$\pm$0.001 & 0.999$\pm$0.001 \\
%         ~ & Exec.Time [s] & 9.360$\pm$0.217 & 3.525$\pm$0.155 &  9.606$\pm$0.565 & 2.236$\pm$0.201 & 4.214$\pm$0.193 \\
%         \cline{1-7}
%         \\[-1em]
%         \multirow{2}{*}{\shortstack{w/\\Encryption}} & MP & 0.957$\pm$0.001 & 0.662$\pm$0.212 & 0.771$\pm$0.010 & 0.911$\pm$0.001 & 0.950$\pm$0.002\\
%         ~ & Exec.Time [s] & 4184.885$\pm$48.201 & 1462.131$\pm$27.655 & 1499.812$\pm$56.000 & 608.712$\pm$11.145 & 606.233$\pm$6.806 \\ 
%     \bottomrule
%     \end{tabular}
% }
% \vspace{-1em}
% \end{table}

\subsection{Attack and Defense Performance} \label{subsec:attack_defense_performance}

\begin{figure}[!htb]
  \centering
  % \vspace{-1em}
    \includegraphics[width=0.99\linewidth]{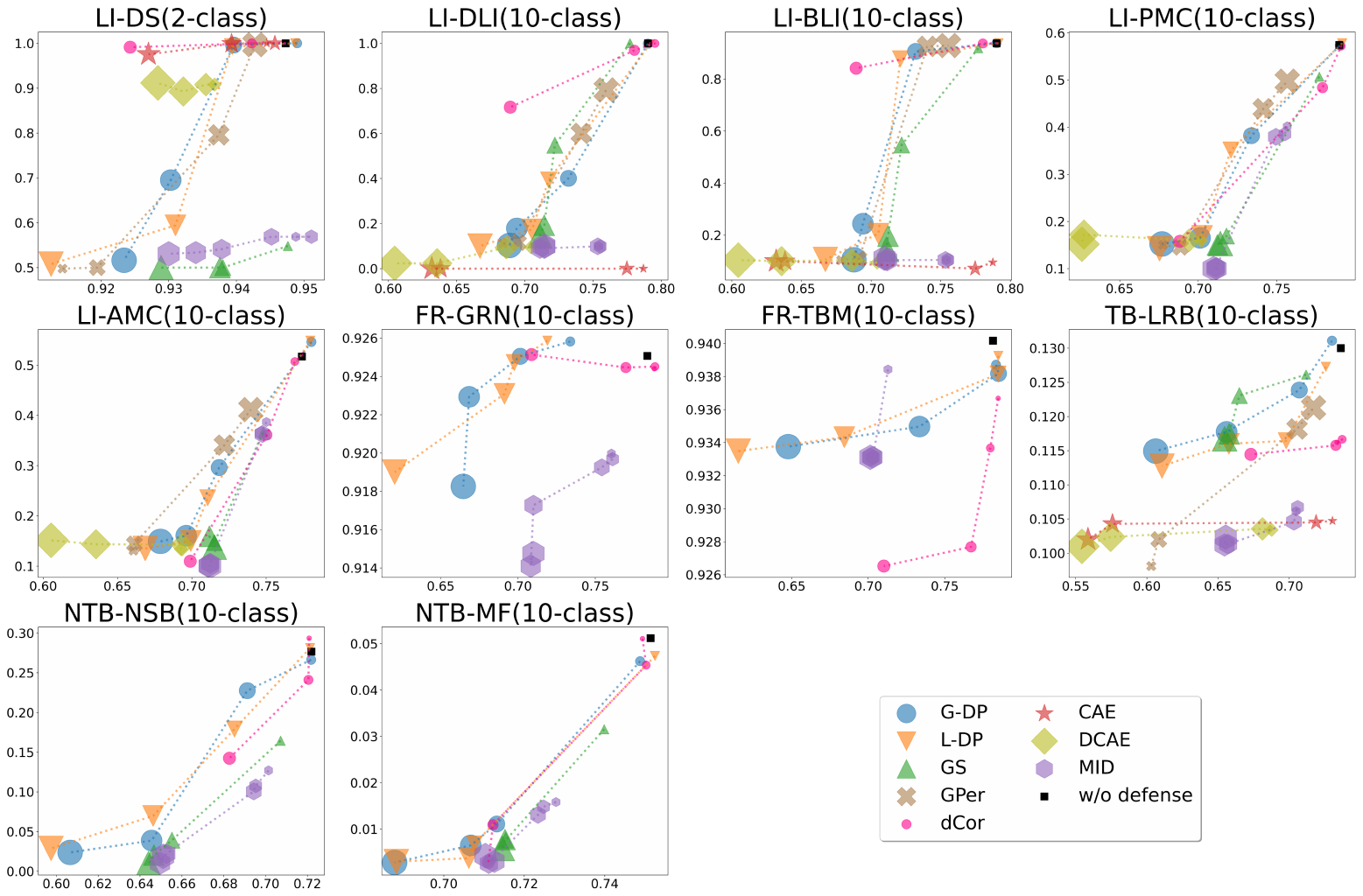}
  \vspace{-1em}
  \caption{MPs and APs for different attacks under defenses [CIFAR10 dataset, aggVFL, FedSGD]}
  \label{fig:cifar10_MPAP}
  \vspace{-1em}
\end{figure}

% \yang{datasets are a little too small. what about CIFAR100?} \tianyuan{CIFAR100 has the same amount of samples as CIFAR10 and thus are of the same size. Also, nuswide is a larger dataset compared to CIFAR100 and CIFAR10.} 
We demonstrate attack and defense results of VFL on the AP-MP graph for each attack on MNIST, CIFAR10 and NUSWIDE datasets under aggVFL setting using FedSGD protocol in \cref{fig:mnist_MPAP,fig:cifar10_MPAP,fig:nuswide_MPAP}.
%In each sub-figure of \cref{fig:mnist_MPAP}, the horizontal and vertical coordinates displays MP and AP respectively with different value range. 
Each point in the figure represents a $(\text{MP},\text{AP})$ pair with the size of markers representing the relative magnitude of the corresponding defense hyper-parameter listed in \cref{tab:defense_param}. %Application details of the defense methods and hyper-parameter descriptions are provided in \cref{subsec:appendix_defense_parameters} with each attack's detailed implementation parameters given in \cref{subsec:appendix_attack_parameters}.
% \tianyuan{Add some basic analysis?} \zixuan{TODO: add some take-aways from the figures} 
Note that although we try to provide comprehensive evaluation for various defenses, we do not force defense onto attacks, meaning that if %we evaluate defense methods only on the specific attacks they were designed for. In other words, if 
a defense mechanism is designed for mitigating label inference attacks only, we do not assess its effectiveness against FR attacks or backdoor attacks. 

We further rank all the defenses of different hyper-parameters based on their C-DCS. Due to space limitation, we show representative results for NUSWIDE dataset in \cref{tab:tiny_nuswide_dcs_ranking}. Full results are shown in \cref{tab:mnist_dcs_ranking,tab:cifar10_dcs_ranking,tab:nuswide_dcs_ranking} in \cref{subsec:appendix_attack_defense_performance} for MNIST, CIFAR10 and NUSWIDE dataset respectively, with calculation detail of each column provided in \cref{subsec:appendix_cdcs_calculation_detail}.%Consequently, our experimental evaluations do not include assessments of GPer, CAE, and DCAE against FR attacks and NTB attacks, while GS is excluded from evaluations involving FR attacks.}}
%{\color{red}{Similarly, we do not force attacks onto settings that are inappropriate according to their original settings meaning that NS attack is only applied on unbalanced NUSWIDE dataset and GRN attack is not applied on NUSWIDE dataset for feature reconstruction at the passive party, as the passive dataset exclusively consists of binary-valued features and takes up more than $60\%$ of the sum of features obtained at both parties.}}

% It's obvious that TBM is a quite strong attack against which no defense shows a good capability. This is because TBM can query the whole model using auxiliary data that is i.i.d to the training dataset to train a generative model that back maps the VFL model output to the auxiliary data. This white-box attack with prior knowledge of data distribution can easily inverse the whole VFL model at a high level. For other attacks, we can see that MID, followed by DCAE, consistently achieves a defense point $(\text{MP}_i^{\text{MID}},\text{AP}_i^{\text{MID}})$ close to ideal defense performance; CAE performs well under attacks that does not obtain auxiliary labeled data; dCor and GPer are each effective for some kind of attack but far from satisfactory in others like LRB; G-DP and L-DP each displays a clear feature of harming the MP for lower AP; GS is not a promising defense for some attacks especially those that exploit the sign feature of gradients like DLI. 

%\yang{results too messy. First discuss the trade-off.}\tianyuan{See bellow.}

%In each figure, a point to the lower-right corner indicates a better defense with a high MP and a low AP. 

\textbf{Attacks pose great threat to VFL.} Comparing the black squares illustrating the MP and AP of the attack against a VFL system without any defense in the sub-figures, we can observe that DS, DLI, BLI and TBM attacks are strong attacks with AP higher than $0.97$, while MF attacks are quite weak with AP below $0.1$. Other attacks, including NS (see \cref{fig:nuswide_MPAP} in \cref{subsec:appendix_attack_defense_performance}), PMC, AMC, GRN and LRB, also pose great threat to the VFL system.

\textbf{Defenses exhibit trade-offs between MP and AP.} For most of the attacks and defenses, we can observe an apparent trade-off between MP and AP, i.e. a lower AP is often gained with increasing harm of MP as defense strength grows, which can be controlled by adjusting defense hyper-parameters. 
An increase of noise level in DP-G and DP-L, sparsification rate in GS, regularization hyper-parameter $\alpha_d$ in dCor, confusional strength $\lambda_2$ in CAE and DCAE, regularization hyper-parameter $\lambda$ in MID or a decrease of DP budget $\epsilon$ in GPer will lead to lower MP and AP.
\textbf{DCS rankings are consistent across various datasets and settings.} As shown in \cref{tab:mnist_dcs_ranking,tab:nuswide_dcs_ranking,tab:cifar10_dcs_ranking}, the results of the C-DCS rankings are generally consistent across all $3$ datasets. %% namely MNIST, CIAFR10 and NUSWIDE, under the same aggVFL setting with FedSGD communication. Also, 
In addition, the C-DCS ranking of the defense methods are still generally consistent even when the VFL model partition and communication protocol changes, as shown in \cref{tab:mnist_dcs_ranking,tab:mnist_split_dcs_ranking,tab:mnist_fedbcd_dcs_ranking} in \cref{subsec:appendix_attack_defense_performance}.As summarised in \cref{fig:dcs_ranking_mean_std}, these results demonstrate the robustness of the proposed DCS metrics, as well as the stableness of relative performance of different defense methods.Note that, T-DCS$_{\text{FR}}$ values are much lower than the T-DCS of other types, indicating that FR attacks are harder to defend than other attacks, which are consistent with human observation (see \cref{fig:visualization_mnist_tbm} in \cref{subsec:visualization_tbm}). %and more effective defense methods against FR attacks are in need.

%{\color{red}{
%\textbf{Similar trend for each defense against each attack across datasets.} 
\textbf{MID, L-DP and G-DP are effective on a wide spectrum of attacks.} MID demonstrates its capability of achieving a relatively lower AP while maintaining a higher MP compared to most other defenses as shown in \cref{fig:mnist_MPAP,fig:nuswide_MPAP,fig:cifar10_MPAP} and  \cref{tab:mnist_dcs_ranking,tab:nuswide_dcs_ranking,tab:cifar10_dcs_ranking}; L-DP and G-DP are also generally effective under most attacks with above average T-DCS and C-DCS; DCAE is effective in defending against LI attacks; %\textbf{CAE is effective only against LI attacks that utilize merely the information of the current sample}%\textbf{CAE consistently performs well across when defending against LI attacks that utilize merely the information of the current sample} by disguising label directly; 
GS demonstrates strong defense ability for most of the LI attacks but performs less than satisfactory on LRB attacks; GPer performs similar to DP-G and DP-L in defending against label related attacks; %as it targets at defending LI attacks and guarantees label-DP \yang{what about other attac}ks?} \tianyuan{Not evaluated... Do we need to include the explanation?}
dCor is less effective in limiting AP under NTB attacks but is largely effective against PMC and AMC attacks as shown in \cref{fig:mnist_MPAP,fig:nuswide_MPAP,fig:cifar10_MPAP}. 
\begin{wrapfigure}{l}{0.55\textwidth}
% \vspace{-1em}
% \captionsetup{font={scriptsize}}
% \begin{figure}[htbp]             
  \centering
  \includegraphics[width=0.99\linewidth]{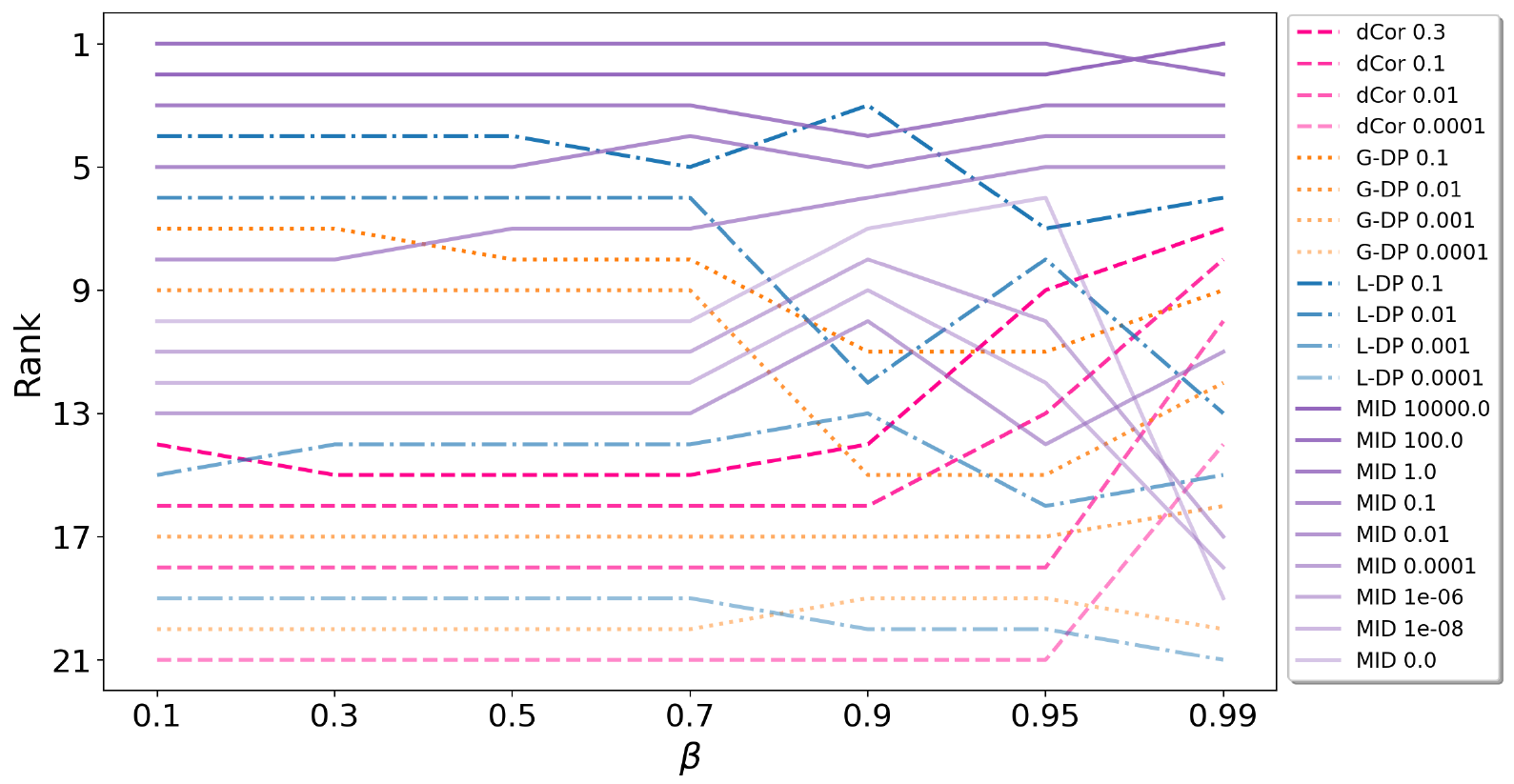}
  \vspace{-2em}
  \caption{Change of C-DCS ranking with the change of $\beta$. [MNIST dataset, aggVFL, FedSGD]}
  \label{fig:change_beta_dcs_mnist}
% \end{figure}
\vspace{-1em} %-1.5em
\end{wrapfigure}
\textbf{Change in $\beta$ does not significantly impact the C-DCS ranking.} $\beta$ in \cref{eq:dcs} represents users' trade-off preference on AP and MP when evaluating defenses, and can be adjusted. Here we use $\beta =0.5$ for our main results. %If the user wants to place more attention on maintaining a high MP, hyper-parameter $\beta$ in \cref{eq:dcs} can be adjusted to a value larger than $0.5$. 
%\yang{it would be interesting to add analysis on the impact of $\beta$ to the ranking.} 
\cref{fig:change_beta_dcs_mnist,fig:change_beta_dcs_cifar10,fig:change_beta_dcs_nuswide} show the change of the ranking results with the change of $\beta$. Overall the relative rankings are not significantly impacted by $\beta$, demonstrating the stableness of the comparison results among various defenses. As $\beta$ grows to large values, e.g. $\geq 0.9$, the metric places overly strong weight on MP, resulting in more variations on the rankings. Specifically, dCor ranks higher with the increase of $\beta$ thanks to its better MP preservation at the cost of a weaker AP limitation.

% \yang{this table does not include any results on CAE or DCAE and others. Should list them as well.For each T-DCS, it would be expected that certain other specific defense would perform the best.} \tianyuan{We have their results, but how should them be ranked?}
% While L-DP and dCor performs better in Targeted Backdoor attack (TB) and Non-Targeted Backdoor attacks (NTB) separately.

\begin{table}[!tb]
% \scriptsize    
\vspace{-1em}
\caption{T-DCS and C-DCS for All Defenses [NUSWIDE dataset, aggVFL, FedSGD]}
\label{tab:tiny_nuswide_dcs_ranking}
\resizebox{0.998\linewidth}{!}{
  \centering
   \begin{tabular}{cc|cccccc|c}
    \toprule
    \textbf{\makecell{Defense\\Name}} & \textbf{\makecell{Defense\\Parameter}} & \bm{$T\text{-}DCS_{LI_{2}}$} &
    \bm{$T\text{-}DCS_{LI_{5}}$} & \bm{$T\text{-}DCS_{LI}$} & \bm{$T\text{-}DCS_{FR}$} & \bm{$T\text{-}DCS_{TB}$} & \bm{$T\text{-}DCS_{NTB}$} & \bm{$C\text{-}DCS$} \\ 
    \midrule 
    MID  & 10000  & 0.7358 & 0.8559 & \textbf{0.8159} & 0.5833 & \textbf{0.7333} & 0.8707 & 0.7508  \\
    MID  & 1.0    & 0.7476 & 0.8472 & 0.8140 & 0.5833 & 0.7331 & 0.8700 & 0.7501  \\
    MID  & 100    & 0.7320 & 0.8536 & 0.8130 & 0.5833 & 0.7326 & \textbf{0.8711} & 0.7500  \\
    G-DP & 0.1    & 0.7375 & 0.8262 & 0.7966 & 0.5863 & 0.7282 & 0.8675 & 0.7447  \\
    L-DP & 0.1    & 0.7389 & 0.8177 & 0.7915 & 0.5863 & 0.7258 & 0.8603 & 0.7410  \\
    MID  & 0.1    & 0.7516 & 0.8259 & 0.8011 & 0.5833 & 0.7172 & 0.8563 & 0.7395  \\
    MID  & 0.01   & 0.7280 & 0.8092 & 0.7822 & 0.5844 & 0.7151 & 0.8627 & 0.7361  \\
    % MID  & 0.0001 & 0.7144 & 0.8097 & 0.7779 & 0.5856 & 0.7040 & 0.8680 & 0.7339  \\
    dCor & 0.3    & \textbf{0.7641} & 0.8411 & 0.8155 & 0.5834 & 0.7289 & 0.8051 & 0.7332  \\
    % G-DP & 0.01   & 0.7391 & 0.7600 & 0.7530 & 0.5863 & 0.7061 & 0.8549 & 0.7251 & 10 \\
    % L-DP & 0.01   & 0.7395 & 0.7525 & 0.7482 & 0.5863 & 0.7148 & 0.8485 & 0.7244 & 11 \\
    % MID  & 1e-06  & 0.7022 & 0.8201 & 0.7808 & 0.5860 & 0.6880 & 0.8408 & 0.7239 & 12 \\
    % dCor & 0.1    & 0.7442 & 0.7617 & 0.7559 & 0.5841 & 0.7259 & 0.8279 & 0.7234 & 13 \\
    % MID  & 1e-08  & 0.7066 & 0.8147 & 0.7787 & 0.5862 & 0.6593 & 0.8410 & 0.7163 & 14 \\
    % MID  & 0.0    & 0.6599 & 0.8097 & 0.7598 & 0.5862 & 0.6590 & 0.8414 & 0.7116 & 15 \\
    % L-DP & 0.001  & 0.7291 & 0.7234 & 0.7253 & 0.5863 & 0.6424 & 0.8329 & 0.6967 & 16 \\
    % G-DP & 0.001  & 0.7175 & 0.7237 & 0.7216 & 0.5863 & 0.6379 & 0.8334 & 0.6948 & 17 \\
    % dCor & 0.01   & 0.7445 & 0.7021 & 0.7162 & 0.5863 & 0.6336 & 0.8295 & 0.6914 & 18 \\
    % L-DP & 0.0001 & 0.6783 & 0.6470 & 0.6574 & 0.5863 & 0.6313 & 0.8293 & 0.6761 & 19 \\
    % G-DP & 0.0001 & 0.6495 & 0.6381 & 0.6419 & 0.5863 & 0.6309 & 0.8290 & 0.6720 & 20 \\
    % $\cdots$ & $\cdots$ & $\cdots$ & $\cdots$ & $\cdots$ & $\cdots$ & $\cdots$ & $\cdots$ & $\cdots$ \\
    dCor & 0.0001 & 0.6496 & 0.6340 & 0.6392 & \textbf{0.5864} & 0.6307 & 0.8287 & 0.6712 \\
    % None & None   & 0.6455 & 0.6339 & 0.6377 & 0.5863 & 0.6318 & 0.8276 & 0.6709 & 22 \\
    \hline
    \\[-1em]
    % GS   & 99.5   & 0.7381 & 0.8142 & 0.7888 & -      & 0.6456 & 0.8415 & -      & -  \\
    GS   & 99.0   & 0.7404 & 0.8060 & 0.7841 & -      & 0.6415 & 0.8408 & -      \\
    % GS   & 97.0   & 0.7414 & 0.7672 & 0.7586 & -      & 0.6376 & 0.8392 & -      & -  \\
    % GS   & 95.0   & 0.7423 & 0.7399 & 0.7407 & -      & 0.6375 & 0.8385 & -      & -  \\
    CAE  & 1.0    & 0.6863 & 0.7822 & 0.7502 & -      & 0.6830  & -           \\
    % CAE  & 0.5    & 0.6808 & 0.7848 & 0.7501 & -      & 0.6733  & -      & -      \\
    % CAE  & 0.1    & 0.6808 & 0.8249 & 0.7768 & -      & 0.6734  & -      & -      \\
    % CAE  & 0.0    & 0.6808 & 0.8212 & 0.7744 & -      & 0.6807  & -      & -      \\
    % DCAE & 1.0    & 0.6716 & 0.8156 & 0.7676 & -      & 0.6771 & -      & -      & -  \\
    % DCAE & 0.5    & 0.6672 & 0.8108 & 0.7629 & -      & 0.6668 & -      & -      & -  \\
    % DCAE & 0.1    & 0.6669 & 0.8651 & 0.7991 & -      & 0.6746 & -      & -      & -  \\
    DCAE & 0.0    & 0.6669 & \textbf{0.8660} & 0.7996 & -      & 0.6816 & -      & -       \\
    % GPer & 10.0   & 0.6877 & 0.6722 & 0.6773 & -      & 0.6222 & -      & -      & -  \\
    % GPer & 1.0    & 0.7230 & 0.7460 & 0.7383 & -      & 0.6315 & -      & -      & -  \\
    % GPer & 0.1    & 0.7395 & 0.8007 & 0.7803 & -      & 0.7042 & -      & -      & -  \\
    GPer & 0.01   & 0.7386 & 0.8412 & 0.8070 & -      & 0.7193 & -      & -       \\
    \bottomrule
    \end{tabular}
}
\vspace{-1em}
\end{table}

%\subsubsection{splitVFL and aggVFL Comparison} \label{subsec:splitVFL_comparison}
\textbf{splitVFL is less vulnerable to attacks than aggVFL.}
% \begin{wrapfigure}{l}{0.36\textwidth}
% \vspace{-2em}
% % \begin{figure}[H]
%     \centering
%     \includegraphics[width=0.99\linewidth]{figures/split_agg_compare/splitagg_hist.png}
%     \vspace{-2em}
%     \caption{DCS gap Distribution, y-axis represents density [MNIST dataset, FedSGD, aggVFL/splitVFL]\tianyuan{@Zixuan, please adjust the x-axis and y-axis's name and font size; please update this figure, including the sub-figures' order}}
%     \label{fig:mnist_splitVFL_DCS_hist}
% % \end{figure}
% \vspace{-2em}
% \end{wrapfigure}
Using DCS metrics, we directly compare all the aforementioned attacks and defenses under aggVFL and splitVFL settings to understand the impact of changing the model partition strategy on VFL's vulnerability against attacks. We mainly use the DCS gap, defined as $\text{DCS}^{\text{splitVFL}}-\text{DCS}^{\text{aggVFL}}$ for each attack-defense point. \cref{fig:mnist_splitVFL_DCS_gap,fig:nuswide_split_DCS_gap} \begin{wrapfigure}{l}{0.55\textwidth}
% \vspace{-1em}
% \begin{figure}[htbp]             
% \captionsetup{font={scriptsize}}
  \centering
  \includegraphics[width=0.99\linewidth]{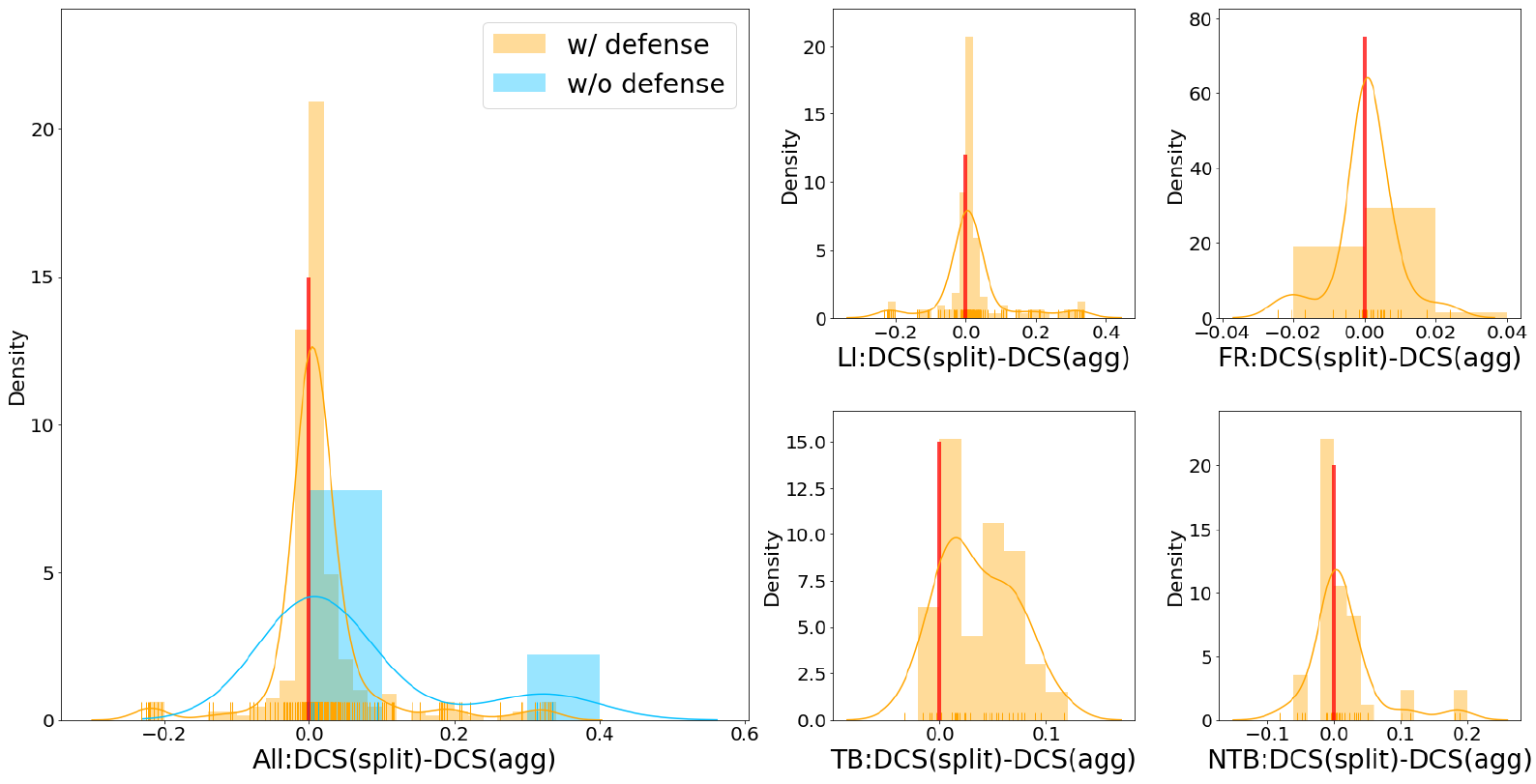}
  \vspace{-1em}
  \caption{DCS gap Distribution, y-axis represents density [MNIST dataset, splitVFL/aggVFL, FedSGD]}   
  \label{fig:mnist_splitVFL_DCS_hist}           
% \end{figure}
\vspace{-0.5em}
\end{wrapfigure}
(see \cref{subsec:appendix_splitVFL_comparison}) show the DCS gap for all the attack-defense points among the $11$ attacks and $8$ defenses with all the evaluated parameters using MNIST and NUSWIDE dataset respectively. 
\cref{fig:mnist_splitVFL_DCS_hist,fig:nuswide_split_hist} displays the distribution of the DCS gaps depicted in \cref{fig:mnist_splitVFL_DCS_gap,fig:nuswide_split_DCS_gap} respectively.
As all the black square points in \cref{fig:mnist_splitVFL_DCS_gap} appear above or close to the red horizontal line at a value of $0.0$ (see also the blue histograms that appear mostly at the right of the vertical line at a value of $0.0$ in \cref{fig:mnist_splitVFL_DCS_hist}), we can conclude that splitVFL is less vulnerable to attacks than aggVFL when no defense is applied. In addition, splitVFL has an overall positive effect on boosting defense performance against attacks as well, as most of the DCS gap is positive in the last subplot of \cref{fig:mnist_splitVFL_DCS_gap} when no attack is applied. Similar results can be seen from \cref{fig:nuswide_split_hist,fig:nuswide_split_DCS_gap} in \cref{subsec:appendix_splitVFL_comparison}. Further analysis are included in \cref{subsec:appendix_splitVFL_comparison}. 

\begin{wrapfigure}{l}{0.55\textwidth}
\vspace{-1em}
% \captionsetup{font={scriptsize}}
% \begin{figure}[htbp]             
      \centering
      \includegraphics[width=0.99\linewidth]{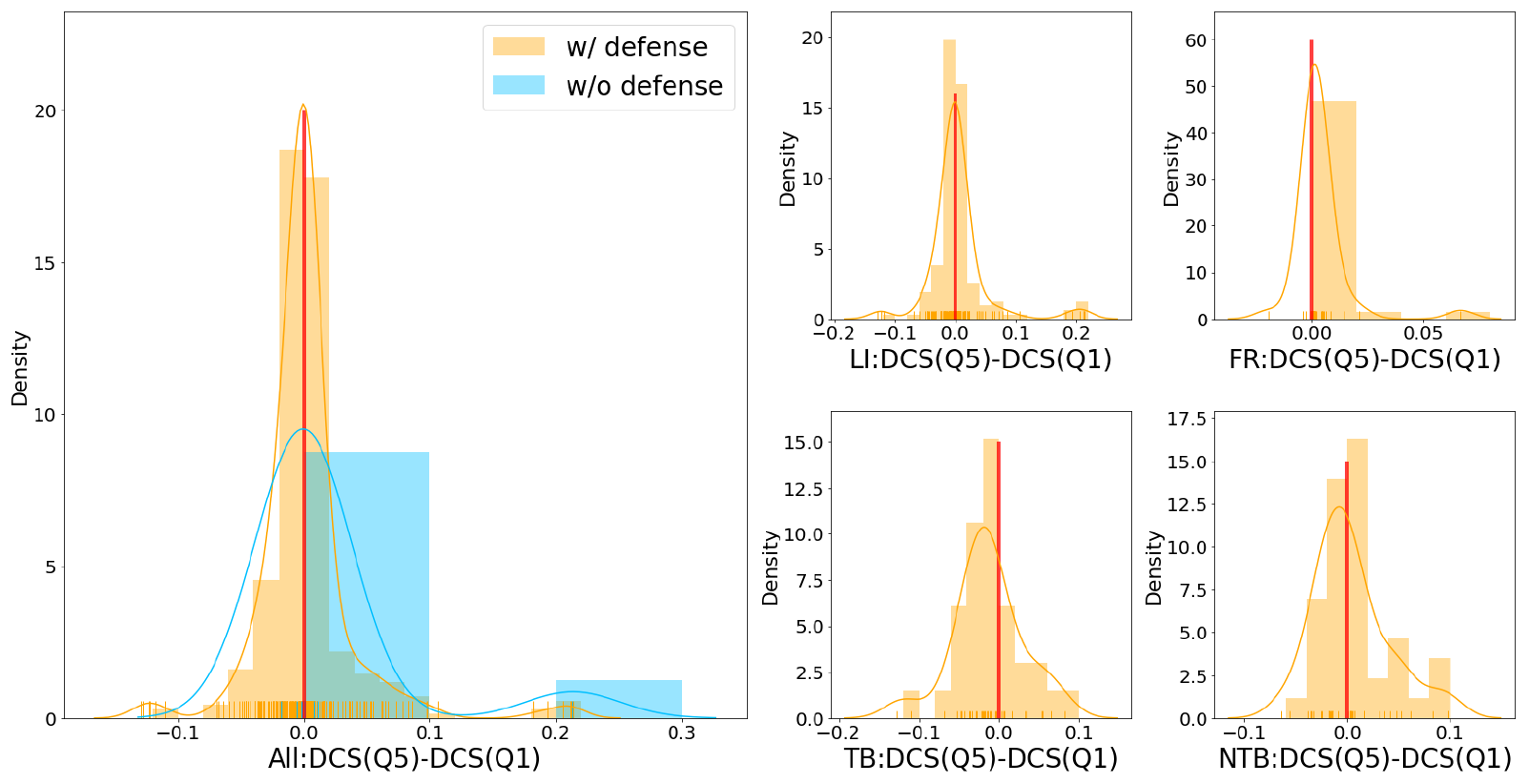}
      \vspace{-1em}
    \caption{DCS gap Distribution, y-axis represents density [MNIST dataset, aggVFL, FedBCD/FedSGD]}
  \label{fig:mnist_FedBCD_DCS_hist}           
% \end{figure}
% \vspace{-1em}
\end{wrapfigure}

\textbf{FedBCD is less vulnerable to attacks than FedSGD.} In addition, we compare DCS gap under FedSGD setting and FedBCD with $Q=5$ to assess the impact of different communication protocols on model's vulnerability to attacks. DCS gap is defined as $\text{DCS}^{\text{FedBCD}}-\text{DCS}^{\text{FedSGD}}$ for each attack-defense point. 
\cref{fig:mnist_FedBCD_DCS_hist,fig:nuswide_FedBCD_hist} provides the distribution of the DCS gaps plotted in \cref{fig:mnist_FedBCD_DCS_gap,fig:nuswide_FedBCD_DCS_gap} (see \cref{subsec:appendix_FedBCD_comparison}) which includes the DCS gap of all the attack-defense points among the $11$ attacks and the $8$ defenses using MNIST and NUSWIDE dataset respectively.
% evaluated in this work as listed in \cref{tab:defense_param}. % The y-axis represents the DCS gap value while the x-axis represents different defenses.
As shown in \cref{fig:mnist_FedBCD_DCS_hist}, the blue histograms generally appear on the right of the vertical line of value $0.0$, indicating that a system with FedBCD protocol is less vulnerable to attacks when no defense method is applied. In addition, a system with FedBCD also has an overall positive effect on boosting defense performance against FR and NTB attacks. This is evidenced by the fact that that the majority of DCS gaps are positive for FR and NTB attacks as shown in \cref{fig:mnist_FedBCD_DCS_hist}.
% Notably, \cref{fig:mnist_FedBCD_DCS_hist} shows that %appear mostly at the right side of the vertical line at a
% %value of $0.0$, indicating that 
% \textit{a system is less vulnerable to attacks under FedBCD setting when no defense method is applied}. 
% When attacks exist, it's apparent from \cref{fig:mnist_FedBCD_DCS_hist} that the majority of DCS gaps still obtain positive values, especially for LI and FR attacks, indicating that \textit{FedBCD has an overall positive effect on boosting defense performance against applied attacks}. 
Similar conclusions can be drawn from \cref{fig:nuswide_FedBCD_hist,fig:nuswide_FedBCD_DCS_gap}.
Further analysis are included in \cref{subsec:appendix_FedBCD_comparison}.

\section{Conclusions and Limitations} % and Societal Impacts
%\tianyuan{Mrs. Liu, limitations and societal impacts are encouraged to be included in NeurIPS submissions as given in the checklist below.}

% \textbf{Conclusions.}  
In this work, we introduce a light-weight VFL framework \verb|VFLAIR| that implements basic VFL training and evaluation flow under multiple model partition, model architectures,communication protocols and attacks and defenses algorithms using datasets of different modality. We also introduce unified evaluation metrics and benchmark model utility, communication and computation efficiency, and defense performance under various VFL settings, which sheds lights on choosing partition, communication and defense techniques in practical deployment. 
%We hope that our work will facilitate future VFL research and benchmark efforts. %We also hope to provide insights into existing attack and defense works and give useful recommendation for protection technique selection in real-world industrial applications.
% In this work, we perform a comprehensive benchmark on existing attacks and defense methods in various VFL settings. We perform extensive experiments with our VFL framework \verb|VFLAIR| under multiple model partition and communication protocols using datasets of different modality. By developing new evaluation metrics with both defense depth and breadth considered, we propose concrete recommendations for selecting defense strategies under different practical VFL deployment scenarios. We hope that our evaluation benchmarks and framework can help expedite the development of VFL evaluations for both future researchers and industrial developers %quickly select an appropriate defense methods for application.} \tianyuan{Mrs. Liu, In backdoorbench, this part appears in the introduction...}
% \textbf{Limitations.} %Although multi-party VFL scenario is supported by \verb|VFLAIR| and can be tested easily with our pipeline, our benchmark evaluation is currently limited to VFL systems with $2$ participants in this work. %We hope to extend to multi-party defense evaluation for future work.
Currently, the library has limited implementations on cryptographic techniques. Combination of non-cryptograhic and cryptographic techniques would be an interesting next step and we plan to add more advanced privacy-preserving and communication-efficient methods to our library. %encryption implementation, since is not implemented for NN-based VFL, we will add this part in the near future.}}

\section{Reproducibility Statement}
We include all the hyper-parameters used for the benchmark experiments in \cref{sec:appendix_experimental_settings}. Our code is also available at
\url{https://github.com/FLAIR-THU/VFLAIR}).
% \url{https://anonymous.4open.science/r/VFLAIR-1ABD}.

% Reference
% {
% \small
\section*{Acknowledgement}
This work was supported by the National Key R\&D Program of China under Grant No.2022ZD0160504, the Tsinghua University(AIR)-Asiainfo Technologies (China) Inc. Joint Research Center, and Tsinghua-Toyota Joint Research Institute inter-disciplinary Program.

\bibliographystyle{iclr2024_conference}
% \bibliography{iclr2024_conference}
\bibliography{ref}
% }

%%%%%%%%%%%%%%%%%%%%%%%%%%%%%%%%%%%%%%%%%%%%%%%%%%%%%%%%%%%%
% \input{checklist}

% \newpage
\clearpage

\appendix
\label{appendix}

%%%%%%%%%%%%%%%%%%%%%%%%%%%%%%%%%%%%%%%%%%%%%%%%%%%%%%%%%%%%

\appendix

\section{Related Work} \label{sec:appendix_related_work}

\textbf{Vertical Federated Learning.} Federated Learning (FL)~\citep{mcmahan2016federated,yang2019federatedbook,yang2019federatedML} is a learning paradigm which allows multiple parties to build a machine learning model collaboratively without centralizing each data owner's local private data. Depending on whether data are partitioned by sample or by feature, FL can be further categorized into Horizontal FL (HFL) and Vertical FL (VFL)~\citep{yang2019federatedML}. VFL~\citep{liu2022fedbcd,cheng2021secureboost,jiang2022signds} is commonly applied in real-world applications in the field of finance and advertising~\citep{cheng2020federated,FATE} where cross-silo participants holding different sets of features of a common group of users jointly build machine learning models while keeping both local data and local model private. In VFL, only one participant possesses sensitive label information and is often referred to as \textit{active party} while others are referred to as \textit{passive parties}. %{\color{red}{
%Depending on whether a tree model or a neural network (NN) model is finally constructed in the VFL training process, VFL system can be categorized into tree-based VFL and NN-based VFL~\citep{liu2022vertical}. }}
%{\color{red}{
%\textbf{NN-based VFL.}
Depending on how model is partitioned among parties, the VFL architecture can be further categorized into aggVFL and splitVFL~\citep{liu2022vertical}. In aggVFL, each party possesses a local sub-model, and a non-trainable aggregation function is used as the global model; while in splitVFL, a trainable aggregation model is used. Communication efficiency issue is a key bottleneck in VFL training. Approaches such as FedBCD~\citep{liu2022fedbcd}, Compressed-VFL~\citep{castiglia2022compressed} (including Quantize and Top-k compression), Flex-VFL~\citep{Castiglia2022FlexibleVF}, AdaVFL~\citep{Jie2022adavfl}, and CELU-VFL~\citep{fu2022communicationefficient} enhance VFL system efficiency by enabling each party to perform multiple local updates during each communication iteration. 
Although raw data and model parameters are not shared in NN-based VFL training and inference procedure, the threats of data leakage and model integrity remain. Existing attacks target either the reconstruction of private data~\citep{li2022label,jiang2022comprehensive} or the compromise of model robustness~\citep{liu2020backdoor,zou2022defending,pang2023adi}. Private label or private features are both potential targets for data leakage attacks. Sample-level gradients (SLI)~\citep{li2022label,fu2021label,sun2021vertical,yang2022differentially} or batch-level gradients (BLI)~\citep{zou2022defending} or trained local models~\citep{fu2021label} can all be exploit to conduct label inference attacks. Model inversion technique~\citep{jin2021cafe,luo2021feature,li2022ressfl,jiang2022comprehensive} for white-box or black-box oracle setting with image or tabular data; or linear equation solving technique~\citep{ye2022feature,peng2022binary} for black-box setting with binary value data can be applied to conduct feature reconstruction attacks. Assigning specific label to triggered samples~\citep{zou2022defending} or adding noise to randomly selected samples~\citep{liu2021rvfr} or failing to transmit collaboration information~\citep{liu2021rvfr} can all result in successful backdoor attacks. These attacks pose significant challenges to VFL settings and necessitate effective countermeasures. To mitigate these threats, multiple defense methods have been proposed. Aside from cryptography techniques such as HE or Secure Multi-Party Computation (MPC)~\citep{yang2019federatedML}, non-cryptography techniques such as reducing information by Adding Noise~\citep{dwork2006DP,li2022label,fu2021label,zou2022defending}, Gradient Sparsification~\citep{aji2017sparse}, Gradient Discretization~\citep{dryden2016communication,fu2021label}, Gradient Compression~\citep{lin2018deep}, or the combination of these information reduction techniques~\citep{shokri2015privacy,fu2021label} are often applied. Additionally, emerging defense mechanisms have been proposed either for specific attacks or multiple type of attacks. Disguising labels~\citep{zou2022defending,jin2021cafe}, label DP~\citep{ghazi2021deep,yang2022differentially}, Dispersed Training~\citep{wang2023beyond} have proven effective in defending against label inference attacks; Dropout and Rounding~\citep{luo2021feature}, Adversarial Training~\citep{sun2021defending,li2022ressfl}, Fabricated Features~\citep{ye2022feature} are proposed to prevent input feature leakage; Robust Feature Recovery~\citep{liu2021rvfr} has demonstrated its effectiveness for eliminating backdoor attacks, while Mutual Information Regularization~\citep{zou2022defending} and Distance Correlation Regularization~\citep{vepakomma2019reducing,sun2022label} seek to mitigate various types of attacks.

\textbf{Tree-based VFL.} Due to its exceptional model performance and its inherent explainability, tree-based VFL has garnered widespread applications. Gradient Boosting Decision Tree (GBDT) is the most commonly employed approach for constructing trees-based VFL with various related algorithms having been proposed~\citep{cheng2021secureboost,chen2021secureboost+,wen2021securexgb,tian2020federboost,xie2022mpfedxgb}. Another noteworthy tree-based ensemble algorithm in VFL is Random Forest (RF)~\citep{tin1995rdf}
% , which has also been widely explored and integrated into VFL. 
which leverages bagging and parallelism optimization techniques to improve training and inference efficiency~\citep{liu2020federated,yao2022efficient}.
While most tree-based VFL methods employ cryptographic approaches to protect sensitive data, some encrypt only partial information to enhance efficiency~\citep{cheng2021secureboost,wu2020pivot} but results in data leakage threats~\citep{chamani2020mitigating,takahashi2023eliminating}. 
Thus, various methods have been investigated aiming to solve the privacy~\citep{cheng2021secureboost,chen2021secureboost+,feng2019securegbm,wen2021securexgb,tian2020federboost,li2022opboost} and efficiency~\citep{li2022opboost} issue.

\section{VFL Framework} \label{sec:appendix_vfl_framework}
We include the training algorithm of NN-based VFL (\cref{alg:NN_VFL_setting}) as well as the description and algorithm (\cref{alg:Tree_VFL_setting}) of the training and inference process of tree-based VFL in this section.
\begin{algorithm}[!htb]
%\SetAlgoLined
 \caption{A Basic VFL Training Procedure using FedSGD.}
 \textbf{Input}:learning rates $\eta_1$ and $\eta_2$\\
 \textbf{Output}: Model parameters $\theta_1$, $\theta_2$ ... $\theta_K$, $\varphi$
 \begin{algorithmic}[1]
 \STATE Party 1,2,\dots,$K$, initialize $\theta_1$, $\theta_2$, ... $\theta_K$, $\varphi$.
 \FOR{each iteration $j=1,2, ...$}
 \STATE Randomly sample a mini-batch of samples $\{\mathbf{x},\mathbf{y}\}\subset\mathcal{D}$ of size n;
 \FOR{each party $k$=1,2,\dots,$K$ in parallel}
 \STATE Party $k$ computes $H_{k} = G_k(\mathbf{x}_{k},\theta_k)$;
 \STATE Party $k$ sends $H_{k}$ to active party $K$;
 \ENDFOR
 \STATE Active party computes the prediction $\hat{\mathbf{y}} = F(H_1,\dots,H_K,\varphi)$ and loss $\mathcal{L}=\frac{1}{n}\ell(\mathbf{y},\hat{\mathbf{y}})$;
 \STATE Active party $K$ updates $\varphi^{j+1} = \varphi^{j} - \eta_1 \frac{\partial \ell}{\partial \varphi}$;
 \STATE Active party $K$ computes and sends $\frac{\partial \mathcal{L}}{\partial H_{k}}$ to all other parties;
 \FOR{each party $k$=1,2,\dots,$K$ in parallel}
 \STATE Party $k$ computes $\nabla_{\theta_k} \mathcal{L} = \frac{\partial \mathcal{L}}{\partial \theta_k}=\frac{\partial \mathcal{L}}{\partial H_k}\frac{\partial H_k}{\partial \theta_k}$;
 \STATE Party $k$ updates $\theta^{j+1}_k = \theta^{j}_k - \eta_2 \nabla_{\theta_k} \mathcal{L}$;
 \ENDFOR
 \ENDFOR
 \end{algorithmic}
 \label{alg:NN_VFL_setting}
\end{algorithm}

We also support using Homomorphic Encryption (HE) with Paillier Encryption to protect transmitted results during training VFL %for linear regression with logit loss using Paillier Encryption, 
~\cite{zou2022defending} (see \cref{alg:he-vertical_federated_learning}). In this protocol, Homomorphic Encryption (HE), denoted as $[[·]]$, is applied to the communicated information, i.e., local model outputs ${H_{k}}$ in forward propagation and respective gradients $\frac{\partial \ell}{\partial H_{k}}$ in backward propagation. %That is, each passive party transmits ency $[[{H_{k}}]]$ to the active party. 
The active party then computes the gradient of loss with respect to $[[H_{k}]]$ under encryption, i.e. $[[\frac{\partial \mathcal{L}}{\partial H_{k}}]]$, and sends the results back to the passive parties for gradient updates. Note Taylor Expansion is used for gradient approximation as Paillier Encryption supports only addition and multiplication, following~\citep{liu2018ftl}. The computed encrypted gradients $[[\frac{\partial \mathcal{L}}{\partial H_{k}}]]$ are subsequently added with a random local mask and transmitted to a Trusted Third Party (TTP) for decryption. The public key for the encryption is generated and distributed to each party by TTP, while the paired private keys are kept at TTP for decryption.

\begin{algorithm}
\caption{A vertical federated learning framework with Homomorphic Encryption~\citep{zou2022defending}}
\label{alg:he-vertical_federated_learning}
\begin{algorithmic}[1]
\REQUIRE Learning rate $\eta$
\ENSURE Model parameters $\theta_1, \theta_2, \ldots, \theta_K$
\STATE Party $1, 2, \ldots, K$ initializes $\theta_1, \theta_2, \ldots, \theta_K$
\STATE Trusted Third Party (TTP) creates encryption pairs, sends public key to each party
\FOR{$j = 1, 2, \ldots$}
    %\STATE Randomly sample $S \subset [N]$
    \FOR{each passive party $k \neq K$ in parallel}
        \STATE $k$ computes, encrypts, and sends $[[H_k]]$ to the active party.
    \ENDFOR
    \STATE Active party $K$ computes and sends $[[\frac{\partial \mathcal{L}}{\partial H_k}]]$ to all other parties
    \FOR{each party $k = 1, 2, \ldots, K$ in parallel}
        \STATE $k$ computes $[[\frac{\partial \mathcal{L}}{\partial \theta_k}]]$ and sends them with a random mask to TTP for decryption.
        \STATE $k$ receives and unmasks $\nabla_{\theta_k} \mathcal{L} \leftarrow \frac{\partial \mathcal{L}}{\partial \theta_k}$ and updates $\theta^{j+1}_k = \theta^j_k - \eta \nabla_{\theta_k} \mathcal{L}$
    \ENDFOR
\ENDFOR
\end{algorithmic}
\end{algorithm}

In tree-based VFL, the active party first broadcasts the set of record indices for the current node. Next, each passive party calculates the percentiles for each feature based on those indices. The passive party then proceeds to create binary splits for each feature by comparing the feature value of each instance to the percentile values. After that, the passive party sends back the statistics of each split necessary for evaluation such as purity to the active party, and the active party selects the best split using a specific evaluation function. To be precise, Random Forest employs gini impurity for classification, while XGBoost utilizes its gain function, which is based on the gradient and hessian. Finally, the active party requests the owner of the best split to send the set of record indices for the children nodes generated by the best split. Tree-based VFL system continues these procedures recursively until certain stop conditions, like depth constraints, are satisfied. \cref{alg:Tree_VFL_setting}
% (in the appendix) 
demonstrates training details of tree-based VFL.

\begin{algorithm}[!htb]
%\SetAlgoLined
 \caption{A Basic Training Process of Tree-based VFL.}
 \textbf{Input}: Evaluation function for a split\\
 \textbf{Output} Trained trees: 
 \begin{algorithmic}[1]

\FOR{each tree $j=1,2, ...$}
    \STATE Root Node $\leftarrow$ randomly sample a subset of record indices;
    \STATE Nodes = [Root Node] 
    \COMMENT{A list of nodes to be divided}
    \WHILE{Nodes is not empty}
    \STATE Current node $\leftarrow$ pop one element of Nodes
    \IF{Current node satisfies the terminate conditions} \STATE Continue \ENDIF
    \STATE Active party broadcasts the set of record indices of the current node to all other parties;
     \FOR{each party $k$=1,2,\dots,$K$ in parallel}
    \STATE Party $k$ receives the set of record indices of a node to be divided;
    \STATE Party $k$ calculates the statistics for all possible splits and sends them to the active party;
    \ENDFOR
    \STATE Active party gathers the statistics of possible splits and picks the best one;
    
    \IF{Party $k$ is selected as the best party}
    \STATE Party $k$ sends the sets of record indices of children nodes generated by the best split;
    \ENDIF
    \STATE Active party receives the sets of record indices of children nodes and  appends them to Node;
    \ENDWHILE
\ENDFOR

\end{algorithmic}
 \label{alg:Tree_VFL_setting}
\end{algorithm}

\section{Quick Guide to Use and Extend VFLAIR} \label{section:appendix_user_guidance}

% {\color{red}{
In this section, we give a step-by-step user guidance on how to use and extend \verb|VFLAIR|.
\paragraph{How to Use.}
\begin{figure}[!htb]
    \centering
    \subfigure[Code download and environment preparation.]{
        \begin{minipage}[t]{0.75\linewidth}
        \centering
        \includegraphics[width=1\linewidth]{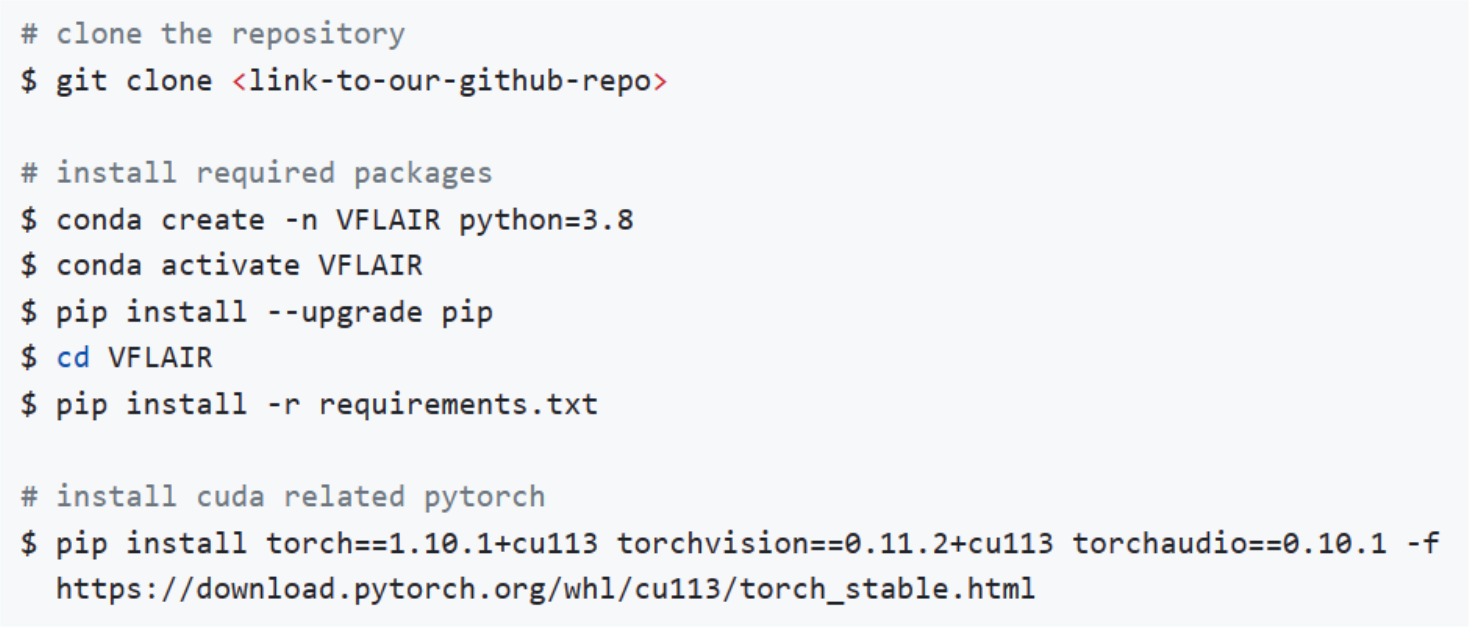}
        \label{fig:user_guidance_code_environment}
        \end{minipage}
    }% 
    
    \subfigure[Example JSON file.]{
        \begin{minipage}[t]{0.94\linewidth}
        \centering
        \includegraphics[width=1\linewidth]{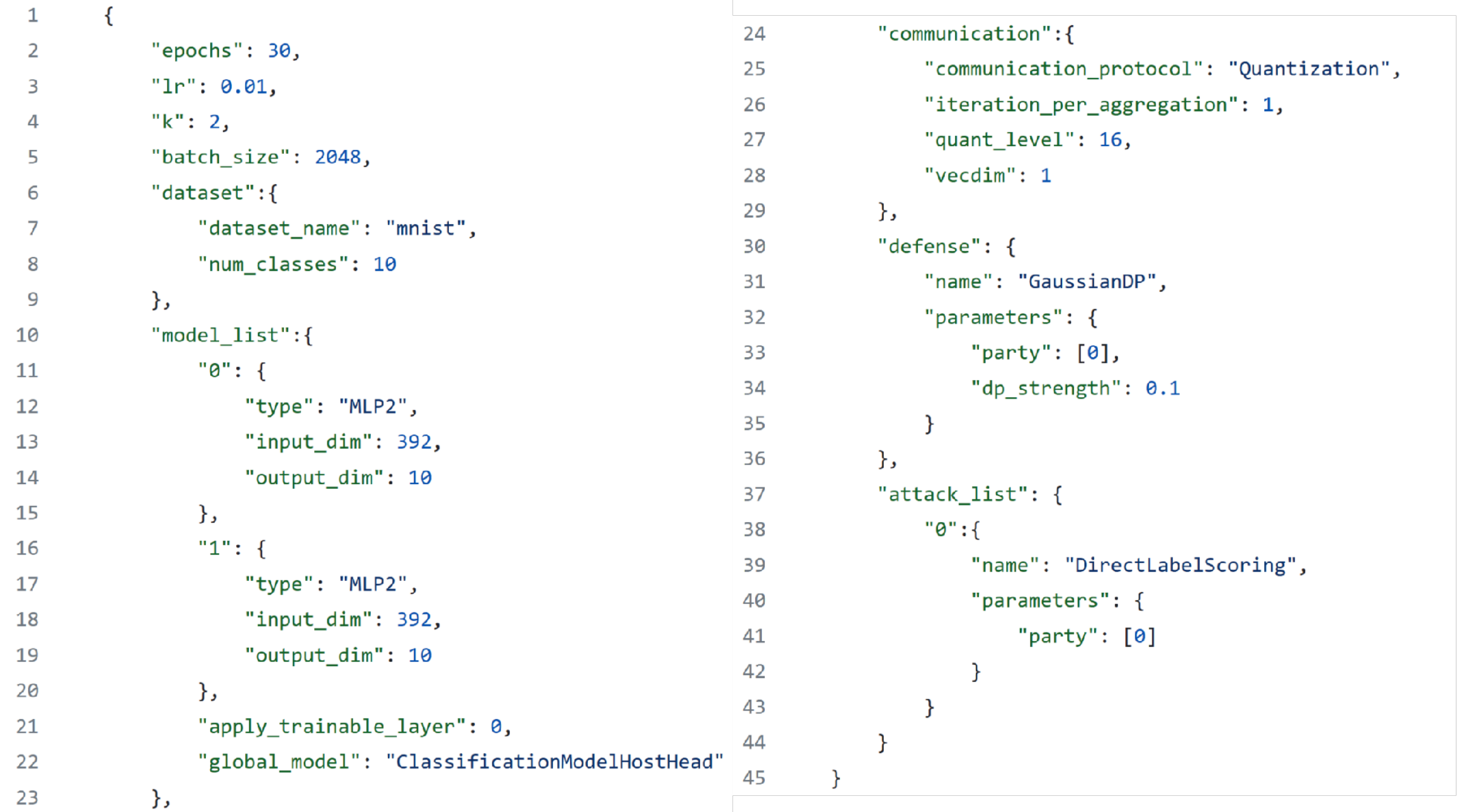}
        \label{fig:user_guidance_config_file}
        \end{minipage}
    }% 
    \vspace{-1em}
    \caption{User guidance.}
\end{figure}

\begin{figure}[!htb]
    \centering
    \includegraphics[width=0.89\linewidth]{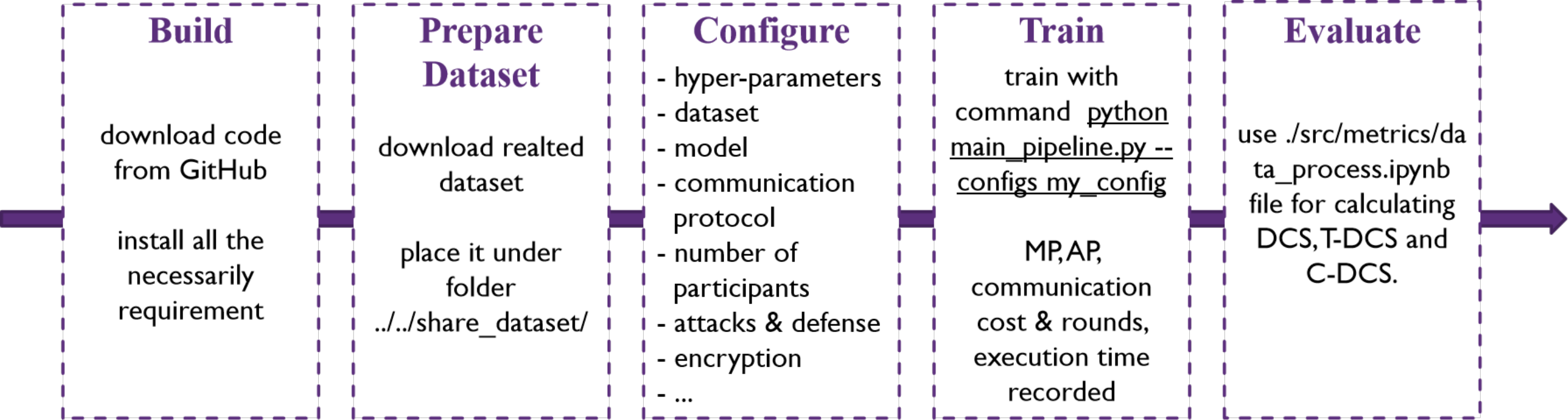}
    % \vspace{-1em}
\caption{Step-by-step user guidance for using VFLAIR.}
\label{fig:user_guidance_step_by_step}
% \vspace{-1em}
\end{figure}

Using \verb|VFLAIR| requires the following steps:
\begin{enumerate}
    \item \textbf{Build.} Download our code repository from GitHub and prepare all the required environments using commands shown in \cref{fig:user_guidance_code_environment}. Hardware requirements are listed in \cref{tab:hardware} in \cref{section:appendix_vfl_workflow}.
    \item \textbf{Prepare Dataset.}  Prepare or download the dataset into folder \verb|../../share_dataset/|, the default folder that contains all the datasets for experiments. 
    \item \textbf{Configure.} Modify the configuration JSON file under \verb|./src/configs/| folder to specify settings, including learning hyper-parameters (e.g. 'epochs' and 'lr' for learning rate etc.), training dataset ('dataset'), training model ('model\_list' for model of each party), model partition ('global\_model' trainable or not), communication protocol ('communication') as well as attacks ('attack\_list') and defense method ('defense'). Then rename (e.g. \verb|my_config.json|) the configuration file and save it under \verb|./src/configs/| folder. An example is shown in \cref{fig:user_guidance_config_file}. Detailed explanations of all the parameters are provided in \verb|./src/configs/README.md| for NN-based VFL and \verb|./src/configs/README_TREE.md| for Tree-based VFL.
    \item \textbf{Train.} Use command \verb|cd src| and \verb|python main_pipeline.py --configs| \verb|my_config| consecutively to start a VFL training. Attack and defense evaluation will also be performed if specified in configuration file. MP, AP, communication rounds, amount of information transmitted each communication round as well as execution time for reaching the specified MP is recorded. % Or by using \verb|python main_paillier.py --configs my_config| to apply encryption to the training and evaluation flow. 
    \item \textbf{Evaluate.} After training is finished, use \verb|./src/metrics/data_process.ipynb| file to perform evaluations, e.g. calculate DCS, T-DCS, C-DCS.
\end{enumerate}
We summarize the above steps in \cref{fig:user_guidance_step_by_step}.

\paragraph{How to Extend.}
Extending our \verb|VFLAIR| platform is also easy thanks to our modularized design of all the relative components.
\begin{itemize}
    \item New hyper-parameter configurations (e.g. new training optimizer specification) can be added by modifying functions in \verb|./src/load/LoadConfigs.py| file. 
    \item New datasets and data partitioning methods can be added by first placing the raw data file under \verb|../../share_dataset/| folder and then modifying functions in \verb|./src/load/LoadDataset.py| file.
    \item New VFL models can be added under \verb|./src/models/| folder and loaded via modifying the \verb|./src/load/LoadModels.py| file with proper personalized modification in the 'model\_list' part (see line 10-23 in \cref{fig:user_guidance_config_file}) of the configuration file.
    \item New communication protocols can be added by modifying the provided training flow in \verb|./src/evaluates/MainTaskVFL.py| file to realize of the new communication protocol or by adding a new python file under the same directory of that file if the user finds the modification significant.
    \item New attacks can be added by inheriting \verb|Attacker| class under \verb|./src/evaluates/attacks/| folder for testing time attacks, or by modifying the main training file \verb|./src/evaluates/MainTaskVFL.py| for training time attack.
    \item New defense methods can be added by modifying \verb|./src/evaluates/defenses/defense_functions.py| file and the main training file \verb|./src/evaluates/MainTaskVFL.py| if necessary.
    \item New party behaviors, e.g. new information exchange strategies for new communication protocols, can be added by modifying \verb|./src/party/party.py|, under \verb|./src/party/| folder.
\end{itemize}

\begin{figure}[!htb]
    \centering
    \subfigure[New JSON file.]{
        \begin{minipage}[t]{0.94\linewidth}
        \centering
        \includegraphics[width=1\linewidth]{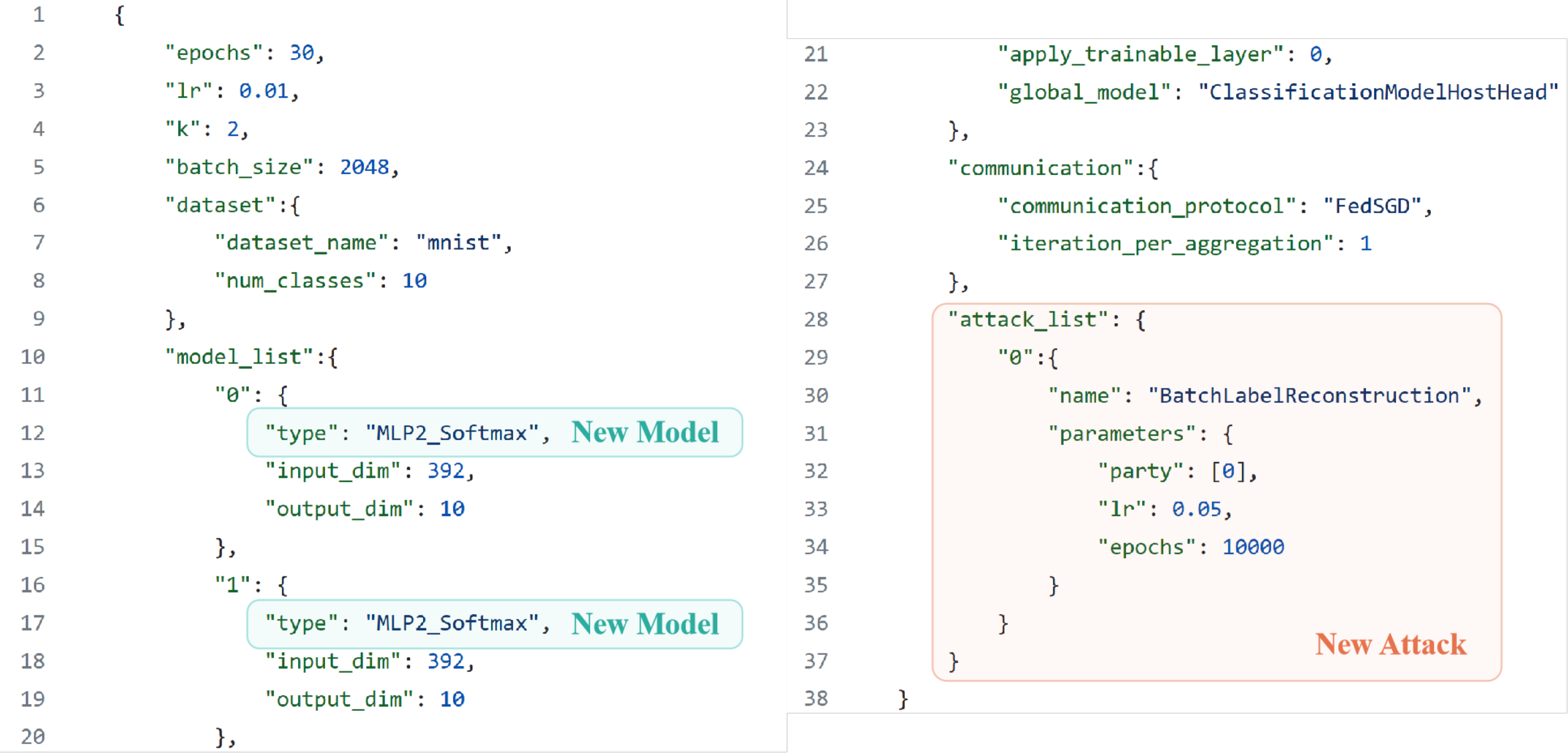}
        \label{fig:user_guidance_new_config_file}
        \end{minipage}%
    }% 
    
    \subfigure[New attack.]{
        \begin{minipage}[t]{0.52\linewidth}
        \centering
        \includegraphics[width=1\linewidth]{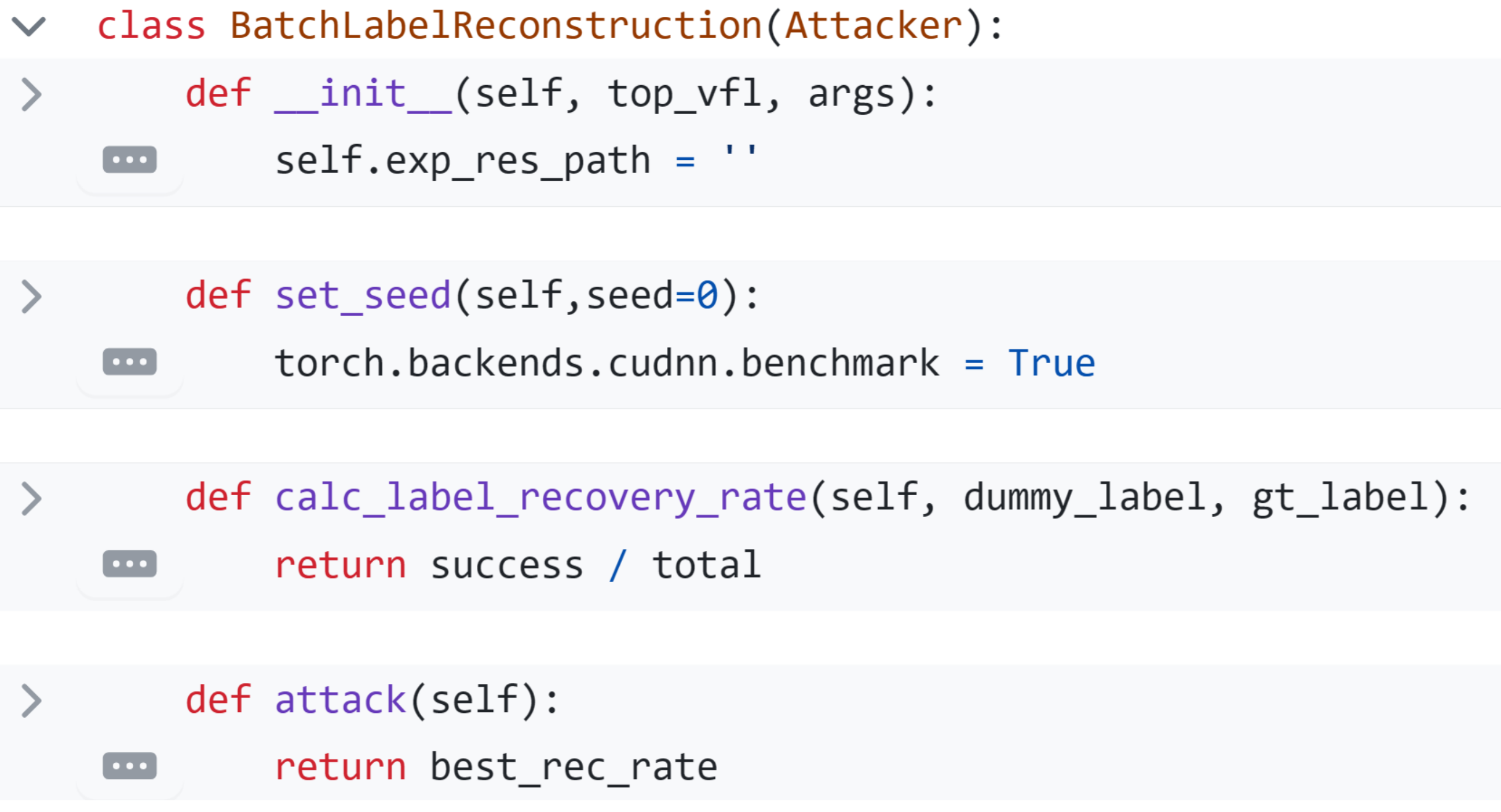}
        \label{fig:user_guidance_new_attack}
        \end{minipage}%
    }% 
    \subfigure[New bottom model.]{
        \begin{minipage}[t]{0.47\linewidth}
        \centering
        \includegraphics[width=1\linewidth]{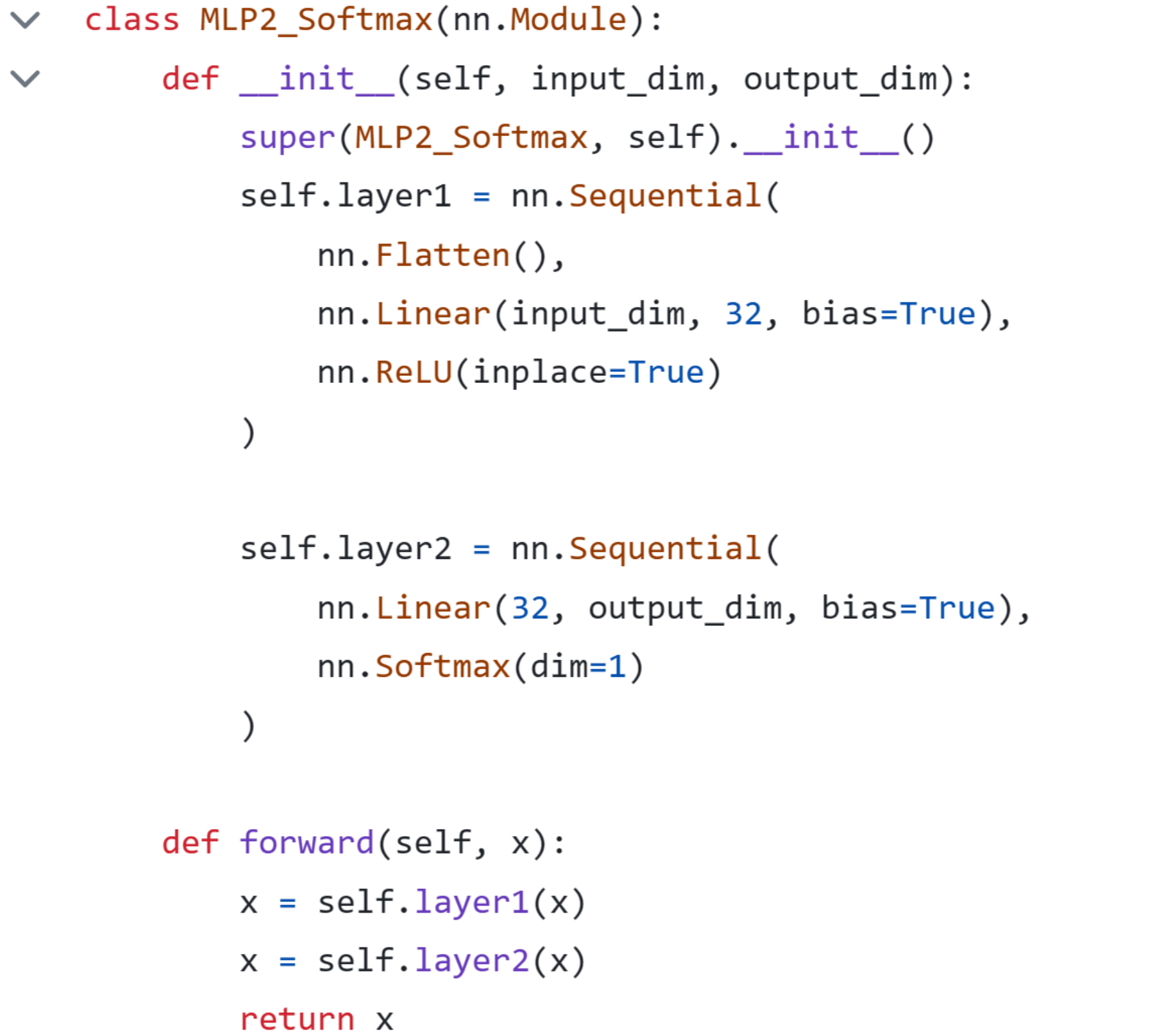}
        \label{fig:user_guidance_new_bottom_model}
        \end{minipage}%
    }% 
    % \sbox{\measurebox}{%
    %     \begin{minipage}[b]{.33\textwidth}
    %     \subfloat[]{
    %     \label{fig:user_guidance_new_config_file}
    %     \includegraphics[width=\textwidth,height=5cm]{figures/user_guidance/demo_new_config_file.pdf}
    %     }
    %     \end{minipage}
    % }
    % \usebox{\measurebox}\qquad
    % \begin{minipage}[b][\ht
    % \measurebox][s]{.64\textwidth}
    % \centering
    %     \subfloat[]{
    %         \label{fig:user_guidance_new_bottom_model}
    %         \includegraphics[width=\textwidth,height=2.9cm]{figures/user_guidance/demo_new_model.pdf}
    %     }
    %     \vfill
    %     \subfloat[]{
    %         \label{fig:user_guidance_new_attack}
    %         \includegraphics[width=\textwidth,height=2cm]{figures/user_guidance/demo_new_attack_class.pdf}
    %     }
    % \end{minipage}
    \caption{Adding new class.}
    \vspace{-1em}
\end{figure}

% \begin{figure}[!htb]
%     \centering
%     \begin{subfigure}[b]{.4\linewidth}
%     \includegraphics[width=\textwidth,height=5cm]{figures/user_guidance/demo_new_config_file.pdf}
%     \caption{New JSON file.} \label{fig:user_guidance_new_config_file}
%     \end{subfigure}\qquad
%     \begin{subfigure}[b]{.65\linewidth}
%     \includegraphics[width=\linewidth]{figures/user_guidance/demo_new_model.pdf}
%     \caption{New bottom model.} \label{fig:user_guidance_new_bottom_model} %,height=2cm
%     \vspace{2ex}
%     \includegraphics[width=\linewidth]{figures/user_guidance/demo_new_attack_class.pdf}
%     \caption{New attack.} \label{fig:user_guidance_new_attack} %,height=2cm
%     \end{subfigure}
    
%     \caption{Adding new class.}
%     \vspace{-1em}
% \end{figure}
\paragraph{Use Case: Adding a new attack.}
%\textbf{Adding a new attack.}
Below is a simple example for adding a new attack named 'BatchLabelReconstruction'. %(which is exactly Batch-level Label Inference (BLI) in our paper, and we just showcase how we implemented this attack as a demo for extending \verb|VFLAIR|). 
% and a new bottom model 'MLP2\_Softmax' to \verb|VFLAIR|. 
\begin{enumerate}
    \item New hyper-parameters for 'BatchLabelReconstruction' attack, e.g. attacker party ('party'), learning rate ('lr') and attack model training epochs ('epochs') in this case, can be added to the configuration file in the 'attack\_list' part (line 29-37 in \cref{fig:user_guidance_new_config_file}). Rename the new configuration file, e.g. \verb|my_new_config.json|, and save it under \verb|./src/configs/|.
    %\item No new datasets or data partitioning methods need to be added in this case, since the configuration uses MNIST dataset with equally partitioned features that \verb|VFLAIR| already supported.
    % \item New VFL bottom model 'MLP2\_Softmax' can be implemented in \verb|./src/models/mlp.py| like shown in \cref{fig:user_guidance_new_bottom_model}. In this case, loading this new model does not need to modify \verb|./src/load/LoadModels.py| file, however, the configuration file needs to be adjusted by specifying this new model as the local model for each party like shown in line 12 and 17 in \cref{fig:user_guidance_new_config_file}.
    %\item No new communication protocol is added in this case.
    \item Implement necessary code for the new attack 'BatchLabelReconstruction' by inheriting \verb|Attacker| class (see \cref{fig:user_guidance_new_attack}) under \verb|./src/evaluates/attacks/| folder for testing time attacks. Save it in file \verb|./src/evaluates/attacks/BatchLabelReconstruction.py|. Note that, the name of the attacker class implemented in the file should be aligned with the name of file, 'BatchLabelReconstruction' in this case. %Since this attack is an inference time attack, there is no need to modify the main training file \verb|./src/evaluates/MainTaskVFL.py|.
    %\item No new defense needs to be added in this case.
    \item Run the new attack with command \verb|python main_pipeline.py --configs| \verb|my_new_config| under \verb|./src/| folder.
\end{enumerate}

\paragraph{Use Case: Adding a new local model.}
We further provide a second example here on adding a new local model 'MLP2\_Softmax' to \verb|VFLAIR|. Assume this local model is for the new attack 'BatchLabelReconstruction' added above.
\begin{enumerate}
    \item New hyper-parameters for specifying the usage of 'MLP2\_Softmax' model as local model for each party are 'type' attribute in 'model\_list' part (line 12 and 17 in \cref{fig:user_guidance_new_config_file}), which should be specified as the name of the model. Then save the modified configuration file \verb|my_new_config.json| still under \verb|./src/configs/|.
    \item New VFL bottom model 'MLP2\_Softmax' can be implemented in \verb|./src/models/mlp.py| like shown in \cref{fig:user_guidance_new_bottom_model}. In this case, loading this new model does not need to modify \verb|./src/load/LoadModels.py| file.
    \item Run 'BatchLabelReconstruction' attack with new local model 'MLP2\_Softmax' with command \verb|python main_pipeline.py --configs| \verb|my_new_config| under \verb|./src/| folder.
\end{enumerate}
%After the above steps, users can 
% \yang{add 1-2 other examples. In principle, all extensible modules should have a detailed description like this in the github document.}
% }}

\section{VFLAIR Workflow}
\label{section:appendix_vfl_workflow}

% {\color{red}{
In this section, we present the key code modules and workflow of \verb|VFLAIR|.

\begin{figure}[!htb]
    \centering
    \includegraphics[width=0.99\linewidth]{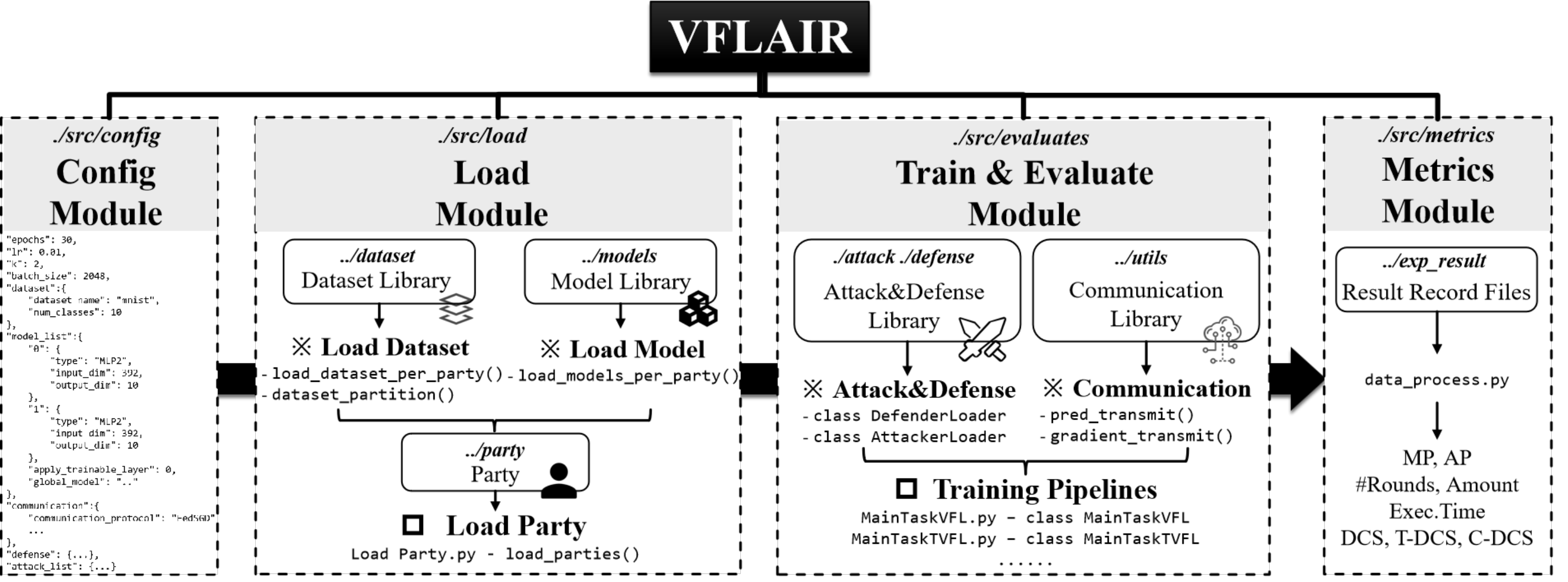}
    % \vspace{-1em}
\caption{Code modules and Workflow of VFLAIR.}
\label{fig:vflair_workflow}
% \vspace{-1em}
\end{figure} 

As shown in \cref{fig:vflair_workflow}, \verb|VFLAIR| contains $4$ key modules, namely Config Module, Load Module, Train \& Evaluate Module and Metrics Module. 

The Config Module first processes the user-specified configurations (see \cref{section:appendix_user_guidance} for detail), then passes it to the Load Module for party preparation. With function \verb|load_parties()|, each party separately loads its dataset and local model with function \verb|load_dataset_per_party()| and \verb|load_models_per_party()| respectively. %Note that, \verb|dataset_partition()| function, which is called inside \verb|load_dataset_per_party()| function, returns merely partial data that should be controlled by the party according to specified data partition in the Config Module.

Afterwards, the training pipeline starts in the Train \& Evaluate Module. Communication protocols are implemented within the training pipeline and are controlled by \verb|pred_transmit()| and \verb|gradient_transmit()| functions. All the supported defense are integrated into the training pipeline. For attacks, training-time attacks (e.g. TB and NTB attacks) are integrated and evaluated within training pipeline while inference-time attacks that do not affect the training flow (e.g. LI and FR attacks) are launched and evaluated after the training pipeline. However, they are all prepared by an \verb|AttackerLoader| object before the training pipeline starts.

Finally, the Metrics Module processes the recorded results using functions implemented in \verb|data_process.py| file and produces evaluation metrics including MP, AP, communication rounds (\#Rounds), amount of information transmitted between parties each round (Amount), execution time (EXec.Time), DCS, T-DCS and C-DCS.

%\subsection{Lightweight Workflow}
We compared our framework \verb|VFLAIR| with FATE, one of the most widely used FL platform supporting a broad range of VFL functionalities, on the system requirements for deployment in \cref{tab:hardware}, to demonstrate that \verb|VFLAIR| is a lightweight framework. According to FATE (\url{https://github.com/FederatedAI/FATE/blob/master/deploy/standalone-deploy/README.md}), deploying a stand-alone version of the FATE framework requires at least a 8 core CPU with 16G memory and 500G hard disk and the downloaded docker package for deployment is of size 4.92G for version 1.7.1.1. However, for our \verb|VFLAIR|, a 1 core CPU with less than 4G memory and less than 4.0G hard disk is required for installation and environment preparation.

\begin{table}[!htb]
\vspace{-1em}
\caption{Comparison of hardware requirements for installation between FATE and VFLAIR.}
\label{tab:hardware}
% \resizebox{0.998\linewidth}{!}{
    \centering
    \begin{tabular}{c|ccc}
    \toprule
        ~ & CPU & memory & Installation required hard disk \\
    \midrule
        FATE (stand-alone, version 1.7.1.1) & 8 core & 16G & 4.92G \\
        \verb|VFLAIR| & 1 core & 4G & $<$4G \\
    \bottomrule
    \end{tabular}
% }
\end{table}
% }}

\section{Detail Definition of Attack Performance and Attack Performance for Ideal Defenses} \label{sec:appendix_ap}
In this section, we give detailed definition of Attack Performance (AP) for each type of attack we studied in the literature. In this work, we use the AP for ideal defense AP*$=0.0$ indicate a complete failure of the attack..
% We believe AP* should equal to the worst Attack Performance (worst AP) of an attack, i.e. the AP of a random guess.
\begin{enumerate}
    \item In \textbf{Label Inference (LI) Attacks}, AP is defined as \textit{the ratio of correctly inferred labels} which can be regarded as the \textit{label inference accuracy}. That is,
    \begin{equation}\label{eq:ap_li} 
        AP_{\text{LI}} = \frac{\text{Number of Samples Correctly Inferred}}{\text{Number of Total Samples}}
    \end{equation}
    One exception is that for LI attacks under binary classification tasks, i.e. NS and DS (see \cref{sec:appendix_attacks} for explanation), we use \textit{the AUC of inferred labels} instead for better measurement especially for unbalanced dataset. For both settings, high label inference accuracy indicates a successful attack. % The worst AP should be equivalent to that of random guess, which equals to the reciprocal of the number of label types.

    % \item In \textbf{Feature Reconstruction (FR) Attacks}, AP is defined as \textit{the difference in feature restoration ability between a random guess and the attacker}. Let $U_0$ be the real data stored in the attacked party, $U_{rand}$ is a random guess on the real data $U_0$, $U_{rec}$ is the data reconstructed by the attacker and function $\text{MSE}(\cdot,\cdot)$ for calculating the mean square error between the $2$ function inputs. Therefore, AP can be expressed as:
    % \begin{equation}\label{eq:ap_fr} 
    %     AP_{\text{FR}} = \frac{\text{MSE}(U_0,U_{rand})-\text{MSE}(U_0,U_{rec})}{\text{MSE}(U_0,U_{rand})}=1-\frac{\text{MSE}(U_0,U_{rand})}{\text{MSE}(U_0,U_{rec})}.
    % \end{equation}
    % Also, a high AP indicates a strong attack. The worst AP should be $0.0$ which means the attacker cannot recover any useful information since it has only equal ability to random guess.
    \item In \textbf{Feature Reconstruction (FR) Attacks}, Mean Squared Error (MSE) is commonly employed as a measure to evaluate the quality of reconstruction, where a smaller MSE signifies a higher-quality reconstruction. So we define the AP for FR attacks as \textit{a negative correlation of MSE with a constant shift}. Let $C$ denote the constant shift, $U_0$ be the real data stored in the attacked party, $U_{rec}$ be the data reconstructed by the attacker and function $\text{MSE}(\cdot,\cdot)$ for calculating the mean square error between the $2$ function inputs, AP can be expressed as:
    \begin{equation}\label{eq:ap_fr} 
        AP_{\text{FR}} = C-\text{MSE}(U_0,U_{rec}).
    \end{equation}
    In our evaluation, we use $C=1.0$ and normalize all raw input features to $[0.0,1.0]$ before training and testing. In this case, $\text{MSE}(U_0,U_{rec})=\mathbb{E}[(u_0^{(f)}-u_{rec}^{(f)})^2] \in [0.0,1.0]$ where $u_0^{(f)},u_{rec}^{(f)}$ are the $f^{th}$ feature of $U_0,U_{rec}$ respectively. Thus, $AP_{\text{FR}} \in [0.0,1.0]$ and a high AP indicates a strong attack. 
    
    \item In \textbf{Targeted Backdoor Attacks}, AP is defined as \textit{the number of trigger inserted samples that is regarded as the target class by the VFL model to the total number of trigger inserted samples}, also \textit{the backdoor accuracy}, achieved by the final VFL model, that is:
    \begin{equation}\label{eq:ap_tb} 
        AP_{\text{TB}} = \frac{\text{Number of Triggered Samples Inferred as Target Class}}{\text{Number of Triggered Samples}}
    \end{equation}
    Same as the $2$ kinds of attacks above, a high AP indicates a successful backdoor attack. % As the samples with trigger attached are selected randomly, the worst AP should be the AP of random guess which is equal to the reciprocal of the number of classes of the main task. 
    
    \item In \textbf{Non-targeted Backdoor (NTB) Attacks}, \textit{the gap of MP on attacked samples relative to the overall MP on all the samples} is defined as AP. That is:
    \begin{equation}\label{eq:ap_ntb}        
        AP_{\text{NTB}} = MP_{\text{all}} - MP_{\text{attacked\_sample}}
    \end{equation}
    Still, a higher AP indicates a more successful attack. % The worst AP should be $0.0$, indicating that this attack does not affect the overall MP.
\end{enumerate}

\section{Evaluated Attacks} \label{sec:appendix_attacks}
In \verb|VFLAIR|, all the attacks listed below are supported and evaluated in the benchmark experiments. Users can easily extend other attacks and evaluate them using our framework.

\subsection{Label Inference (LI) Attacks}
In \textbf{Label Inference (LI) Attacks}, the attacker (also the passive party) hopes to steal the sensitive label information kept at active part. There are multiple ways for the passive attacker to infer private labels.
% , like exploiting sample-level gradient information ~\citep{li2022label,fu2021label,sun2022label,yang2022differentially}, or using batch-level gradient information~\citep{zou2022defending}, or inferring from the trained local models~\citep{fu2021label}. 
In this work, we evaluate $6$ label inference attacks to assess the vulnerability of a VFL system under label leakage threat, including: Norm-based Scoring (NS)~\citep{li2022label}, Direction-based Scoring (DS)~\citep{li2022label} and Direct Label Inference (DLI)~\citep{li2022label,zou2022defending} that exploit sample-level gradients, Batch-level Label Inference (BLI)~\citep{zou2022defending} that uses batch-level gradients, Passive Model Completion (PMC)~\citep{fu2021label} and Active Model Completion (AMC)~\citep{fu2021label} that infer label from trained local models.
% In \textbf{Label Inference (LI) Attacks}, the attacker (also the passive party) hopes to steal the sensitive label information kept at active part. There are multiple ways for the passive attacker to infer private labels, like exploiting sample-level gradient information ~\citep{li2022label,fu2021label,sun2022label,yang2022differentially}, or using batch-level gradient information~\citep{zou2022defending}, or inferring from the trained local models~\citep{fu2021label}. In this work, we considered the below label inference attack to evaluate the vulnerability of a VFL system facing under label leakage threat.
\begin{enumerate}
    \item Norm-based Scoring (NS)~\citep{li2022label}. NS is designed for binary classification task, and is based on an experimental observation that the norm of gradient $||g||_2$ for a positive sample is generally larger than that of a negative sample in an unbalanced dataset. Thus the passive party can calculate the norm of sample-level gradients transmitted back by the active party and then infer the sensitive label information according to the norm values.
    \item Direction-based Scoring (DS)~\citep{li2022label}. Similar to NS, DS also aims to infer label in binary classification task. DS is based on the observation that for any given sample pairs $x_a,x_b$, with their sample-level gradient denoted as $g_a,g_b$, the cosine similarity of these $2$ gradients $cos(g_a,g_b)=g_a^Tg_b/(||g_a||_2||g_b||_2)$ has a positive value if $x_a,x_b$ are of the same class, and a negative value if $x_a,x_b$ are of opposite classes. Attack can be launched using merely gradients that are transmitted back from the active party. However, to make the attacker more powerful, in the implementation of the attack, the gradient of one positive sample $g_+$ is additionally given.
    % \item {\color{red}{Spectral Attack (SA)~\citep{sun2022label}}} PA is also designed for binary classication tasks and utilize a proven fact that score $s=|<f(X)-\mu_F,v>|$ is likely to be different between positive and negative samples. Intuitively, this indicates that the mean value of the embedding distribution of the positive samples and that of the negative samples are far from each other. $f(X)$ is the embedding of a mini-batch and $\mu_F,v$ are its empirical mean and the top singular vector of the corresponding covariance matrix respectively. \tianyuan{@zixuan}
    \item Direct Label Inference (DLI)~\citep{li2022label,zou2022defending}. DLI is based on the observation that at passive party, for sample $\{x_i,y_i\}$, $y_i^{th}$ element of the gradient transmitted back from active party $g_i$ is the only element having opposite sign compared to others, thus the label information is revealed.
    \item Batch-level Label Inference (BLI)~\citep{zou2022defending}. There are cases where sample-level gradient information is protected, when using encryption techniques for example, and only batch-level gradient is revealed to passive party. BLI is designed to infer labels from this kind of setting. Inversion model is constructed and trained at passive party to invert label information from batch-level gradients.
    \item Passive Model Completion (PMC)~\citep{fu2021label}. PMC exploits the information of label inherit in the trained local model at passive party. By adding a randomly initialized linear layer on top of the trained local model to get a "completion model" and fine-tune the "completion model" with auxiliary label data, the passive party is able to guess the label of each sample.
    \item Active Model Completion (AMC)~\citep{fu2021label}. AMC is an enhanced version of PMC where the attacker (passive party) use a malicious local optimizer which adaptively scales up the gradient of each parameter in the adversary’s bottom model. This results in a more accurate trained local model as the overall VFL model is tricked to rely more on the maliciously optimized local model at passive party. With a more accurate local model, passive party is expected to obtain a better attack using the same model completion techniques as PMC.
\end{enumerate}

\subsection{Feature Reconstruction Attacks}
In \textbf{Feature Reconstruction (FR) Attacks}, the attacker aims to recover other parties' local features from its local model and all information received from other innocent parties. Both active and passive party can be the attacker when labels are not needed for completing the attack. When neural network models serves as local models, targeted features are limited to binary values~\citep{ye2022feature,peng2022binary} under a black-box setting in which the attacker neither has any knowledge of nor has access for querying the local model of the party under attack. Generation based on model inversion can be employed to reconstruct data by querying the trained model in a black-box oracle manner to recover tabular data~\citep{luo2021feature}. For white-box setting that has access to the trained model, model inversion techniques can also be used to recover image data~\citep{jin2021cafe} with the knowledge of trained local model, or with prior knowledge about data~\citep{li2022ressfl,jiang2022comprehensive}. In this work, we test the following $2$ FR attacks:
% In this work we tested $2$ FR attacks including: 
% % white-box attack CAFE~\citep{jin2021cafe} by model inversion and 
% Generative Regression Network (GRN)~\citep{luo2021feature} using generative model that relies on querying the trained mode in a black-box manner, as well as Training-based Back Mapping by model inversion (TBM)~\citep{li2022ressfl} that relies on auxiliary i.i.d. data.
\begin{enumerate}
    % \item {\color{red}{CAFE~\citep{jin2021cafe}.}} With the assumption that the attacker knows the model structure of the party attacked (i.e. white-box), CAFE divides the entire data leakage attack procedure into several steps and performs step by step recovery of data by gradient inversion. \tianyuan{@zixuan}
    % CAFE is tested to be a strong feature reconstruction attack compared to former data reconstruction attacks in HFL.
    
    \item Generative Regression Network (GRN)~\citep{luo2021feature}. By querying the trained VFL model, GRN attack reconstructs data features by matching the VFL prediction of real and reconstructed data. A generative model is trained to map random noise to targeted features.
    
    \item {Training-based Back Mapping by model inversion (TBM)~\citep{li2022ressfl}.} When auxiliary i.i.d. data of the local private data is available at the attacker, TBM utilizes these data to train a generative model in order to map the embedding feature of the victim party back to the original input feature. A strong assumption is used for TBM attack in which the attacker can query the whole trained VFL model with the data it obtained.
\end{enumerate}

\subsection{Targeted Backdoor Attacks}
\textbf{Targeted Backdoor (TB) Attacks} have a clear incorrect leading target and aims to manipulate the VFL model's behavior on samples marked with backdoor related features. We evaluated Label Replacement Backdoor (LRB)~\citep{zou2022defending} in this work to assess the vulnerability of a VFL system under targeted backdoor threat. 
\begin{enumerate}
    \item Label Replacement Backdoor (LRB)~\citep{zou2022defending} aims to assign an attacker-chosen label (target label) $\tau$ to input data with a specific pattern (i.e. a trigger). The passive attacker is assumed of knowing a few clean samples from the target class. The triggered poison samples are created locally by adding triggers to randomly selected samples from the passive attacker's own data. In training, the attacker replaces the embedding of a known clean sample from the target class with that of a triggered poison sample to replace the corresponding label of the poisoned sample.
\end{enumerate}

\subsection{Non-targeted Backdoor Attack}
Unlike TB attacks, \textbf{Non-targeted Backdoor (NTB) Attacks} only aim to affect the utility of the VFL model. Methods like adding noise to some randomly selected samples~\citep{zou2022defending} or by adding missing features~\citep{liu2021rvfr} can be exploit during inference. The attacks used to assess the safety of a VFL system under NTB are listed below:
\begin{enumerate}  
    \item Noisy-Sample Backdoor (NSB)~\citep{zou2022defending}. In NSB attack, random noise $\sim \mathcal{N}(0,2)$ is attached to arbitrarily selected samples by passive party to harm the model utility of VFL.
    
    % {\color{red}{\item Adversarial-Sample Backdoor (ASB)~\citep{liu2021rvfr}}} 
    
    % {\item Noisy-Label Backdoor (NLB)~\citep{jiang2022towards}}. In NLB attack, random noise is add to the label possessed by active party, simulating circumstances where high-quality labeled data is difficult to obtain, which is very common in reality.
    % \zixuan{Do we still need this attack? I've found one paper under VFL setting and aligned our code with theirs }

    \item Missing Feature (MF)~\citep{liu2021rvfr}. In MF attack, the embedding of passive party's local features of some samples are missing either due to the unstable network issue or due to the intentional hiding by the passive attacker. Missed features are treated as all $0$ in the implementation.
\end{enumerate}
Detail attack hyper-parameters settings can be seen in \cref{subsec:appendix_attack_parameters}.

\section{Evaluated Defenses} \label{sec:appendix_defenses}
In \verb|VFLAIR|, all the defense methods listed in below are supported and evaluated in the benchmark experiments. Like for the attacks, users can easily extend other defense techniques and evaluate them using our framework.

\begin{enumerate}
    \item Differential privacy with Gaussian noise (G-DP)~\citep{dwork2006DP,li2022label,zou2022defending} or Laplace noise (L-DP)~\citep{dwork2006DP,fu2021label,zou2022defending} added to gradients or local model predictions to defend against attacks launched at passive or active party.
    
    \item Gradient Sparsification (GS)~\citep{aji2017sparse,fu2021label,zou2022defending} defends against attacks by dropping elements in gradients that are close to $0$.
    
    % \item {\color{red}{Discrete Gradients (DG)~\citep{fu2021label} by mapping gradients into discrete value bins.}}
    
    % \item {\color{red}{Random Response with Prior (RRwP)~\citep{ghazi2021deep} by randomly replacing some real label with fake ones according to prior knowledge of label distribution during training to gain label differential privacy (label DP).}} \tianyuan{@zixuan}
    
    \item Gradient Perturb (GPer)~\citep{yang2022differentially} defends against label leakage attacks by perturbing the gradients with the sum of gradients from each class added with random scalars. Label-DP is guaranteed using GPer.
    
    \item Distance Correlation (dCor)~\citep{sun2022label,vepakomma2019reducing} defends against attacks by applying correlation regularization. When the passive party applies this defense to defend against feature reconstruction attacks, distance correlation is calculated between input features $X_k$ and local embedding $G(X_k,\theta_k)$ at each party in order to limit redundant information of $X_k$ kept in $G(X_k,\theta_k)$. In contrast, when active party applies this defense to defend against label inference attacks or backdoor attacks, distance correlation is calculated between label $Y$ and $G(X_k,\theta_k)$ of each party to limit redundant information of $Y$ kept in $G(X_k,\theta_k)$. $\log(dCor(X_k,G(X_k,\theta_k))$ and $\log(dCor(Y,G(X_k,\theta_k))$ are used in practice instead of $dCor(X_k,G(X_k,\theta_k)$ and $dCor(Y,G(X_k,\theta_k)$ to stabilize training.
    
    \item Confusional AutoEncoder (CAE)~\citep{zou2022defending} defends against label related attacks by disguising labels with an encoder and reconstruct the original label with the paired decoder. Confusion is added to map one class to multiple classes to further disguise label information.
    
    \item Discrete-gradient-enhanced CAE (DCAE)~\citep{zou2022defending} defends against attacks by applying discrete gradients along with CAE to get a stronger defense.
    
    \item Mutual Information regularization Defense method (MID)~\citep{zou2023mutual} defends against attacks by limiting the information of label contained in local embedding. 
\end{enumerate}
Detailed hyper-parameters of these defenses can be found in \cref{subsec:appendix_defense_parameters}.

\section{Experimental Settings} \label{sec:appendix_experimental_settings}

We mainly use NVIDIA GeForce RTX 3090 for all the benchmark experiments except for tree-based VFL related experiments for which we use Intel(R) Xeon(R) CPU E5-2650 v2 instead. All the experiments are repeated $5$ times with different random seeds. The stopping criterion is determined as reaching a predefined number of epochs, while ensuring convergence at the same time. The reported results include both the mean values and the corresponding standard deviations. Other experimental settings are listed below.
% For each VFL main task performance benchmark experiments, the mean and standard deviation of MP are reported. While for each attack and defense related experiment, the mean value of both AP and MP are recorded for each attack under each defense. 
% Corresponding DCS, T-DCS and C-DCS are then calculated using the mean values of AP and MP. 

\subsection{Models and Datasets} \label{subsec:appendix_model_dataset}

\begin{table}[!tb]
\caption{Summary of evaluated datasets under NN-based VFL. In "\#Samples" column, the values denote the number of training and testing samples separately. In "Feature Partition" column, if the number of features is equal among each party, we present only one number; '[]' marks the number of features after extending discrete feature to one-hot features; '/' is used to separate the feature partition for 2-party VFL and 4-party VFL, otherwise the number stands for the feature partition of 2-party VFL. In "\#Parameters / \#Nodes" column, the value denotes the number of trainable parameters for each local model for neural network or logistic regression local model and denotes the number of nodes in total of each side for tree-based local model. $I$ and $C$ stands for the number of input features specified in \cref{subsec:appendix_model_dataset} and \#Class respectively. (p) and (a) refer to passive and active party respectively.} % In "\#Samples with Auxiliary" column, the values denote the number of training, testing and auxiliary samples separately.
\label{tab:datasets}
% \resizebox{0.998\linewidth}{!}{
%     \centering
%     \begin{tabular}{c|c|c|c|c|c}
%     \toprule
%         Dataset & \#Class & Balance & \#Samples & \#Samples with Auxiliary & Bottom Model\\
%          % &  &  & (train, test) & (train, test, auxiliary) & \\
%     \midrule
%         MNIST~\citep{MNISTdataset} & $10$ & yes & $60000; 10000$ & {\color{red}{$54000; 10000; 6000$}} & MLP-2\\
%         CIFAR10~\citep{krizhevsky2009learning} & $10$ & yes & $50000; 10000$ & {\color{red}{??}} & {\color{red}{??}} \\
%         NUSWIDE~\citep{NUSWIDEdataset} & $5$ & no & {\color{red}{??}} & {\color{red}{??}} & {\color{red}{??}} \\
%         Breast Cancer~\citep{street1993nuclear} & $2$ & {\color{red}{??}} & {\color{red}{??}} & {\color{red}{??}} & {\color{red}{??}} \\
%     \bottomrule
%     \end{tabular}
%     }
    % \centering
    % \begin{tabular}{c|c|c|c|c}
    % \toprule
    %     Dataset & \#Class & Balance & \#Samples & Local Model\\
    %      % &  &  & (train, test) & (train, test, auxiliary) & \\
    % \midrule
    %     MNIST~\citep{MNISTdataset} & $10$ & yes & $60000; 10000$ & MLP-2\\
    %     CIFAR10~\citep{krizhevsky2009learning} & $10$ & yes & $50000; 10000$ & {\color{red}{??}} \\
    %     NUSWIDE~\citep{NUSWIDEdataset} & $5$ & no & {\color{red}{??}} & {\color{red}{??}} \\
    %     Breast Cancer~\citep{street1993nuclear} & $2$ & {\color{red}{??}} & {\color{red}{??}} & {\color{red}{??}} \\
    % \bottomrule
    % \end{tabular}
% }
\resizebox{0.998\linewidth}{!}{    
    \centering
    \begin{tabular}{c|c|c|c|c|c}
    \toprule
        Dataset & \#Class & \#Samples & \shortstack{Feature\\Partition} & \shortstack{Local Model\\(active \& passive party)} & \shortstack{\#Parameters / \#Nodes}\\
         % &  & (train, test) & (train, test, auxiliary) & \\
    \midrule
        % MNIST~\citep{MNISTdataset} & $10$ & $60000; 10000$ & MLP-2 (active \& passive party)\\
        % CIFAR10~\citep{krizhevsky2009learning} & $10$ & $50000; 10000$ & Resnet18 (active \& passive party)\\
        % CIFAR100~\citep{krizhevsky2009learning} & $100$ & $50000; 10000$ & Resnet18 (active \& passive party)\\
        % NUSWIDE~\citep{NUSWIDEdataset} & $5$ & $60000; 40000$ & MLP-2 (active \& passive party) \\
        % Breast Cancer~\citep{street1993nuclear} & $2$ & $454; 114$ & MLP-2 (active \& passive party)\\
        % Diabetes~\citep{Diabetes1999dataset} & 2 & $614;154$ & Logistic Regression (active \& passive party) \\
        % Adult Income~\citep{AdultIncome1996dataset} & 2 & $34153;14637$ & MLP-4 (active \& passive party)\\
        % Credit~\citep{Dua:2019}  & 2 & $20000;10000$ & RandomForest/XGBoost (active \& passive party)\\
        % Nursery~\citep{Dua:2019} & 5 & $8640;4320$ & RandomForest/XGBoost (active \& passive party)\\
        MNIST~\citep{MNISTdataset} & $10$ & $60000; 10000$ & $392$ & MLP-2 ($I$-$32$-$C$) & $12.9$K \\ % 12906
        \hline
        \\[-1em]
        CIFAR10~\citep{krizhevsky2009learning} & $10$ & $50000; 10000$ & $512$ / $258$ & Resnet18 & $11.17$M \\ % 11173962
        \hline
        \\[-1em]
        CIFAR100~\citep{krizhevsky2009learning} & $100$ & $50000; 10000$ & $512$ / $258$ & Resnet18 & $11.17$M \\ % 11173962
        \hline
        \\[-1em]
        NUSWIDE~\citep{NUSWIDEdataset} & $5$ & $60000; 40000$ & $1000$ (p), $634$ (a) & MLP-2 ($I$-$32$-$C$) & $32.2$K (p), $20.5$K (a)\\ % 32197(passive), 20485(active)
        \hline
        \\[-1em]
        Breast Cancer~\citep{street1993nuclear} & $2$ & $454; 114$ & $15$ & MLP-2 ($I$-$32$-$C$) & $2.3$K \\ % 2306
        \hline
        \\[-1em]
        Diabetes~\citep{Diabetes1999dataset} & 2 & $614;154$ & $4$ & Logistic Regression & 10 \\
        \hline
        \\[-1em]
        Adult Income~\citep{AdultIncome1996dataset} & 2 & $34153;14637$ & $7$ [$15$ (p), $93$ (a)] & MLP-4 ($I$-$64$-$128$-$64$-$C$) & $17.7$K (p), $22.7$K (a) \\
        \hline
        \\[-1em]
        Criteo~\citep{guo2017deepfm_Criteo} & 2 & $183362;1650263$ & $13$ (p), $26$ (a) & Wide\&Deep~\citep{cheng2016wide} & $1040.1$M (p), $73.2$K (a) \\
        \hline
        \\[-1em]
        Avazu~\citep{qu2018product_Avazu} & 2 & $727722;80857$ & $9$ (p), $13$ (a) & Wide\&Deep~\citep{cheng2016wide} & $144.1$M (p), $208.1$M (a) \\
        \hline
        \\[-1em]
        Cora~\citep{mccallum2000automating_Cora} & 7 & $140;1000$ & $716$ & 2-layer GCN~\citep{kipf2017semisupervised} & $23.2$K \\ % $2708;2708$
        \hline
        \\[-1em]
        News20-S5~\citep{lang1995News20} & 5 & $4000;1000$ & $49658$ & MLP-5 & $3.2$M \\
        \hline
        \\[-1em]
        \multirow{3}{*}{Credit~\citep{Dua:2019}} & \multirow{3}{*}{2} & \multirow{3}{*}{$24000;6000$} & \multirow{3}{*}{$12$ (p), $11$ (a)} & Logistic Regression & $26$ (p), $24$ (a) \\
        \cline{5-6}
        \\[-1em]
        ~ & ~ & ~ & ~ & MLP-4 ($I$-$100$-$50$-$20$-$C$) & $7.4$K (p), $7.3$K (a) \\
        \cline{5-6}
        \\[-1em]
        ~ & ~ & ~ & ~ & RandomForest / XGBoost & $580$ / $620$ \\
        \hline
        \\[-1em]
        \multirow{3}{*}{Nursery~\citep{Dua:2019}} & \multirow{3}{*}{5} & \multirow{3}{*}{$10368;2592$} & \multirow{3}{*}{$4$} & Logistic Regression & $40$ (p), $65$ (a) \\
        \cline{5-6}
        \\[-1em]
        ~ & ~ & ~ & ~ & MLP-3 ($I$-$200$-$100$-$C$) & $22.2$K (p), $23.2$K (a) \\
        \cline{5-6}
        \\[-1em]
        ~ & ~ & ~ & ~ & RandomForest / XGBoost & $402$ / $415$ \\
    \bottomrule
    \end{tabular}
    % \vspace{0.5em}
}
\end{table}

We construct our benchmark experiments on the following $9$ datasets. The local models used for each dataset are listed in \cref{tab:datasets}. 
Additionally, for the splitVFL setting, we employ a $1$-layer MLP model as the global model for datasets other than Cora, for which a 1-layer graph convolution layer is adopted to pair with the GCN it adopts as local model, while a non-trainable global softmax function is used for the aggVFL setting.
\begin{itemize}
    \item MNIST~\citep{MNISTdataset}. The MNIST dataset comprises handwritten digits and consists of a training set with $60,000$ examples, along with a test set containing $10,000$ examples, distributed across $10$ classes. All samples have been standardized in size and centered within $32\times32$ pixel grayscale images. In our experiments, we utilize the entire dataset, and each sample is horizontally divided into equal halves and assigned to respective parties.
    \item CIFAR10~\citep{krizhevsky2009learning}. The CIFAR10 dataset consists of $60,000$ colour images, each of size $32\times32$, of $10$ classes, with $6,000$ images per class: $5,000$ for training and $1,000$ for testing. All the data are used in our experiments. Each sample is horizontally split in to equal halves and assigned to respective parties under 2-party VFL setting, or is equally split into $4$ parts of size $16\times16$ and assigned to respective parties under 4-party VFL setting.
    \item CIFAR100~\citep{krizhevsky2009learning}. Similar to CIFAR10 dataset, CIFAR100 dataset consists of $60,000$ $32\times32$ colour images, but are distributed across $100$ classes containing $600$ images each: $500$ for training and $100$ for testing. All the data are used in our experiments. Each sample is horizontally split in to equal halves and assigned to each party under 2-party VFL setting, or is equally split into $4$ parts of size $16\times16$ and assigned to each party under 4-party VFL setting.
    \item NUSWIDE~\citep{NUSWIDEdataset}. NUSWIDE dataset is a web image dataset that includes: (1) $269,648$ images and the associated tags from Flickr; (2) $5$ types of low-level features extracted from the images and $1$ bag of words feature of $500$ dimension; (3) ground-truth label for $81$ concepts. In our experiments, we use all the $5$ low-level image features, that is a total of $634$ features, for the active party and use the top $1000$ frequent tags associated to the images for the passive party. We use only data from 'buildings', 'grass', 'animal', 'water' and 'person' in our experiments. For binary classification tasks, we use only data from 'clouds' and 'person'.
    \item Breast Cancer~\citep{street1993nuclear}.  Breast Cancer dataset is a tabular dataset consists of $30$ statistical descriptions of $568$ breast tumors which are categorized as either malignant (cancerous) or benign(non-cancerous). In our experiments, we use $20\%$ of the whole dataset samples for testing and the rest for training. Each party possesses $15$ statistical description features.
    \item Diabetes~\citep{Diabetes1999dataset}. The Diabetes dataset comprises $768$ samples, each accompanied by $8$ diagnostic measurements corresponding to individual patients, along with the diagnosis indicating whether the patient has diabetes. We also use $20\%$ of the whole dataset samples for testing and the rest for training. Each party possesses $4$ diagnostic measurements.
    \item Adult Income~\citep{AdultIncome1996dataset}. The Adult Income dataset consists of $14$ distinct features, including demographic information such as age, education, and occupation, collected from a dataset of $48,790$ individuals. The primary task associated with this dataset is to predict whether a person earns an annual income exceeding $50$K. We use $30\%$ of the whole dataset samples for testing and the rest for training. Category features are first changed to one-hot features before sending into the model for prediction. Passive party obtains the $6$ non-category features as well as a category feature 'workclass', while the active party controls the rest $7$ category features. After extending the categorical discrete features, $15$ and $93$ features are kept separately at each party.
    \item Criteo~\citep{guo2017deepfm_Criteo} The Criteo dataset contains real world click-through data of display advertisements served by Criteo of $7$ days and whether the advertisement has been clicked or not. The primary task is to predict whether clicktion is done. Only $1,833,625$ samples from the whole dataset is used in our experiments with $90\%$ used for training and the rest $10\%$ used for testing following previous work~\citep{fu2022towards}. Each sample has $26$ anonymous categorical features assigned to passive party and $13$ continuous features assigned to active party also following previous work~\citep{fu2022towards}.
    \item Avazu~\citep{qu2018product_Avazu} The Avazu dataset contains $11$ days real world click-through data from Avazu. The primary task is also to predict whether clicktion is done. We use only $808,579$ samples from the whole dataset in our experiments with $90\%$ for training and rest $10\%$ for testing following previous work~\citep{fu2022towards}. Each sample has $9$ anonymous categorical features assigned to passive party and $13$ categorical features that are not anonymous assigned to active party also following previous work~\citep{fu2022towards}.
    \item Cora~\citep{mccallum2000automating_Cora} The Cora dataset is a citation network dataset with nodes representing computer science research papers and edges representing citations between them. The task is to predict the category of a node, which falls into one of the following $7$ categories: 'Neural\_Networks', 'Probabilistic\_Methods', 'Genetic\_Algorithms', 'Theory', 'Case\_Based', 'Reinforcement\_Learning' and 'Rule\_Learning'. A total of $2708$ nodes are provided in the dataset, with $140$ nodes with labels and are used as training data. We use $1000$ of the rest for testing. A total of $1432$ features are provided for each node and we split them equally and signed to each party.
    \item News20-S5~\citep{lang1995News20} The News20 dataset, also the 20 Newsgroups dataset, is a collection of approximately $20,000$ newsgroup documents, partitioned (nearly) evenly across 20 different newsgroups. We use only the first $5$ categories, namely 'alt.atheism', 'comp.graphics', 'comp.os.ms-windows.misc', 'comp.sys.ibm.pc.hardware' and 'comp.sys.mac.hardware' for our experiments, resulting in a total of $5,000$ samples. We use $80\%$ of the data for training and the rest $20\%$ for testing. Each party controls $49,658$ features in our setting.
    \item Credit~\citep{Dua:2019}. Credit dataset comprises $30,000$ samples, each with $23$ features, including attributes like age and education level. The primary objective here is to predict the likelihood of default payment. For our experimentation, we allocate $20\%$ of the entire dataset for testing, while the remaining $80\%$ is designated for training. While the passive party owns the $12$ features related to the amount of bill statement and previous statement, the active party possesses the other $11$ features related background information and repay statement.
    \item Nursery~\citep{Dua:2019}. Nursery dataset consists of $12,960$ samples of $8$ features with the task of predicting the recommendation level of applications for nursery schools. We also use $20\%$ of the whole dataset samples for testing and the rest for training. While the active party owns $4$ features related to family structure and financial standings, the passive party has other $4$ features.
\end{itemize}

\subsection{VFL Main Task Training Hyper-parameters}\label{subsec:appendix_main_task_parameters}

% \yang{what is the learning rate of FedBCD? should provide all experimental details and hyperparameters in Appendix.}
We introduce the hyper-parameters uesd for MP evaluations and benchmarks showed in \cref{tab:NN_MP,tab:4party_MP,tab:tree_MP,tab:communication_MP,tab:real_world_dataset_MP,tab:different_local_models_MP} in the following.

For NN-based VFL, the learning rate and training epochs use for reporting the MP listed in \cref{tab:NN_MP,tab:4party_MP,tab:communication_MP,tab:real_world_dataset_MP,tab:different_local_models_MP} are included in \cref{tab:NN_MP_parameters,tab:NN_MP_parameters_communication}. A batchsize of $1024$ is used throughout all the experiments (except for MNIST, Criteo, Avazau and News20-S5 which uses a batchsize of $2048, 8192, 8192, 128$ respectively) and is not listed in the table. For tree-based VFL, for reporting the MP listed in \cref{tab:tree_MP}, each party is equipped with a number of $5$ trees each of depth $6$ under all circumstances. Note that learning rate is only utilized for XGBoost and is set to $0.003$ in the experiments. 
% For tree-based VFL, the learning rate and the number of trees used for reporting the MP listed in \cref{tab:tree_MP} are listed in \cref{tab:tree_MP_parameters}. 

% In order to ensure robustness and aptness across varied VFL scenarios, we meticulously tweak the configurations. In splitVFL, we employ a single Fully Connected Layer as the global model, whereas in aggVFL, the global model is an average function. Compared with FedSCG, we reduce the learning rate to 0.005 for MNIST in a FedBCD scenario. Also, the training epoch for PMC and AMC on MNIST is aligned to 100, enabling a thorough exploration and revelation of the attack functions.

% \guzx{I only write configurations for MNIST and CIFAR10,please add NUSWIDE and BreastCancer into the context.}

\begin{table}[!tb]
\caption{Hyper-parameters for \cref{tab:NN_MP,tab:4party_MP,tab:real_world_dataset_MP,tab:different_local_models_MP}. LR is short for learning rate.}
\label{tab:NN_MP_parameters}
\resizebox{0.998\linewidth}{!}{
    \centering
    \begin{tabular}{c||c|c||c|c||c|c||c|c}
    \toprule
    \multirow{2}{*}{\shortstack{Dataset \\ \,}} & \multicolumn{2}{c||}{aggVFL, FedSGD} & \multicolumn{2}{c||}{aggVFL, FedBCD} & \multicolumn{2}{c||}{splitVFL, FedSGD} & \multicolumn{2}{c}{splitVFL, FedBCD} \\
    \cline{2-9}
    \\[-1em]
    ~ & LR & epochs & LR & epochs & LR & epochs & LR & epochs \\
    \midrule
    MNIST & 0.01 & 30 & 0.005 & 30 & 0.01 & 30 & 0.005 & 30 \\
    CIFAR10 (2-party) & 0.001 & 30 & - & - & 0.001 & 30 & - & - \\
    CIFAR10 (4-party) & 0.001 &30  & - & - & 0.001 & 1024 & - & - \\
    CIFAR100 (2-party) & 0.01 & 30 & - & - & 0.01 & 30 & - & - \\
    CIFAR100 (4-party) &  0.001& 40 & - & - & 0.001 & 40 & - & - \\
    NUSWIDE & 0.003 & 10 & 0.003 & 10 & 0.006 & 10 & 0.003 & 10 \\
    Breast Cancer & 0.05 & 50 &  0.005& 50 &0.05  &50  &0.005  & 50 \\
    Diabetes & 0.05 & 80 & 0.01 & 80 & 0.05 & 80 & 0.01 & 80 \\
    Adult Income & 0.01 & 50 & 0.001 & 50 & 0.01 & 50 & 0.001 & 50 \\
    Criteo & 0.0001 & 5 & - & - & 0.0001 & 5 & - & - \\
    Avazu & 0.0001 & 10 & - & - & 0.0001 & 10 & - & - \\
    Cora & 0.01 & 20 & - & - & 0.01 & 20 & - & - \\
    News20-S5 & 0.001 & 80 & - & - & 0.002 & 80 &  & - \\
    Credit (LR) & 0.6 & 300 & - & - & 0.6 & 300 & - & - \\
    Credit (NN) & 0.01 & 40 & - & - & 0.01 & 40 & - & - \\
    Nursery (LR) & 0.5 & 40 & - & - & 0.5 & 40 & - & - \\
    Nursery (NN) & 0.01 & 40 & - & - & 0.01 & 40 & - & - \\
    \bottomrule
\end{tabular}
}

\end{table}

\begin{table}[!tb]
\caption{Hyper-parameters for \cref{tab:communication_MP}. Note that the hyper-parameters for the FedSGD and FedBCD in \cref{tab:communication_MP} are the same as that in \cref{tab:NN_MP}.}
\label{tab:NN_MP_parameters_communication}
\centering
% \resizebox{0.60\linewidth}{!}{
    \begin{tabular}{c||c|c||c|c||c|c|c}
    \toprule
    \multirow{2}{*}{\shortstack{Dataset}} & \multicolumn{2}{c||}{Quantize} & \multicolumn{2}{c||}{Top-k} & \multicolumn{3}{c}{CELU-VFL} \\
    \cline{2-8}
    \\[-1em]
    ~ & LR & epochs & LR & epochs & LR & epochs & $\xi$\\
    \midrule
    MNIST & 0.03 & 30 & 0.05 & 30 & 0.008 & 30 & 0.5 \\
    NUSWIDE & 0.003 & 10 & 0.003 & 10 & 0.008 & 10 & 0.8 \\
    \bottomrule
\end{tabular}
% }
\end{table}

% \begin{table}[!tb]
% \centering
% \caption{Hyper-parameters for \cref{tab:tree_MP}. LR is short for learning rate.}
% \label{tab:tree_MP_parameters}
% \resizebox{0.998\linewidth}{!}{
%     \centering
%     \begin{tabular}{c||c|c|c||c|c|c||c|c|c||c|c|c}
%     \toprule
%    \multirow{2}{*}{\shortstack{Dataset \\ \,}} & \multicolumn{3}{c||}{\shortstack{Random Forest \\ w/o Encryption \\ \,}} & \multicolumn{3}{c||}{\shortstack{XGBoost\\ w/o Encryption \\ \,}} & \multicolumn{3}{c||}{\shortstack{Random Forest \\ w/ Encryption \\ \,}} & \multicolumn{3}{c}{\shortstack{XGBoost\\ w/ Encryption\\ (a.k.a. SecureBoost)}} \\
%   % \begin{tabular}[c]{@{}c@{}}\end{tabular} &
%   % \begin{tabular}[c]{@{}c@{}}XGBoost\\ w/o Encryption\end{tabular} &
%   % \begin{tabular}[c]{@{}c@{}}Random Forest \\ w/ Encryption\end{tabular} &
%   % \begin{tabular}[c]{@{}c@{}}XGBoost\\ w/ Encryption\\ (a.k.a. SecureBoost)\end{tabular} \\ 
%     \cline{2-13}
%     \\[-1em]
%     ~ & LR & Depth & \#Trees & LR & Depth & \#Trees & LR & Depth & \#Trees & LR & Depth & \#Trees \\
%     \midrule
%     Credit  & - & 6 & 5 & 0.03 & 6 & 5 & - & 6 & 5 & 0.03 & 6 & 5 \\
%     Nursery & - & 6 & 5 & 0.03 & 6 & 5 & - & 6 & 5 & 0.03 & 6 & 5  \\
%     \bottomrule
% \end{tabular}
% }
% \end{table}

\subsection{Attack Hyper-parameters} \label{subsec:appendix_attack_parameters}

Our evaluated attacks can be categorized into \textbf{Label Inference} (LI) attacks, \textbf{Feature Reconstruction} (FR) attacks, \textbf{Targeted Backdoor} (TB) attacks and \textbf{Non-targeted Backdoor} (NTB) attacks. Each attack is launched separately on the VFL systems trained with the hyper-parameters listed above in \cref{subsec:appendix_main_task_parameters}. Specific hyper-parameters for each attack is listed in below if exist.
\begin{itemize}
    \item For LI attacks, attacker is the passive party. 
    In \textbf{NS}~\citep{li2022label}, \textbf{DS}~\citep{li2022label} and \textbf{DLI}~\citep{li2022label,zou2022defending}, no other attack related hyper-parameter needs to be specified. However, for MNIST dataset, in order to achieve a higher MP, we use a learning rate of $0.001$ for VFL model training under binary classification tasks (DS attack).
    In \textbf{BLI}~\citep{zou2022defending}, learning rate and number of training epoch for the inference model are set to $0.05$ and $10000$ respectively. 
    In \textbf{AMC} and \textbf{PMC}~\citep{fu2021label}, we randomly take $4$ training samples from each class to form the auxiliary labeled dataset for training of the "completion model". Both the trained local model and the classification head used to complete the model are fine-tuned using the auxiliary labeled dataset and some non-labeled data from the training dataset. Number of "completion model" training epochs, learning rate and training batchsize are set to $25,0.002,16$ separately for MNIST and CIFAR10 datasets and are set to $20,0.002,16$ separately for NUSWIDE dataset. Note that for better attack stability, we change the VFL main task learning epoch to $100$ for PMC and AMC under all the $3$ datasets.
    
    \item For FR attacks, attacker is the active party. 
    In \textbf{GRN}~\citep{luo2021feature}, following the original paper, the reconstruction model is trained for $60$ epochs with a batchsize of $1024$ for all datasets, as well as a learning rate of $0.005$ for MNIST dataset and $0.0001$ for CIFAR10 dataset.
    While in \textbf{TBM}~\citep{li2022ressfl}, the reconstruction model is trained for $50$ epochs with a learning rate of $0.0001$ and a batchsize of $32$. Note that, since TBM requires auxiliary data with the same distribution as the training data, we use $10\%$ of the training data to form the auxiliary dataset and train the VFL model using only the rest $90\%$ following the original work~\citep{li2022ressfl}.

    \item For TB attacks, \textbf{LRB}~\citep{zou2022defending} is evaluated with the passive party being the attacker. Target class is randomly selected for each training. Following previous work~\citep{zou2022defending}, $1\%$ of the data is randomly selected and attached with specific trigger of $4$ pixels to form backdoor samples for MNIST and CIFAR10 datasets, while samples that have value $1$ at the last bit of the tag data, which take up less than $1\%$ of the whole dataset, are treated as backdoor samples for NUSWIDE dataset. 
    % \tianyuan{For Breast Cancer dataset, we select samples ... as the backdoor data samples which takes up less than $1\%$ of the whole dataset, same to that of NUSWIDE.}

    \item For NTB attacks, attack is done by the passive party at inference time. 
    In \textbf{NSB}~\citep{zou2023mutual}, $1\%$ of the testing samples are randomly selected and added with random noise sampled from $\mathcal{N}(0,2)$ by the attacker party. 
    % In \textbf{NLB)}~\citep{jiang2022towards}, two noise type("asymmetric" \& "pairflip") is evaluated with a noise rate of $0.2$. 
    In \textbf{MF}~\citep{liu2021rvfr}, a missing rate of $0.25$ is evaluated which means during inference, a quarter of the intermediate local model output from the passive attacker is replaced with $\mathbf{0}$ before transmitted to the active party for aggregation.

\end{itemize}

\subsection{Defense Hyper-parameters} \label{subsec:appendix_defense_parameters}
To ensure a comprehensive benchmark, we evaluate a total of 8 defense methods using our proposed benchmark pipeline. For each defense method, we systematically test different parameters to ensure a thorough and adequate evaluation. When defenses are applied to a VFL under a particular attack, all the hyper-parameters are not changed compared to that when that attack is applied. Detailed defense related experimental hyper-parameter settings are listed below:
\begin{itemize}
    \item In \textbf{L-DP} or \textbf{G-DP}~\citep{dwork2006DP,li2022label,zou2022defending}, gradients are first 2-norm clipped with $0.2$ when the defense is applied at the active party. When passive party applies this defense, noises are added to normalized local prediction results. Then noise with "diversity" scale parameter or standard deviation ranging from $0.0001$ to $1.0$ is added to clipped gradients.

    \item In \textbf{GS}~\citep{aji2017sparse,fu2021label,zou2022defending}, drop rate ranging from $95.0\%$ to $99.5\%$ are evaluated in the experiments. 
    
    \item In \textbf{GPer}~\citep{yang2022differentially}, we conduct tests with various standard deviations for the noise variable $u$ with the value of range $0.0001$ to $0.1$.
    
    \item In \textbf{dCor}~\citep{sun2022label,vepakomma2019reducing}, we set distance correlation regularizer coefficient $\lambda$ from $0.0001$ to $0.3$ and record the corresponding results.
    
    \item In \textbf{CAE}~\citep{zou2022defending}, we consider various confusional levels with $\lambda_2$ value ranging from $0.0$ to $1.0$  while in \textbf{DCAE}~\citep{zou2023mutual}, we select the same set of confusional levels and fix the bin number for gradient discrete to $12$.
    
    \item In \textbf{MID}~\citep{zou2023mutual}, we test the scenario where both active party and passive party apply MID, which suit the real world application since both sides need to protect themselves against potential attacks. Hyper-parameter $\lambda$ that signifies the strength of the defense ranges from $0.0$ to $1000.0$. 
\end{itemize}

\subsection{C-DCS Calculation Detail} \label{subsec:appendix_cdcs_calculation_detail}
T-DCS values and C-DCS values are all calculated with respect to $1$ single dataset. All T-DCS scores, i.e. $T\text{-}DCS_{LI_{2}}, T\text{-}DCS_{LI_{10}}, T\text{-}DCS_{LI_{5}}, T\text{-}DCS_{LI}, T\text{-}DCS_{FR}, T\text{-}DCS_{TB}, T\text{-}DCS_{NTB}$ are calculated by averaging the DCS values of each evaluated attack belonging to that type under the particular dataset. Then C-DCS values are calculated with \cref{eq:c-dcs} using only $T\text{-}DCS_{LI}, T\text{-}DCS_{FR}, T\text{-}DCS_{TB}, T\text{-}DCS_{NTB}$.

\section{Additional Experimental Results} \label{sec:appendix_results}

\subsection{Additional Main Task Performance Results} \label{subsec:appendix_MP}
We place the comparison results of MP and communication rounds (\#Rounds) of NN-based VFL with different local models in \cref{tab:different_local_models_MP} here in the appendix. Together with \cref{tab:tree_MP}, we compare the performance of linear regression, tree and neural network model architecture under $2$ different datasets. 
% \yang{if possible, add another dataset to show that tree model is typically used for tabular data. Current conclusion is not very strong here.} %Analysis has already been included in \cref{subsec:MP}.
% We also include the real-world dataset results in \cref{tab:real_world_dataset_MP}. %with analysis already included in \cref{subsec:MP} as well.}}

\begin{table}[!htb]
\caption{Comparison of MP and communication rounds using different local models in NN-based VFL with FedSGD communication protocol. We mainly follow~\citep{ye2022feature} for the selection of MLP model architecture.}
\label{tab:different_local_models_MP}
\resizebox{0.998\linewidth}{!}{
    \centering
    \begin{tabular}{c|c||c|c||c|c}
    \toprule
    \multirow{2}{*}{} & \multirow{2}{*}{} & \multicolumn{2}{c||}{Credit} & \multicolumn{2}{c}{Nursery} \\
    \cline{3-6}
    \\[-1em]
    ~ & ~ & Linear Regression & Neural Network (MLP-4) & Linear Regression & Neural Network (MLP-3) \\
    \midrule
    \multirow{2}{*}{aggVFL} & MP & 0.820$\pm$0.001 & 0.826$\pm$0.001 & 0.938$\pm$0.001 & 0.999$\pm$0.001  \\
    ~ & \#Rounds & 22$\pm$10 & 56$\pm$20 & 58$\pm$19 & 77$\pm$13 \\
    \midrule
    \multirow{2}{*}{splitVFL} & MP & 0.821$\pm$0.000 & 0.826$\pm$0.001 & 0.931$\pm$0.005 & 0.999$\pm$0.001 \\
    ~ & \#Rounds & 21$\pm$6 & 85$\pm$20 & 52$\pm$4 & 102$\pm$19 \\
    \bottomrule
    \end{tabular}
}
\vspace{-1em}
\end{table}

Table~\ref{tab:result-paillier-lr} presents the MP and running times with and without encryption. The number of epochs for Credit with LR is reduced to 40 since the training with HE takes a lot of time. The MP is different from those of Table~\ref{tab:different_local_models_MP} since the experiment of Table~\ref{tab:result-paillier-lr} uses the logit-loss even for multi-class classification. In terms of MP, a slight performance degradation occurs due to the approximation error caused by Taylor Expansion. The increase in running time due to encryption is more than 1000 times, as we do not currently support GPU for matrix operations on encrypted values.

\begin{table}[!htb]
\caption{MP and execution time under aggVFL with Homomorphic Encryption.} 
\label{tab:result-paillier-lr}
\centering
\resizebox{0.7\linewidth}{!}{
\centering
\begin{tabular}{@{}c|c|c|c@{}}
\toprule
Dataset                  &                    & Linear Regression (w/o encryption) & Linear Regression (w/ encryption)  \\ \midrule
\multirow{2}{*}{Credit}  & MP                & 0.821 $\pm$ 0.000   & 0.805 $\pm$ 0.002      \\
                         & Exec.Time {[}s{]} & 3 $\pm$ 0           & 11149 $\pm$ 357 \\ \midrule
\multirow{2}{*}{Nursery} & MP                & 0.918 $\pm$ 0.001   & 0.912 $\pm$ 0.001      \\
                         & Exec.Time {[}s{]} & 1 $\pm$ 0           & 2971 $\pm$ 57  \\ \bottomrule
\end{tabular}
}
\vspace{-1em}
\end{table}

% \subsection{Additional Detail Analysis of Attack and Defense Performance}

\subsection{Additional Attack and Defense Performance with More Datasets} \label{subsec:appendix_attack_defense_performance}
\begin{figure}[!htb]
  \centering
    \includegraphics[width=1.0\linewidth]{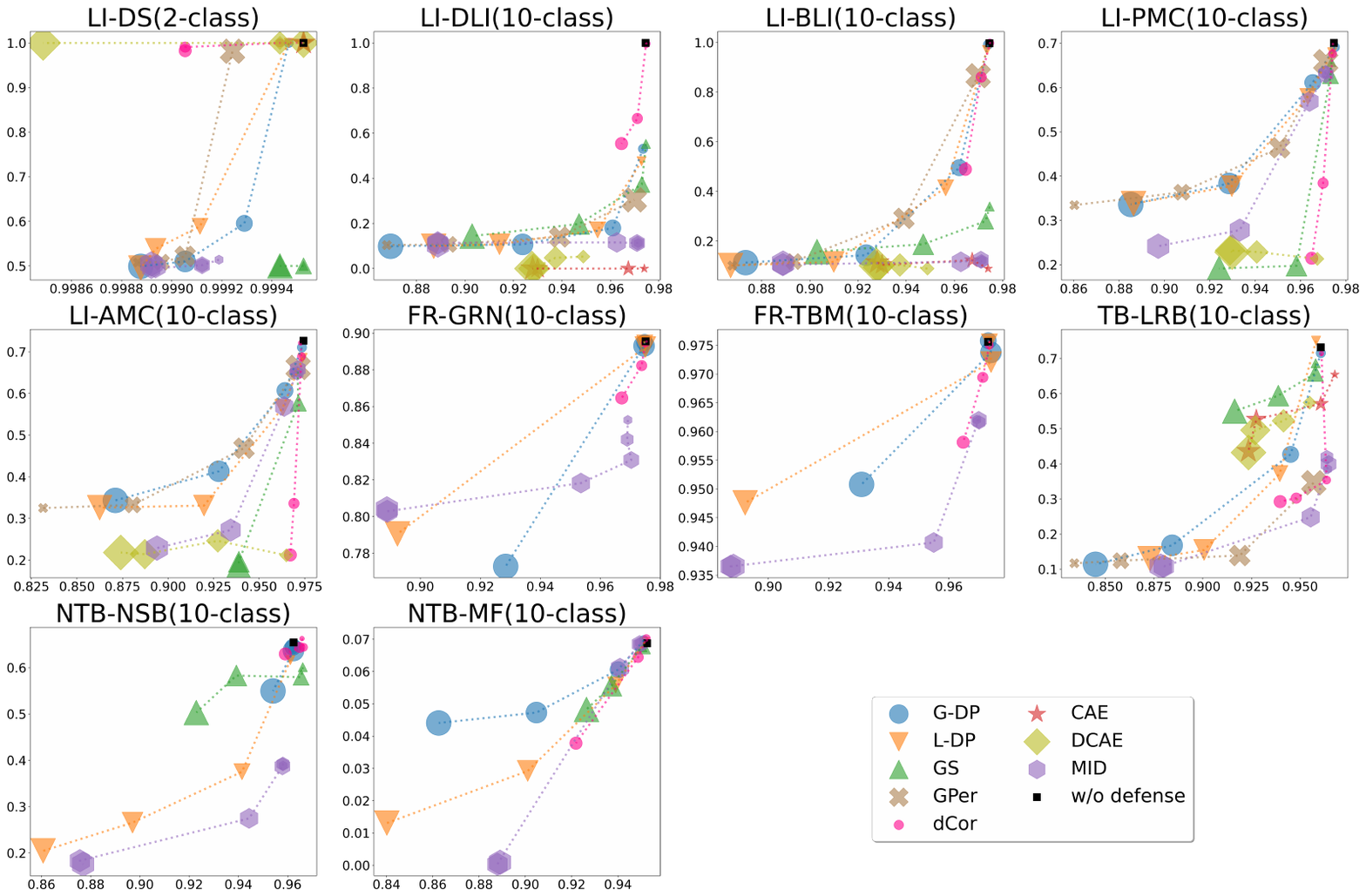}
  \vspace{-1em}
  \caption{MPs and APs for different attacks under defenses [MNIST dataset, aggVFL, FedSGD]}
  \label{fig:mnist_MPAP}
\end{figure}

\begin{figure}[!htb]
  \centering
    \includegraphics[width=1.0\linewidth]{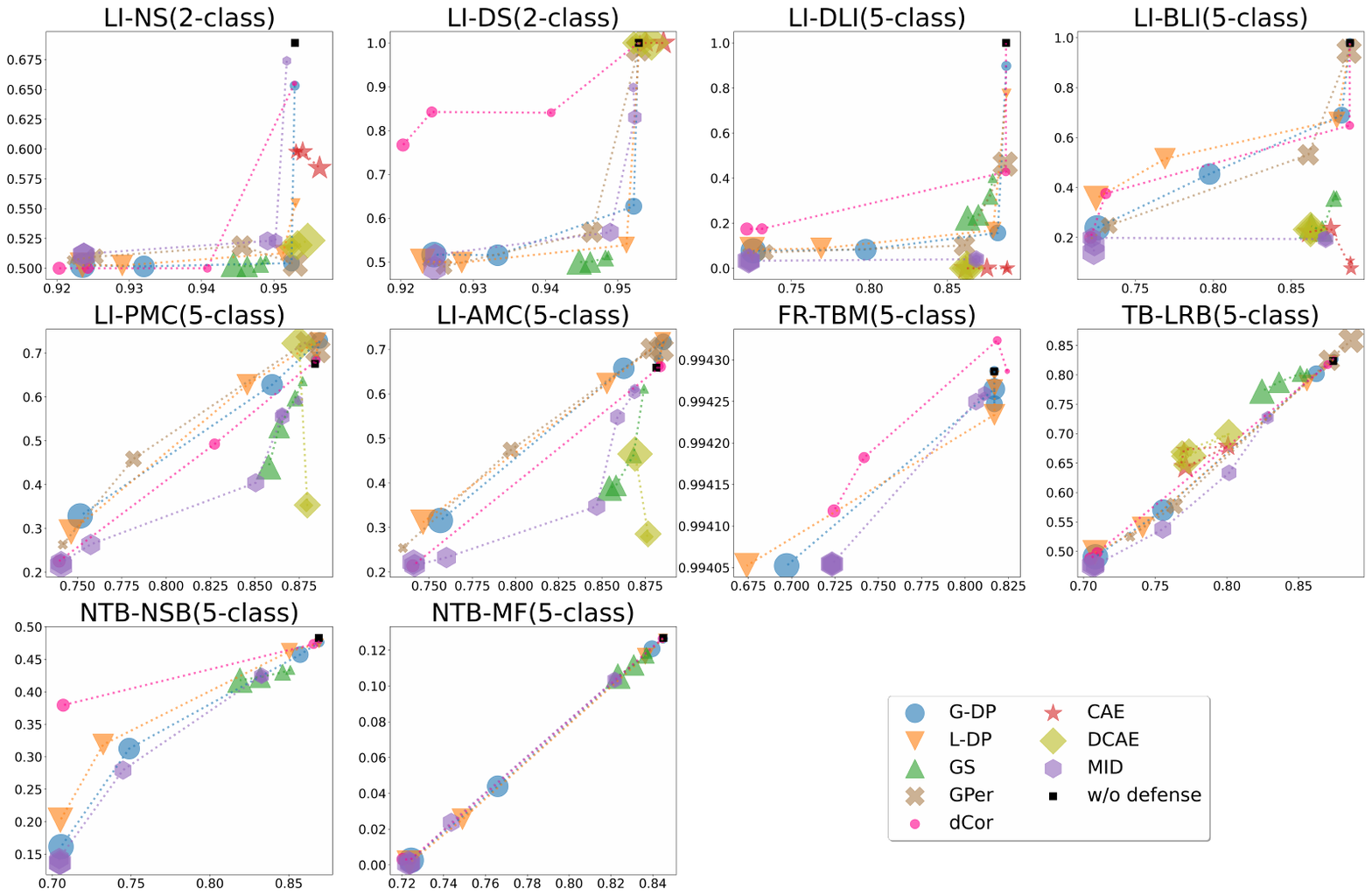}
    \vspace{-1em}
  \caption{MPs and APs for different attacks under defenses [NUSWIDE dataset, aggVFL, FedSGD]}
\label{fig:nuswide_MPAP}
\end{figure}

Due to space limit, we place the plot for MPs and APs under MNIST (\cref{fig:mnist_MPAP}) and NUSWIDE (\cref{fig:nuswide_MPAP}) datasets in this section with detailed analysis of the results. A brief summary of the below analysis is already included in \cref{subsec:attack_defense_performance}.

\subsubsection{MP and AP}
\underline{\textit{\textbf{MID}}} defense is capable of achieving a relatively lower AP while maintaining a similar MP compared to most other defenses, demonstrating its effectiveness in defending against a wide spectrum of attacks by reducing the information of label $Y$ and local feature $X_p$ kept in $H_p$ with a mutual information (MI) regularizer. % \tianyuan{There are a few exceptions in FR attacks under CIFAR10 dataset likely due to fact that ...} 

\underline{\textit{\textbf{dCor}}} targets at attacks that explore the information contained in $H_p$ for deducing $Y$ or $X_p$ by limiting the distance correlation between $H_p$ and $Y$ or $X_p$ for defending against attacks launched by passive or active party respectively. This defense ideology is similar to that of MID, which directly regularized the mutual information between $Y$ and $H_p$. %resulting in slightly different performances under some cases. 
MID performs better than dCor on gradient-based LI attacks like DS, DLI and BLI, likely due to the fact that limiting the MI between $H_p$ and $Y$ simultaneously limits the dependency between the correlated gradient $g_p$ at passive party and $Y$, which is better than reducing the distance correlation between $H_p$ and $Y$. Additionally, when compared to other attacks, dCor appears less effective in limiting AP under NTB attacks across all the $3$ datasets which is reasonable since this defense is not designed for defending against attacks that introduce information loss into transmitted data.

\underline{\textit{\textbf{CAE}} and \textit{\textbf{DCAE}}} focus on disguising label by mapping real one-hot labels to soft-fake labels. CAE consistently performs well across all datasets when defending against DLI, BLI and LRB attacks, which utilize merely the information of the current sample without any auxiliary information or data. For PMC and AMC attacks that rely on auxiliary label information, DCAE is more effective with the help of information reduction from quantization of gradients. The limited effectiveness of DCAE against DS attack can be attributed to the fact that the quantization is not effective in perturbing the direction of the gradient, while DS attack relies on the cosine similarity between the gradient of each sample and a known sample. These $2$ attacks are not designed for FR and NTB attacks. 

\underline{\textit{\textbf{GS}}} is a defense that injects information loss by sparsifying gradients, i.e. setting gradient elements with small absolute value to $0.0$, during training. As it is the gradients, which is directly related to labels but not input features, that GS modifies to defend against potential attacks, we do not benchmark its performance on FR attacks following previous works~\citep{luo2021feature,li2022ressfl}.
GS shows a strong defense ability for most of the LI attacks but exhibits less than satisfactory defense results against LRB attack, which replace the gradients of selected samples with those related to target labels to alter their labels. This is likely due to the fact that GS still preserves the gradient elements with larger absolute values which are responsible for learning the triggered patterns injected by the backdoor attack. 

\underline{\textit{\textbf{DP-G}} and \textit{\textbf{DP-L}}} are defenses that inject noise to gradients or local model output to defend against attacks launched by passive or active party respectively. They show similar trend and display a clear trade off between MP and AP in all the attacks. 

\underline{\textit{\textbf{GPer}}} targets at defending label inference attacks so we apply it only on label related attacks, i.e. LI and LRB attacks. GPer guarantees label-DP by perturbing the gradients and performs similarly to DP-G and DP-L in defending against these attacks, consistent with the original work~\citep{yang2022differentially} which shows that GPer performs the same as adding isotropical noise to gradients.

\subsubsection{Visualization of TBM Attack}\label{subsec:visualization_tbm}
\begin{figure}[!htb]
  \centering
    \includegraphics[width=0.99\linewidth]{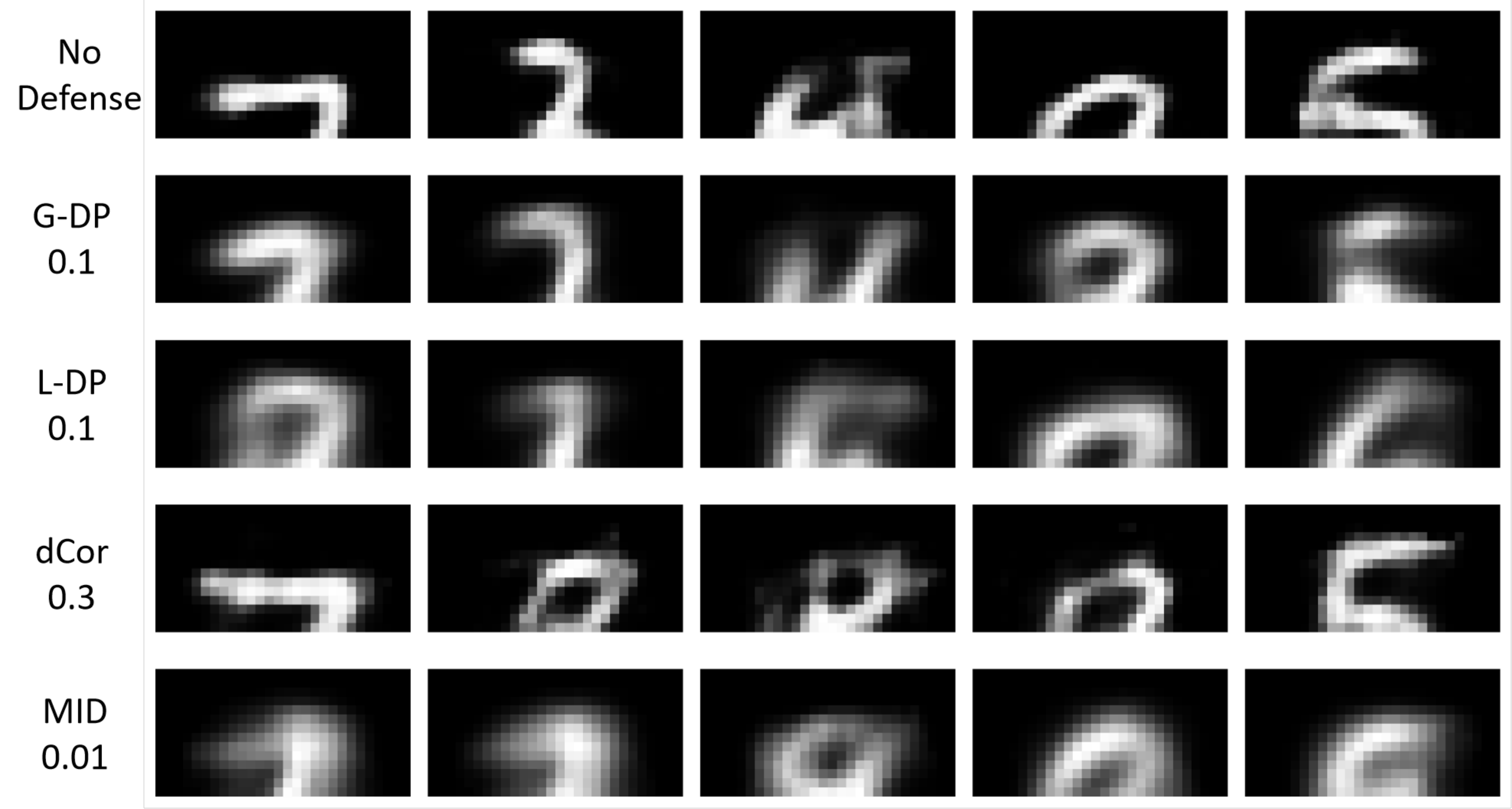}
    \vspace{-1em}
  \caption{Visualization of the feature reconstruction results of TBM attack for MNIST dataset. All the defense hyper-parameters selected are the most effective ones with highest DCS among the particular defense method.}
\label{fig:visualization_mnist_tbm}
\end{figure}

We show the visualization of reconstructed features of TBM attack under various defenses with MNIST dataset in \cref{fig:visualization_mnist_tbm}. All the defense hyper-parameters selected are the most effective one with highest DCS among the particular defense method, demonstrated in \cref{fig:mnist_MPAP} (sub-figure for TBM) and \cref{tab:mnist_dcs_ranking}. It's clear from \cref{fig:visualization_mnist_tbm} that, although the reconstructed images are noisy when defense exits, the contour can still be seen easily. This implies the difficulty for defending against TBM attack which exploits auxiliary i.i.d. data for feature reconstruction guidance.

\subsubsection{DCS, T-DCS and C-DCS ranking}
We also place the full T-DCS and C-DCS table for MNIST (\cref{tab:mnist_dcs_ranking}), CIFAR10 (\cref{tab:cifar10_dcs_ranking}), NUSWIDE (\cref{tab:nuswide_dcs_ranking}) datasets here due to limit of space. Basic analysis has already been included in \cref{subsec:attack_defense_performance}. %Note that, %although the top ranking defenses for each attack type is different for CIFAR10 compared to the other $2$ datasets, 
The overall relative rankings of all the defenses are still similar to that of other datasets.

\begin{table}[!h]
% \scriptsize    
\caption{T-DCS and C-DCS for All Defenses [MNIST dataset, aggVFL, FedSGD]}
\label{tab:mnist_dcs_ranking}
\resizebox{0.998\linewidth}{!}{
  \centering
   \begin{tabular}{cc|cccccc|c|c}
    \toprule
    \textbf{\makecell{Defense\\Name}} & \textbf{\makecell{Defense\\Parameter}} & \bm{$T\text{-}DCS_{LI_{2}}$} &
    \bm{$T\text{-}DCS_{LI_{10}}$} & \bm{$T\text{-}DCS_{LI}$} & \bm{$T\text{-}DCS_{FR}$} & \bm{$T\text{-}DCS_{TB}$} & \bm{$T\text{-}DCS_{NTB}$} & \bm{$C\text{-}DCS$} & \textbf{Ranking} \\ 
    \midrule 
       MID & 100 & 0.7371  & 0.8810  & 0.8523  & 0.6184  & 0.9070  & \textbf{0.9093}  & 0.8217  & 1  \\ 
        MID & 1.0 & 0.7395  & 0.8731  & 0.8464  & 0.6186  & \textbf{0.9080}  & 0.9070  & 0.8200  & 2  \\ 
        MID & 10000 & 0.6989  & 0.8939  & 0.8549  & 0.6145  & 0.9071  & 0.9026  & 0.8197  & 3  \\ 
        L-DP & 0.1 & 0.7407  & 0.8566  & 0.8334  & 0.6190  & 0.8957  & 0.8855  & 0.8084  & 4  \\ 
        MID & 0.1 & 0.6937  & 0.8577  & 0.8249  & 0.6205  & 0.8964  & 0.8885  & 0.8076  & 5  \\ 
        L-DP & 0.01 & 0.7244  & 0.8577  & 0.8310  & 0.6029  & 0.8920  & 0.8918  & 0.8044  & 6  \\ 
        MID & 0.01 & \textbf{0.7411}  & 0.8185  & 0.8031  & 0.6169  & 0.8504  & 0.8948  & 0.7913  & 7  \\ 
        G-DP & 0.1 & 0.7392  & 0.8528  & 0.8301  & \textbf{0.6220}  & 0.8913  & 0.8208  & 0.7910  & 8  \\ 
        G-DP & 0.01 & 0.7350  & 0.8461  & 0.8238  & 0.6026  & 0.8813  & 0.8167  & 0.7811  & 9  \\ 
        MID & 0.0 & 0.7335  & 0.8058  & 0.7913  & 0.6095  & 0.7862  & 0.8676  & 0.7636  & 10  \\ 
        MID & 1e-06 & 0.7382  & 0.8048  & 0.7914  & 0.6125  & 0.7798  & 0.8680  & 0.7629  & 11  \\ 
        MID & 1e-08 & 0.7382  & 0.8063  & 0.7927  & 0.6110  & 0.7714  & 0.8669  & 0.7605  & 12  \\ 
        MID & 0.0001 & 0.6564  & 0.8037  & 0.7742  & 0.6168  & 0.7720  & 0.8653  & 0.7571  & 13  \\ 
        L-DP & 0.001 & 0.7061  & 0.7720  & 0.7588  & 0.6022  & 0.7911  & 0.8725  & 0.7562  & 14  \\ 
        dCor & 0.3 & 0.5901  & 0.7999  & 0.7579  & 0.6083  & 0.8277  & 0.8241  & 0.7545  & 15  \\ 
        dCor & 0.1 & 0.5881  & 0.7241  & 0.6969  & 0.6046  & 0.8234  & 0.8203  & 0.7363  & 16  \\ 
        G-DP & 0.001 & 0.7038  & 0.7564  & 0.7458  & 0.6021  & 0.7677  & 0.8200  & 0.7339  & 17  \\ 
        dCor & 0.01 & 0.5858  & 0.6305  & 0.6216  & 0.6022  & 0.7999  & 0.8189  & 0.7106  & 18  \\ 
        L-DP & 0.0001 & 0.5858  & 0.6727  & 0.6553  & 0.6020  & 0.6533  & 0.8238  & 0.6836  & 19  \\ 
        G-DP & 0.0001 & 0.5858  & 0.6632  & 0.6477  & 0.6026  & 0.6640  & 0.8196  & 0.6835  & 20  \\ 
        dCor & 0.0001 & 0.5858  & 0.6258  & 0.6178  & 0.6020  & 0.6645  & 0.8161  & 0.6751  & 21  \\ 
        % None & None & 0.5858  & 0.6255  & 0.6175  & 0.6020  & 0.6589  & 0.8176  & 0.6740  & 22  \\
        \hline
        \\[-1em]
        GS & 99.5 & 0.7388  & 0.8873  & \textbf{0.8576}  & - & 0.7189  & 0.8454  & - & - \\ 
        GS & 99.0 & 0.7388  & 0.8780  & 0.8502  & - & 0.7038  & 0.8313  & - & - \\ 
        GS & 97.0 & 0.7388  & 0.7573  & 0.7536  & - & 0.6825  & 0.8305  & - & - \\ 
        GS & 95.0 & 0.7388  & 0.7232  & 0.7263  & - & 0.6732  & 0.8268  & - & - \\ 
        CAE & 1.0 & 0.5858  & 0.7960  & 0.7539  & - & 0.7632  & - & - & - \\ 
        CAE & 0.1 & 0.5858  & 0.8144  & 0.7687  & - & 0.7124  & - & - & - \\ 
        CAE & 0.5 & 0.5858  & 0.7937  & 0.7521  & - & 0.7280  & - & - & - \\ 
        CAE & 0.0 & 0.5858  & 0.8218  & 0.7746  & - & 0.6837  & - & - & - \\ 
        DCAE & 1.0 & 0.5858  & 0.9021  & 0.8389  & - & 0.7648  & - & - & - \\ 
        DCAE & 0.5 & 0.5858  & 0.9039  & 0.8402  & - & 0.7396  & - & - & - \\ 
        DCAE & 0.1 & 0.5858  & 0.8990  & 0.8364  & - & 0.7299  & - & - & - \\ 
        DCAE & 0.0 & 0.5858  & \textbf{0.9094}  & 0.8447  & - & 0.7111  & - & - & - \\ 
        GPer & 10.0 & 0.5903  & 0.7016  & 0.6794  & - & 0.8031  & - & - & - \\ 
        GPer & 1.0 & 0.7306  & 0.8097  & 0.7939  & - & 0.9045  & - & - & - \\ 
        GPer & 0.1 & 0.7357  & 0.8546  & 0.8308  & - & 0.8926  & - & - & - \\ 
        GPer & 0.01 & 0.7360  & 0.8525  & 0.8292  & - & 0.8857  & - & - & - \\ 
    \bottomrule
    \end{tabular}
}
\end{table}

\begin{table}[!h]
% \scriptsize    
\caption{T-DCS and C-DCS for All Defenses [CIFAR10 dataset, aggVFL, FedSGD]}
\label{tab:cifar10_dcs_ranking}
\resizebox{0.998\linewidth}{!}{
  \centering
   \begin{tabular}{cc|cccccc|c|c}
    \toprule
    \textbf{\makecell{Defense\\Name}} & \textbf{\makecell{Defense\\Parameter}} & \bm{$T\text{-}DCS_{LI_{2}}$} &
    \bm{$T\text{-}DCS_{LI_{10}}$} & \bm{$T\text{-}DCS_{LI}$} & \bm{$T\text{-}DCS_{FR}$} & \bm{$T\text{-}DCS_{TB}$} & \bm{$T\text{-}DCS_{NTB}$} & \bm{$C\text{-}DCS$} & \textbf{Ranking} \\ 
    \midrule 
    MID & 0.01 & 0.7232  & 0.9172  & 0.8784  & 0.6035  & 0.8942  & 0.9286  & 0.8262  & 1 \\ 
    MID & 1.0 & 0.7260  & 0.9173  & \textbf{0.8791}  & 0.6039  & 0.8931  & 0.9286  & 0.8262  & 2 \\ 
    MID & 100 & 0.7270  & 0.9161  & 0.8783  & 0.6039  & 0.8931  & 0.9280  & 0.8258  & 3 \\ 
    MID & 10000 & 0.7276  & 0.9159  & 0.8782  & 0.6016  & 0.8945  & 0.9284  & 0.8257  & 4 \\ 
    MID & 0.1 & 0.7001  & \textbf{0.9175}  & 0.8740  & 0.6016  & 0.8906  & 0.9284  & 0.8237  & 5 \\ 
    MID & 1e-06 & 0.7132  & 0.8600  & 0.8306  & 0.6024  & 0.9123  & 0.9323  & 0.8194  & 6 \\ 
    MID & 1e-08 & 0.7132  & 0.8594  & 0.8301  & 0.6023  & 0.9123  & 0.9313  & 0.8190  & 7 \\ 
    MID & 0.0 & 0.7132  & 0.8552  & 0.8268  & 0.6023  & 0.9121  & 0.9289  & 0.8175  & 8 \\ 
    MID & 0.0001 & 0.7118  & 0.8575  & 0.8283  & 0.6034  & 0.9061  & 0.9302  & 0.8170  & 9 \\ 
    G-DP & 0.01 & 0.7316  & 0.8965  & 0.8635  & 0.6019  & 0.8809  & 0.9152  & 0.8154  & 10 \\ 
    L-DP & 0.01 & 0.7365  & 0.8987  & 0.8663  & 0.6021  & 0.8751  & 0.9146  & 0.8145  & 11 \\ 
    L-DP & 0.001 & 0.7041  & 0.8787  & 0.8437  & 0.6023  & 0.8892  & 0.9211  & 0.8141  & 12 \\ 
    G-DP & 0.1 & 0.7324  & 0.8947  & 0.8622  & \textbf{0.6097}  & 0.8670  & 0.9083  & 0.8118  & 13 \\ 
    G-DP & 0.001 & 0.6704  & 0.8713  & 0.8311  & 0.6023  & 0.8879  & 0.9241  & 0.8114  & 14 \\ 
    L-DP & 0.1 & 0.7353  & 0.8919  & 0.8606  & 0.6034  & 0.8696  & 0.9058  & 0.8098  & 15 \\ 
    dCor & 0.3 & 0.6116  & 0.8109  & 0.7710  & 0.6012  & 0.8883  & 0.9245  & 0.7963  & 16 \\ 
    dCor & 0.1 & 0.5879  & 0.7695  & 0.7332  & 0.6030  & 0.8962  & 0.9175  & 0.7875  & 17 \\ 
    L-DP & 0.0001 & 0.5876  & 0.7611  & 0.7264  & 0.6023  & 0.9049  & 0.9083  & 0.7855  & 18 \\ 
    G-DP & 0.0001 & 0.5868  & 0.7487  & 0.7163  & 0.6025  & 0.9046  & 0.8993  & 0.7807  & 19 \\ 
    dCor & 0.0001 & 0.5858  & 0.6578  & 0.6434  & 0.6031  & 0.9162  & \textbf{0.9558}  & 0.7796  & 20 \\ 
    dCor & 0.01 & 0.5859  & 0.6839  & 0.6643  & 0.6042  & 0.9162  & 0.8982  & 0.7708  & 21 \\ 
    % None & None & 0.5858  & 0.6577  & 0.6433  & 0.6026  & 0.9095  & 0.8944  & 0.7624  & 22 \\ 
    \hline
    \\[-1em]
    GS & 99.5 & 0.7387  & 0.8956  & 0.8642  & - & 0.8877  & 0.9276  & - & - \\ 
    GS & 99.0 & 0.7388  & 0.8810  & 0.8526  & - & 0.8888  & 0.9282  & - & - \\ 
    GS & 97.0 & 0.7388  & 0.8367  & 0.8171  & - & 0.8882  & 0.9286  & - & - \\ 
    GS & 95.0 & 0.7388  & 0.8049  & 0.7916  & - & 0.8895  & 0.9295  & - & - \\ 
    CAE & 1.0 & 0.5918  & 0.8592  & 0.8057  & - & 0.8482  & - & - & - \\ 
    CAE & 0.5 & 0.5860  & 0.8633  & 0.8079  & - & 0.8557  & - & - & - \\ 
    CAE & 0.1 & 0.5858  & 0.8498  & 0.7970  & - & 0.9179  & - & - & - \\ 
    CAE & 0.0 & 0.5858  & 0.8419  & 0.7907  & - & \textbf{0.9214}  & - & - & - \\ 
    DCAE & 1.0 & 0.6080  & 0.8678  & 0.8159  & - & 0.8464  & - & - & - \\ 
    DCAE & 0.5 & 0.6131  & 0.8780  & 0.8250  & - & 0.8556  & - & - & - \\ 
    DCAE & 0.1 & 0.6090  & 0.8977  & 0.8400  & - & 0.9039  & - & - & - \\ 
    DCAE & 0.0 & 0.6090  & 0.9038  & 0.8449  & - & 0.9066  & - & - & - \\ 
    GPer & 10.0 & 0.6422  & 0.8082  & 0.7750  & - & 0.8937  & - & - & - \\ 
    GPer & 1.0 & 0.6976  & 0.8913  & 0.8526  & - & 0.8838  & - & - & - \\ 
    GPer & 0.1 & 0.7385  & 0.8946  & 0.8634  & - & 0.8715  & - & - & - \\ 
    GPer & 0.01 & \textbf{0.7392}  & 0.8931  & 0.8623  & - & 0.8700  & - & - & - \\ 
    \bottomrule
    \end{tabular}
}
\end{table}

\begin{table}[!h]
% \scriptsize    
\caption{T-DCS and C-DCS for All Defenses [NUSWIDE dataset, aggVFL, FedSGD]}
\label{tab:nuswide_dcs_ranking}
\resizebox{0.998\linewidth}{!}{
  \centering
   \begin{tabular}{cc|cccccc|c|c}
    \toprule
    \textbf{\makecell{Defense\\Name}} & \textbf{\makecell{Defense\\Parameter}} & \bm{$T\text{-}DCS_{LI_{2}}$} &
    \bm{$T\text{-}DCS_{LI_{5}}$} & \bm{$T\text{-}DCS_{LI}$} & \bm{$T\text{-}DCS_{FR}$} & \bm{$T\text{-}DCS_{TB}$} & \bm{$T\text{-}DCS_{NTB}$} & \bm{$C\text{-}DCS$} & \textbf{Ranking} \\ 
    \midrule 
    MID  & 10000  & 0.7358 & 0.8559 & \textbf{0.8159} & 0.5833 & \textbf{0.7333} & 0.8707 & 0.7508 & 1  \\
    MID  & 1.0    & 0.7476 & 0.8472 & 0.8140 & 0.5833 & 0.7331 & 0.8700 & 0.7501 & 2  \\
    MID  & 100    & 0.7320 & 0.8536 & 0.8130 & 0.5833 & 0.7326 & \textbf{0.8711} & 0.7500 & 3  \\
    G-DP & 0.1    & 0.7375 & 0.8262 & 0.7966 & 0.5863 & 0.7282 & 0.8675 & 0.7447 & 4  \\
    L-DP & 0.1    & 0.7389 & 0.8177 & 0.7915 & 0.5863 & 0.7258 & 0.8603 & 0.7410 & 5  \\
    MID  & 0.1    & 0.7516 & 0.8259 & 0.8011 & 0.5833 & 0.7172 & 0.8563 & 0.7395 & 6  \\
    MID  & 0.01   & 0.7280 & 0.8092 & 0.7822 & 0.5844 & 0.7151 & 0.8627 & 0.7361 & 7  \\
    MID  & 0.0001 & 0.7144 & 0.8097 & 0.7779 & 0.5856 & 0.7040 & 0.8680 & 0.7339 & 8  \\
    dCor & 0.3    & \textbf{0.7641} & 0.8411 & 0.8155 & 0.5834 & 0.7289 & 0.8051 & 0.7332 & 9  \\
    G-DP & 0.01   & 0.7391 & 0.7600 & 0.7530 & 0.5863 & 0.7061 & 0.8549 & 0.7251 & 10 \\
    L-DP & 0.01   & 0.7395 & 0.7525 & 0.7482 & 0.5863 & 0.7148 & 0.8485 & 0.7244 & 11 \\
    MID  & 1e-06  & 0.7022 & 0.8201 & 0.7808 & 0.5860 & 0.6880 & 0.8408 & 0.7239 & 12 \\
    dCor & 0.1    & 0.7442 & 0.7617 & 0.7559 & 0.5841 & 0.7259 & 0.8279 & 0.7234 & 13 \\
    MID  & 1e-08  & 0.7066 & 0.8147 & 0.7787 & 0.5862 & 0.6593 & 0.8410 & 0.7163 & 14 \\
    MID  & 0.0    & 0.6599 & 0.8097 & 0.7598 & 0.5862 & 0.6590 & 0.8414 & 0.7116 & 15 \\
    L-DP & 0.001  & 0.7291 & 0.7234 & 0.7253 & 0.5863 & 0.6424 & 0.8329 & 0.6967 & 16 \\
    G-DP & 0.001  & 0.7175 & 0.7237 & 0.7216 & 0.5863 & 0.6379 & 0.8334 & 0.6948 & 17 \\
    dCor & 0.01   & 0.7445 & 0.7021 & 0.7162 & 0.5863 & 0.6336 & 0.8295 & 0.6914 & 18 \\
    L-DP & 0.0001 & 0.6783 & 0.6470 & 0.6574 & 0.5863 & 0.6313 & 0.8293 & 0.6761 & 19 \\
    G-DP & 0.0001 & 0.6495 & 0.6381 & 0.6419 & 0.5863 & 0.6309 & 0.8290 & 0.6720 & 20 \\
    dCor & 0.0001 & 0.6496 & 0.6340 & 0.6392 & \textbf{0.5864} & 0.6307 & 0.8287 & 0.6712 & 21 \\
    % None & None   & 0.6455 & 0.6339 & 0.6377 & 0.5863 & 0.6318 & 0.8276 & 0.6709 & 22 \\
    \hline
    \\[-1em]
    GS   & 99.5   & 0.7381 & 0.8142 & 0.7888 & -      & 0.6456 & 0.8415 & -      & -  \\
    GS   & 99.0   & 0.7404 & 0.8060 & 0.7841 & -      & 0.6415 & 0.8408 & -      & -  \\
    GS   & 97.0   & 0.7414 & 0.7672 & 0.7586 & -      & 0.6376 & 0.8392 & -      & -  \\
    GS   & 95.0   & 0.7423 & 0.7399 & 0.7407 & -      & 0.6375 & 0.8385 & -      & -  \\
    CAE  & 1.0    & 0.6863 & 0.7822 & 0.7502 & -      & 0.6830  & -      & -     & -   \\
    CAE  & 0.5    & 0.6808 & 0.7848 & 0.7501 & -      & 0.6733  & -      & -     & -   \\
    CAE  & 0.1    & 0.6808 & 0.8249 & 0.7768 & -      & 0.6734  & -      & -     & -   \\
    CAE  & 0.0    & 0.6808 & 0.8212 & 0.7744 & -      & 0.6807  & -      & -     & -   \\
    DCAE & 1.0    & 0.6716 & 0.8156 & 0.7676 & -      & 0.6771 & -      & -      & -  \\
    DCAE & 0.5    & 0.6672 & 0.8108 & 0.7629 & -      & 0.6668 & -      & -      & -  \\
    DCAE & 0.1    & 0.6669 & 0.8651 & 0.7991 & -      & 0.6746 & -      & -      & -  \\
    DCAE & 0.0    & 0.6669 & \textbf{0.8660} & 0.7996 & -      & 0.6816 & -      & -      & -  \\
    GPer & 10.0   & 0.6877 & 0.6722 & 0.6773 & -      & 0.6222 & -      & -      & -  \\
    GPer & 1.0    & 0.7230 & 0.7460 & 0.7383 & -      & 0.6315 & -      & -      & -  \\
    GPer & 0.1    & 0.7395 & 0.8007 & 0.7803 & -      & 0.7042 & -      & -      & - \\
    GPer & 0.01   & 0.7386 & 0.8412 & 0.8070 & -      & 0.7193 & -      & -      & -  \\
    \bottomrule
    \end{tabular}
}
\end{table}

\begin{table}[!h]
% \scriptsize    
\caption{T-DCS and C-DCS for All Defenses [MNIST dataset, splitVFL, FedSGD]}
\label{tab:mnist_split_dcs_ranking}
\resizebox{0.998\linewidth}{!}{
  \centering
   \begin{tabular}{cc|cccccc|c|c}
    \toprule
    \textbf{\makecell{Defense\\Name}} & \textbf{\makecell{Defense\\Parameter}} & \bm{$T\text{-}DCS_{LI_{2}}$} &
    \bm{$T\text{-}DCS_{LI_{10}}$} & \bm{$T\text{-}DCS_{LI}$} & \bm{$T\text{-}DCS_{FR}$} & \bm{$T\text{-}DCS_{TB}$} & \bm{$T\text{-}DCS_{NTB}$} & \bm{$C\text{-}DCS$} & \textbf{Ranking} \\ 
    \midrule
MID & 1.0 & 0.7380  & 0.8846  & 0.8553  & 0.6065  & 0.9051  & 0.9237  & 0.8226  & 1 \\ 
L-DP & 0.1 & 0.7398  & 0.8700  & 0.8439  & 0.6156  & 0.9092  & 0.9203  & 0.8222  & 2 \\ 
G-DP & 0.1 & 0.7350  & 0.8654  & 0.8393  & 0.6139  & 0.9058  & \textbf{0.9271}  & 0.8215  & 3 \\ 
MID & 10000 & 0.7372  & \textbf{0.8921}  & \textbf{0.8611 } & 0.6064  & 0.9071  & 0.9062  & 0.8202  & 4 \\ 
MID & 100 & 0.7371  & 0.8896  & 0.8591  & 0.6082  & 0.9085  & 0.9034  & 0.8198  & 5 \\ 
L-DP & 0.01 & 0.7381  & 0.8603  & 0.8359  & 0.6026  & 0.9099  & 0.9147  & 0.8158  & 6 \\ 
MID & 0.1 & 0.7398  & 0.8652  & 0.8401  & \textbf{0.6241}  & \textbf{0.9113}  & 0.8871  & 0.8156  & 7 \\ 
G-DP & 0.01 & 0.7393  & 0.8562  & 0.8328  & 0.6031  & 0.9060  & 0.9170  & 0.8147  & 8 \\ 
dCor & 0.3 & 0.5934  & 0.8876  & 0.8288  & 0.6194  & 0.9072  & 0.8732  & 0.8071  & 9 \\ 
L-DP & 0.001 & 0.7189  & 0.8365  & 0.8130  & 0.6038  & 0.8682  & 0.8663  & 0.7878  & 10 \\ 
MID & 0.01 & 0.7364  & 0.8455  & 0.8237  & 0.6140  & 0.8421  & 0.8536  & 0.7834  & 11 \\ 
G-DP & 0.001 & \textbf{0.7404}  & 0.8269  & 0.8096  & 0.6041  & 0.8109  & 0.8775  & 0.7755  & 12 \\ 
dCor & 0.1 & 0.5983  & 0.8519  & 0.8012  & 0.6133  & 0.8519  & 0.8212  & 0.7719  & 13 \\ 
MID & 1e-06 & 0.7388  & 0.8140  & 0.7990  & 0.6173  & 0.7641  & 0.8394  & 0.7549  & 14 \\ 
MID & 1e-08 & 0.7388  & 0.8158  & 0.8004  & 0.6135  & 0.7615  & 0.8421  & 0.7544  & 15 \\ 
MID & 0.0 & 0.7388  & 0.8147  & 0.7995  & 0.6146  & 0.7524  & 0.8456  & 0.7530  & 16 \\ 
dCor & 0.01 & 0.5870  & 0.8020  & 0.7590  & 0.6048  & 0.8111  & 0.8174  & 0.7481  & 17 \\ 
MID & 0.0001 & 0.7388  & 0.8137  & 0.7988  & 0.6182  & 0.7324  & 0.8312  & 0.7451  & 18 \\ 
L-DP & 0.0001 & 0.6472  & 0.8041  & 0.7727  & 0.6046  & 0.7699  & 0.8268  & 0.7435  & 19 \\ 
G-DP & 0.0001 & 0.6253  & 0.7972  & 0.7628  & 0.6036  & 0.7586  & 0.8212  & 0.7365  & 20 \\ 
dCor & 0.0001 & 0.5858  & 0.7906  & 0.7496  & 0.6053  & 0.7538  & 0.8164  & 0.7313  & 21 \\ 
\hline
\\[-1em]
GS & 95.0 & 0.7388  & 0.8080  & 0.7941  & - & 0.7380  & 0.8345  & -  & - \\ 
GS & 97.0 & 0.7388  & 0.8133  & 0.7984  & - & 0.7355  & 0.8428  & -  & - \\ 
GS & 99.0 & 0.7388  & 0.8334  & 0.8145  & - & 0.7590  & 0.8558  & -  & - \\ 
GS & 99.5 & 0.7388  & 0.8487  & 0.8267  & - & 0.7872  & 0.8636  & -  & - \\ 
CAE & 1.0 & 0.5858  & 0.8252  & 0.7773  & - & 0.7741  & - & -  & - \\ 
CAE & 0.5 & 0.5858  & 0.8282  & 0.7797  & - & 0.7439  & - & -  & - \\ 
CAE & 0.1 & 0.5858  & 0.8365  & 0.7863  & - & 0.7535  & - & -  & - \\ 
CAE & 0.0 & 0.5858  & 0.8410  & 0.7899  & - & 0.7574  & - & -  & - \\ 
DCAE & 1.0 & 0.5858  & 0.7824  & 0.7431  & - & 0.7882  & - & -  & - \\ 
DCAE & 0.5 & 0.5858  & 0.7862  & 0.7461  & - & 0.7882  & - & -  & - \\ 
DCAE & 0.1 & 0.5858  & 0.7773  & 0.7390  & - & 0.7882  & - & -  & - \\ 
DCAE & 0.0 & 0.5858  & 0.7758  & 0.7378  & - & 0.7882  & - & -  & - \\ 
GPer & 10.0 & 0.5916  & 0.8037  & 0.7613  & - & 0.8490  & - & -  & - \\ 
GPer & 1.0 & 0.7295  & 0.8460  & 0.8227  & - & 0.9064  & - & -  & - \\ 
GPer & 0.1 & 0.7369  & 0.8632  & 0.8379  & - & 0.9039  & - & -  & - \\ 
GPer & 0.01 & 0.7365  & 0.8712  & 0.8442  & - & 0.9025  & - & -  & - \\ 
    \bottomrule
    \end{tabular}
}
\end{table}

\begin{table}[!h]
% \scriptsize    
\caption{T-DCS and C-DCS for All Defenses [MNIST dataset, aggVFL, \textbf{FedBCD(Q=5)}]}
\label{tab:mnist_fedbcd_dcs_ranking}
\resizebox{0.998\linewidth}{!}{
  \centering
   \begin{tabular}{cc|cccccc|c|c}
    \toprule
    \textbf{\makecell{Defense\\Name}} & \textbf{\makecell{Defense\\Parameter}} & \bm{$T\text{-}DCS_{LI_{2}}$} &
    \bm{$T\text{-}DCS_{LI_{10}}$} & \bm{$T\text{-}DCS_{LI}$} & \bm{$T\text{-}DCS_{FR}$} & \bm{$T\text{-}DCS_{TB}$} & \bm{$T\text{-}DCS_{NTB}$} & \bm{$C\text{-}DCS$} & \textbf{Ranking} \\ 
    \midrule
    MID & 1.0 & 0.6707  & 0.8794  & 0.8376  & 0.6166  & 0.9034  & 0.9397  & 0.8243  & 1  \\ 
    MID & 10000 & 0.6917  & 0.8729  & 0.8366  & 0.6133  & 0.9034  & \textbf{0.9417}  & 0.8238  & 2  \\ 
    MID & 100 & 0.6908  & 0.8750  & 0.8381  & 0.6087  & \textbf{0.9039}  & 0.9347  & 0.8214  & 3  \\ 
    MID & 0.1 & 0.6121  & 0.8730  & 0.8208  & 0.6199  & 0.9030  & 0.9413  & 0.8212  & 4  \\ 
    L-DP & 0.1 & 0.7403  & 0.8369  & 0.8176  &\textbf{ 0.6525 } & 0.8669  & 0.8506  & 0.7969  & 5  \\ 
    G-DP & 0.1 & 0.7385  & 0.8402  & 0.8199  & 0.6262  & 0.8704  & 0.8532  & 0.7924  & 6  \\ 
    G-DP & 0.01 & \textbf{0.7404 } & 0.8376  & 0.8182  & 0.6037  & 0.8976  & 0.8464  & 0.7915  & 7  \\ 
    L-DP & 0.01 & 0.7387  & 0.8331  & 0.8142  & 0.6055  & 0.8959  & 0.8472  & 0.7907  & 8  \\ 
    MID & 0.01 & 0.6119  & 0.8357  & 0.7910  & 0.6166  & 0.8075  & 0.9118  & 0.7817  & 9  \\ 
    L-DP & 0.001 & 0.6821  & 0.7915  & 0.7696  & 0.6033  & 0.8447  & 0.8668  & 0.7711  & 10  \\ 
    G-DP & 0.001 & 0.7132  & 0.7836  & 0.7696  & 0.6051  & 0.8032  & 0.8527  & 0.7576  & 11  \\ 
    dCor & 0.3 & 0.5900  & 0.8329  & 0.7843  & 0.6071  & 0.8000  & 0.8154  & 0.7517  & 12  \\ 
    MID & 1e-06 & 0.6156  & 0.8089  & 0.7702  & 0.6144  & 0.7441  & 0.8727  & 0.7504  & 13  \\ 
    MID & 0.0001 & 0.6155  & 0.8110  & 0.7719  & 0.6217  & 0.7166  & 0.8753  & 0.7464  & 14  \\ 
    MID & 0.0 & 0.6156  & 0.8072  & 0.7689  & 0.6176  & 0.7396  & 0.8542  & 0.7451  & 15  \\ 
    MID & 1e-08 & 0.6156  & 0.8092  & 0.7705  & 0.6224  & 0.7237  & 0.8522  & 0.7422  & 16  \\ 
    dCor & 0.1 & 0.5960  & 0.7971  & 0.7569  & 0.6077  & 0.6955  & 0.8115  & 0.7179  & 17  \\ 
    L-DP & 0.0001 & 0.5858  & 0.7246  & 0.6969  & 0.6044  & 0.7083  & 0.8241  & 0.7084  & 18  \\ 
    G-DP & 0.0001 & 0.5858  & 0.7194  & 0.6927  & 0.6034  & 0.6963  & 0.8196  & 0.7030  & 19  \\ 
    dCor & 0.01 & 0.5858  & 0.6900  & 0.6691  & 0.6052  & 0.6894  & 0.8121  & 0.6940  & 20  \\ 
    dCor & 0.0001 & 0.5858  & 0.6763  & 0.6582  & 0.6055  & 0.6714  & 0.8086  & 0.6859  & 21  \\ 
    \hline
    \\[-1em]
    GS & 99.5 & 0.7388  & 0.9005  & 0.8682  & - & 0.7086  & 0.8587  & - & - \\ 
    GS & 99.0 & 0.7388  & 0.8896  & 0.8594  & - & 0.7093  & 0.8506  & - & - \\ 
    GS & 97.0 & 0.7388  & 0.7718  & 0.7652  & - & 0.6683  & 0.8483  & - & - \\ 
    GS & 95.0 & 0.7388  & 0.7458  & 0.7444  & - & 0.6585  & 0.8359  & - & - \\ 
    CAE & 1.0 & 0.5858  & 0.9319  & 0.8627  & - & 0.7450  & - & - & - \\ 
    CAE & 0.5 & 0.5858  & 0.9367  & 0.8665  & - & 0.8109  & - & - & - \\ 
    CAE & 0.1 & 0.5858  & 0.9664  & 0.8903  & - & 0.6757  & - & - & - \\ 
    CAE & 0.0 & 0.5858  & \textbf{0.9681 } & \textbf{0.8917}  & - & 0.6724  & - & - & - \\ 
    DCAE & 1.0 & 0.5858  & 0.8734  & 0.8159  & - & 0.7311  & - & - & - \\ 
    DCAE & 0.5 & 0.5858  & 0.8776  & 0.8192  & - & 0.8041  & - & - & - \\ 
    DCAE & 0.1 & 0.5858  & 0.8867  & 0.8265  & - & 0.6619  & - & - & - \\ 
    DCAE & 0.0 & 0.5858  & 0.8919  & 0.8307  & - & 0.6896  & - & - & - \\ 
    GPer & 10.0 & 0.5901  & 0.7562  & 0.7230  & - & 0.7515  & - & - & - \\ 
    GPer & 1.0 & 0.7317  & 0.8216  & 0.8036  & - & 0.8848  & - & - & - \\ 
    GPer & 0.1 & 0.7341  & 0.8398  & 0.8186  & - & 0.8829  & - & - & - \\ 
    GPer & 0.01 & 0.7366  & 0.8381  & 0.8178  & - & 0.8683  & - & - & - \\ 
    \bottomrule
    \end{tabular}
}
\end{table}

\begin{figure}[htbp]             
  \centering
  \includegraphics[width=0.60\linewidth]{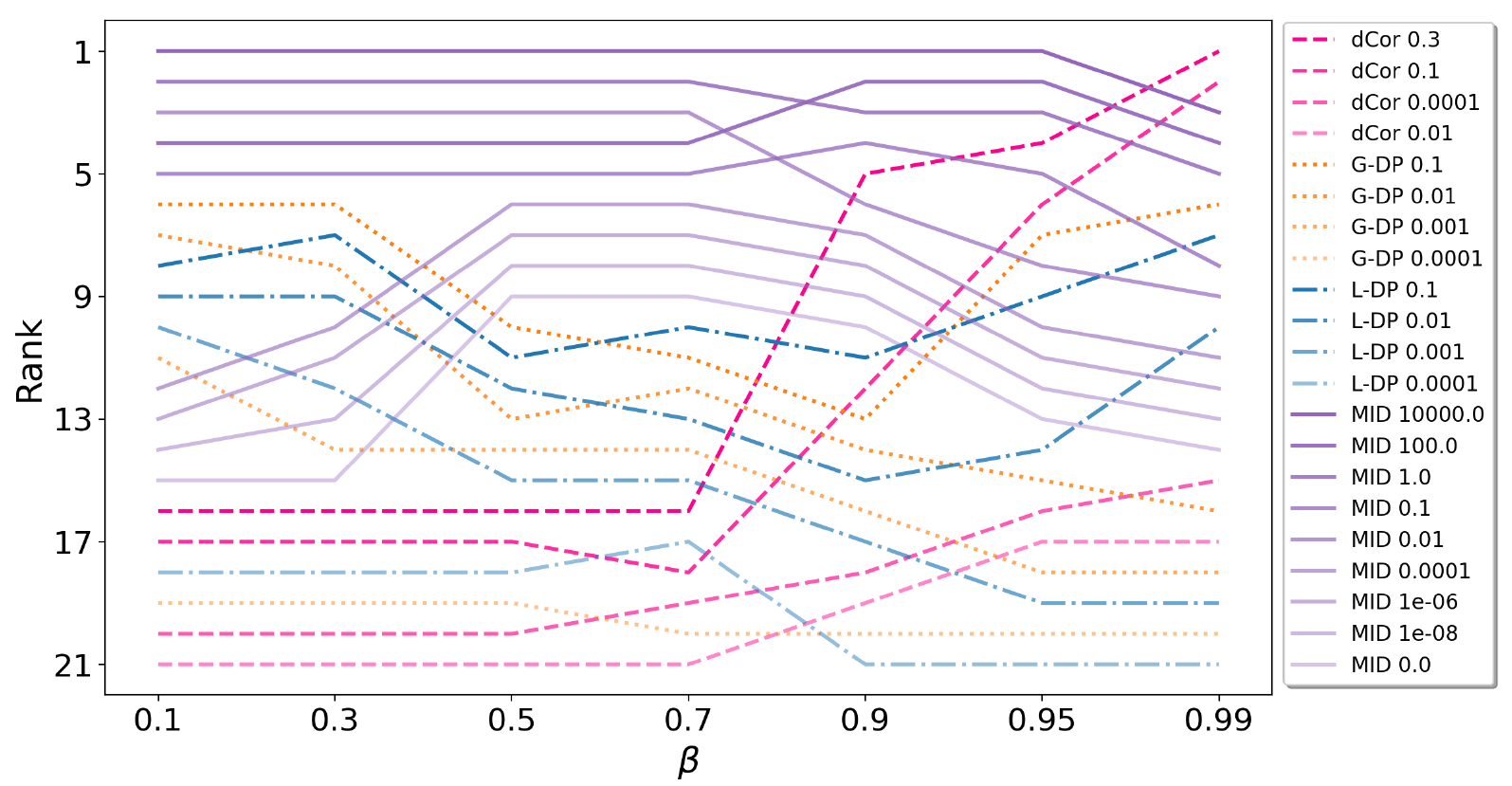}
  \vspace{-1em}
  \caption{Change of C-DCS ranking with the change of $\beta$. [CIFAR10 dataset, aggVFL, FedSGD]}
  \label{fig:change_beta_dcs_cifar10}           
\end{figure}

\begin{figure}[htbp]             
  \centering
  \includegraphics[width=0.60\linewidth]{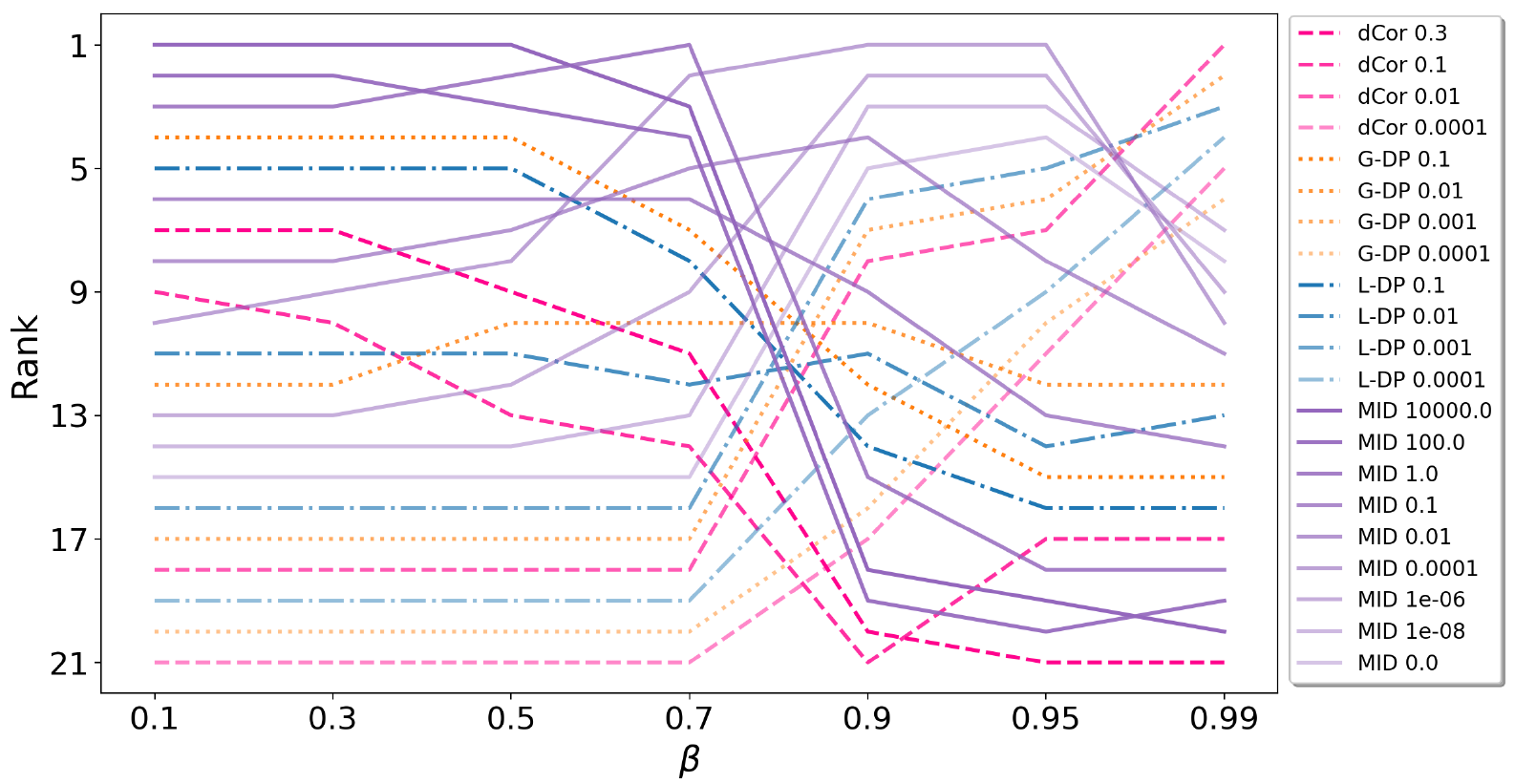}
  \vspace{-1em}
  \caption{Change of C-DCS ranking with the change of $\beta$. [NUSWIDE dataset, aggVFL, FedSGD]}
  \label{fig:change_beta_dcs_nuswide}           
\end{figure}

We also show the change in C-DCS ranking under different $\beta$ for CIFAR10 and NUSWIDE here due to space limit. Similar results can be seen from \cref{fig:change_beta_dcs_cifar10,fig:change_beta_dcs_nuswide} while the ranking fluctuates more under NUSWIDE dataset compared to the other $2$ datasets, probably because of the larger drop of MP for all the defenses when applied to NUSWIDE dataset.

% \begin{wrapfigure}{l}{0.60\textwidth}
% \vspace{-1em}
% \vspace{-1em}
% \end{wrapfigure}

% \begin{wrapfigure}{l}{0.60\textwidth}
% \vspace{-1em}
% \vspace{-1em}
% \end{wrapfigure}

\subsubsection{Additional Results on splitVFL and aggVFL Comparison} \label{subsec:appendix_splitVFL_comparison}
\begin{figure}[htbp]             
  \centering
    \includegraphics[width=0.99\linewidth]{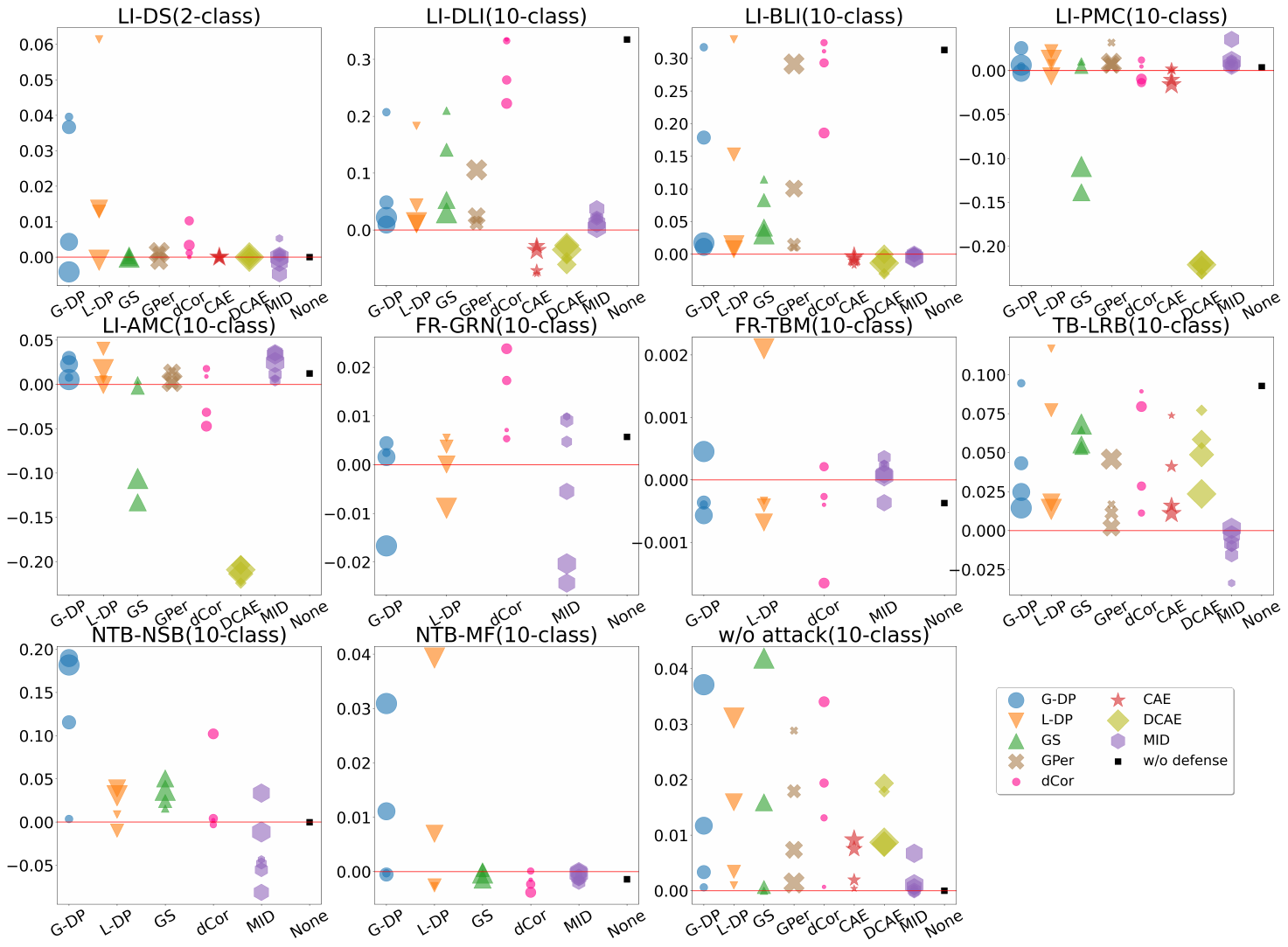}
  \caption{DCS gap for each attack-defense point [MNIST dataset, splitVFL/aggVFL, FedSGD]}
  \label{fig:mnist_splitVFL_DCS_gap}
  \vspace{-2em}
\end{figure}

\begin{figure}[!htb]
  \centering
    \includegraphics[width=0.6\linewidth]{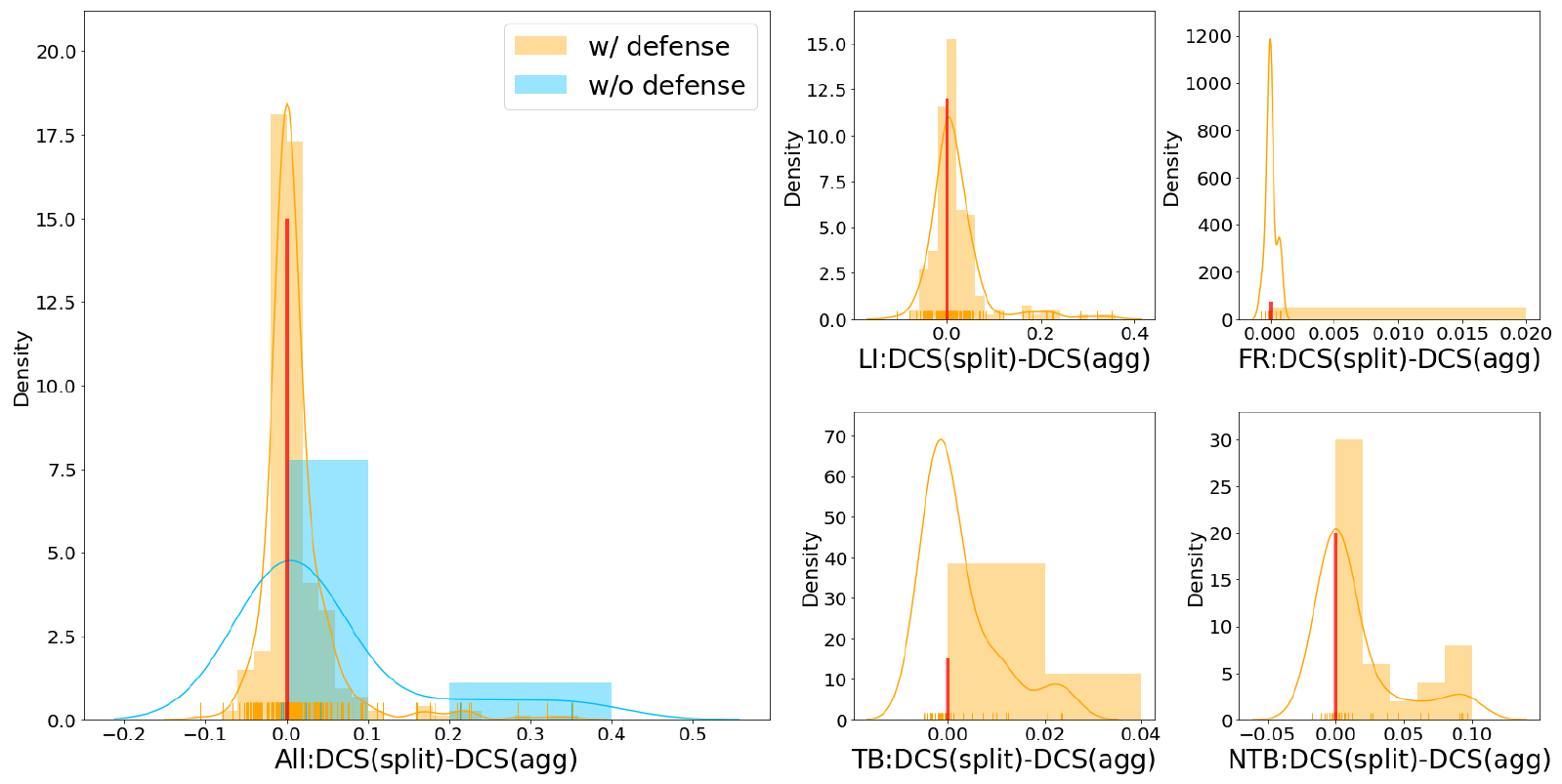}
  \caption{DCS gap Distribution, y-axis represents density [NUSWIDE dataset, splitVFL/aggVFL, FedSGD]}
  \label{fig:nuswide_split_hist}
\end{figure}

\begin{figure}[!htb]
  \centering
    \includegraphics[width=0.99\linewidth]{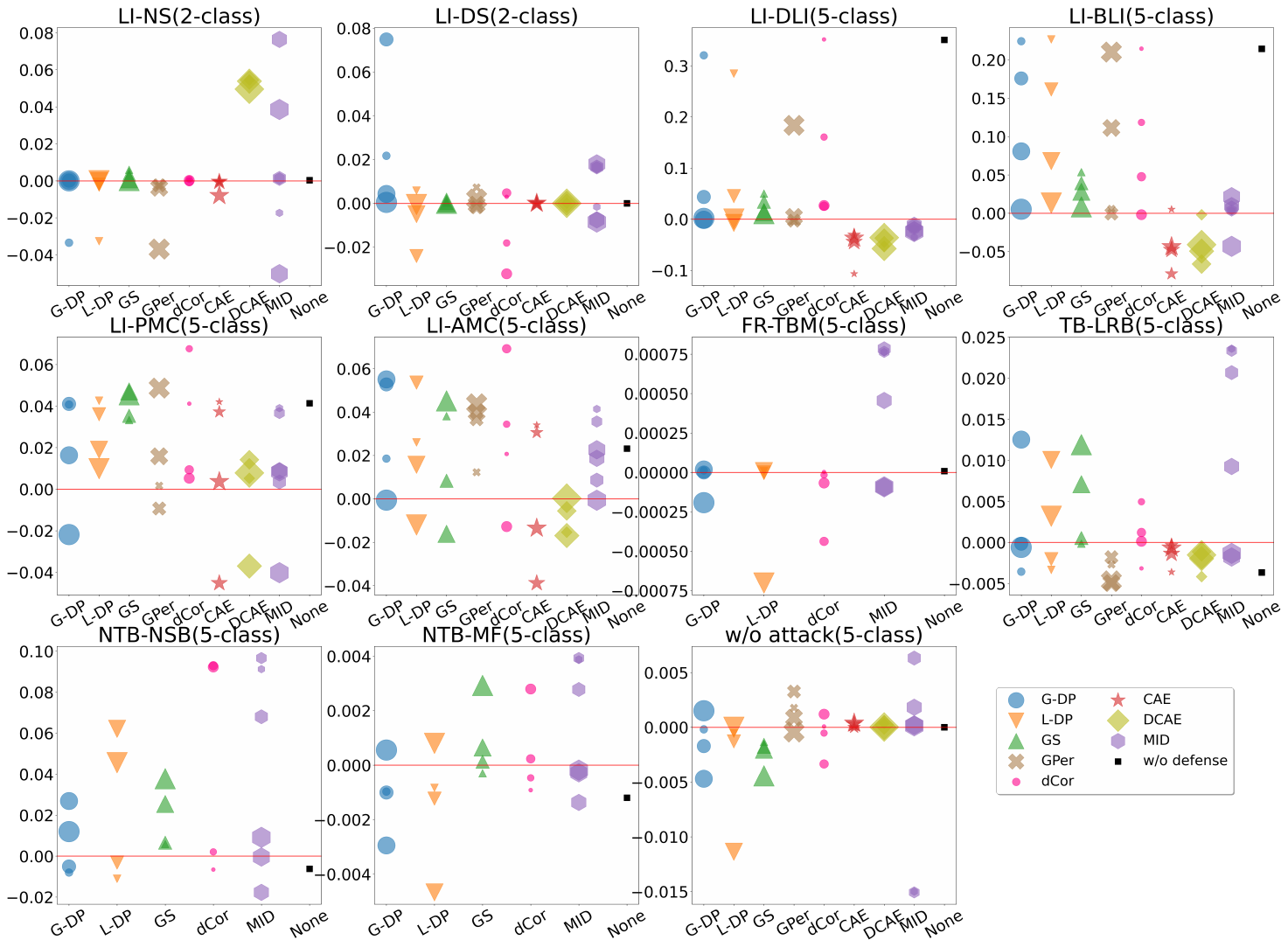}
  \caption{DCS gap for each attack-defense point [NUSWIDE dataset, splitVFL/aggVFL, FedSGD]}
  \label{fig:nuswide_split_DCS_gap}
\end{figure}

We further expand the discussion of the conclusions on the comparison between splitVFL and aggVFL as well as the comparison between FedBCD and FedSGD given in \cref{subsec:attack_defense_performance}
% \cref{subsec:splitVFL_comparison} 
here due to space limitations.

The conclusion that \textit{when no defense is applied, a splitVFL system is less vulnerable to attacks than aggVFL} is also evident by the results under NUSWIDE dataset, with all the black square points in \cref{fig:nuswide_split_DCS_gap} appearing above or close to the red horizontal line at a value of $0.0$ and the blue histograms appearing mostly at the right of the vertical line at a value of $0.0$ in \cref{fig:nuswide_split_hist}, indicating a positive DCS gap. This shows that a global trainable model can be beneficial to model robustness and safety. Specifically, the DCS gap is pronounced for attacks that directly exploit the gradients, i.e., DLI and BLI attacks. 

The conclusion that \textit{splitVFL has an overall positive effect on boosting defense performance against attacks} is also shown from \cref{fig:mnist_splitVFL_DCS_hist,fig:nuswide_split_hist} in which most DCS gaps for defenses are positive, especially for LI, FR and TB attacks. This implies that splitVFL architecture exhibits greater robustness against potential attacks.
However, not all defenses are enhanced in splitVFL setting. For example, in \cref{fig:mnist_splitVFL_DCS_gap}, MID results in minor negative DCS gaps in several attacks like GRN and NSB, so is GS in MC attacks.

The DCS ranking under splitVFL setting with FedSGD communication protocol using MNIST dataset is also included in \cref{tab:mnist_split_dcs_ranking}, which is quite similar to that in \cref{tab:mnist_dcs_ranking} under aggVFL setting with FedSGD communication protocol using MNIST dataset. This indicates the robustness of our DCS evaluation metrics as well as the inner consistency of defense abilities under different settings.

\subsubsection{Additional Results on FedBCD and FedSGD Comparison} \label{subsec:appendix_FedBCD_comparison}

\begin{figure}[!htb]
  \centering
    \includegraphics[width=0.99\linewidth]{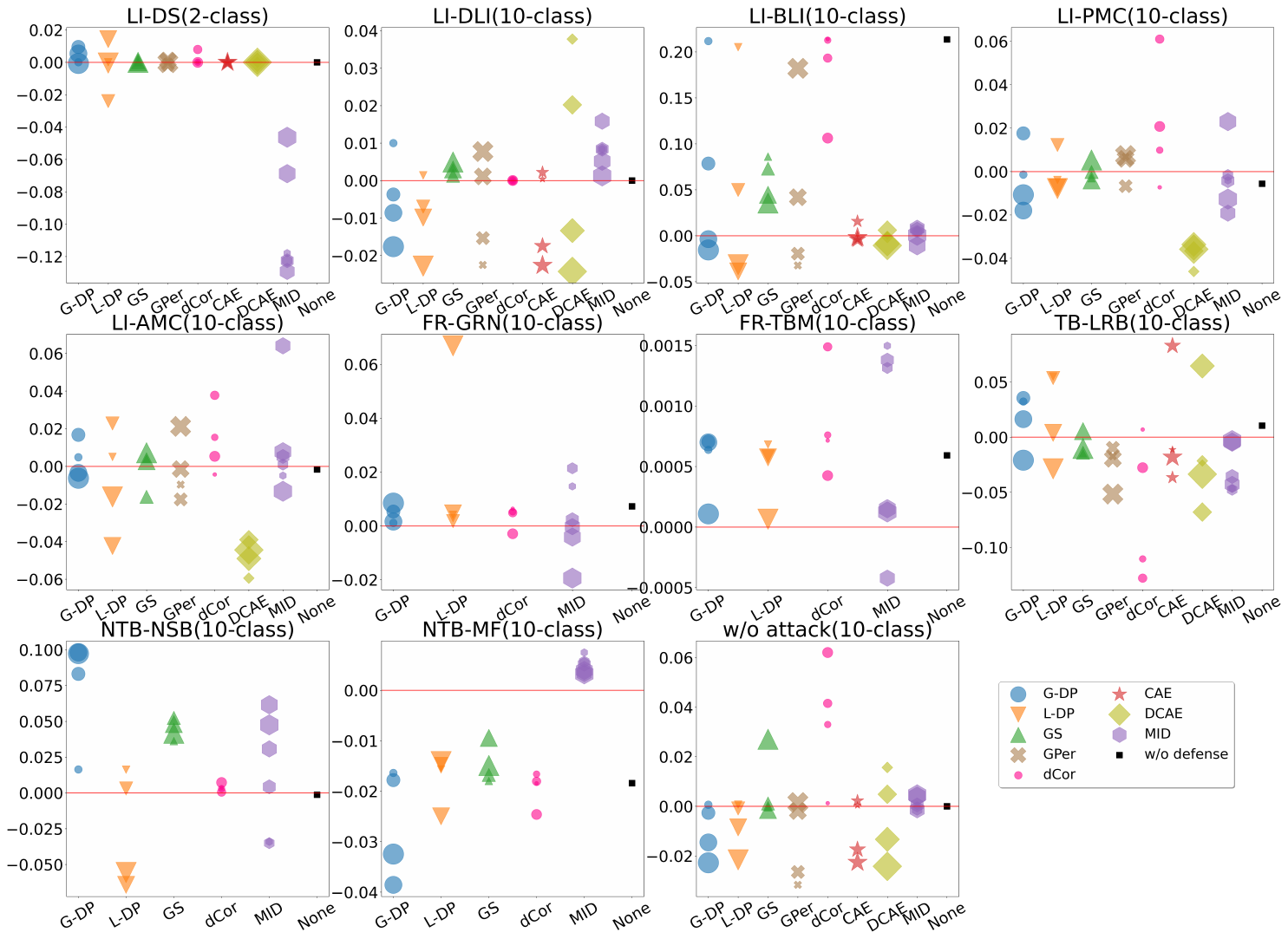}
  \caption{DCS gap for each attack-defense point [MNIST dataset, aggVFL, FedBCD/FedSGD]}
  \label{fig:mnist_FedBCD_DCS_gap}
\end{figure}

\begin{figure}[!htb]
  \centering
    \includegraphics[width=0.6\linewidth]{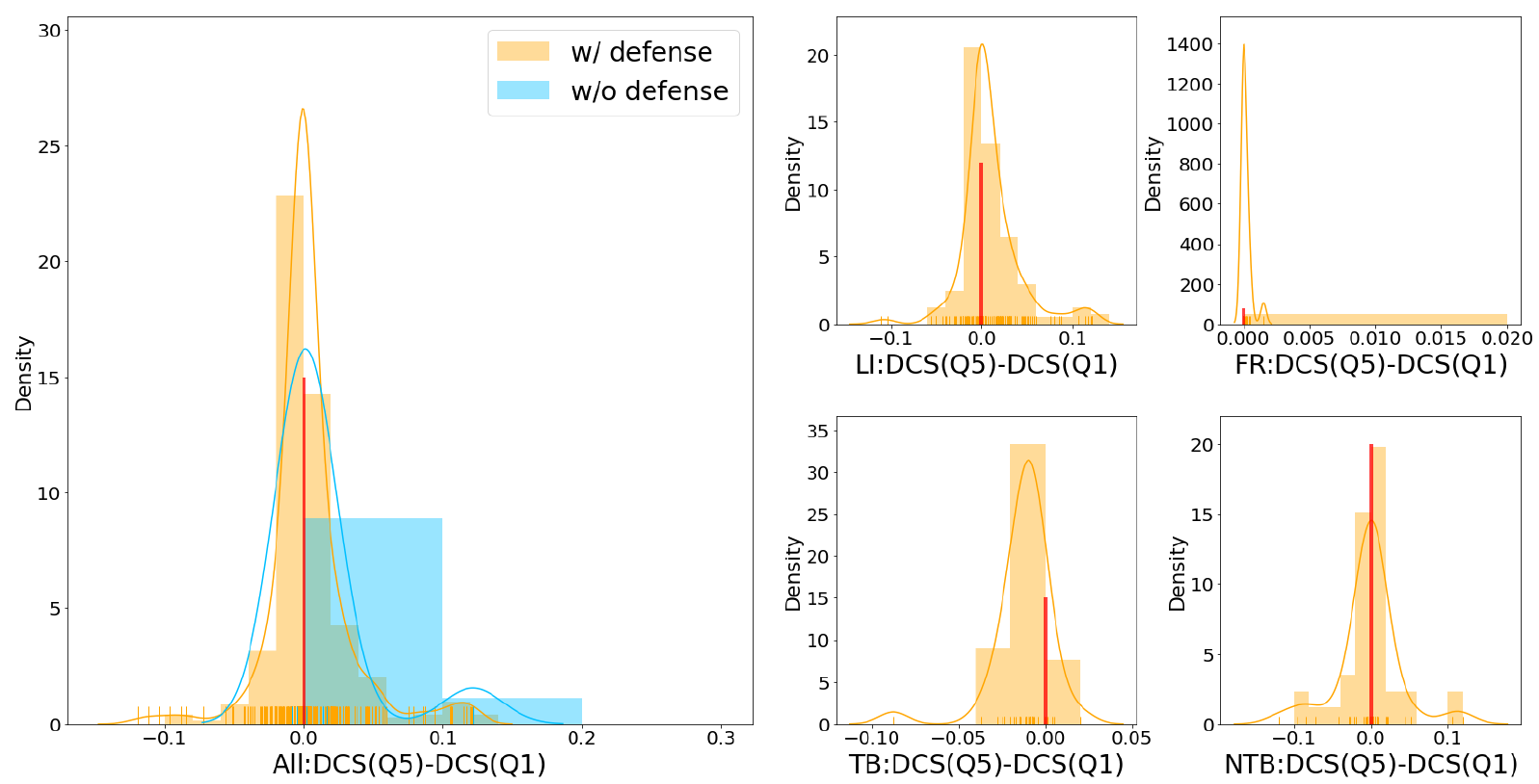}
  \caption{DCS gap Distribution, y-axis represents density [NUSWIDE dataset, aggVFL, FedBCD/FedSGD]}
  \label{fig:nuswide_FedBCD_hist}
\end{figure}

\begin{figure}[!htb]
  \centering
    \includegraphics[width=0.99\linewidth]{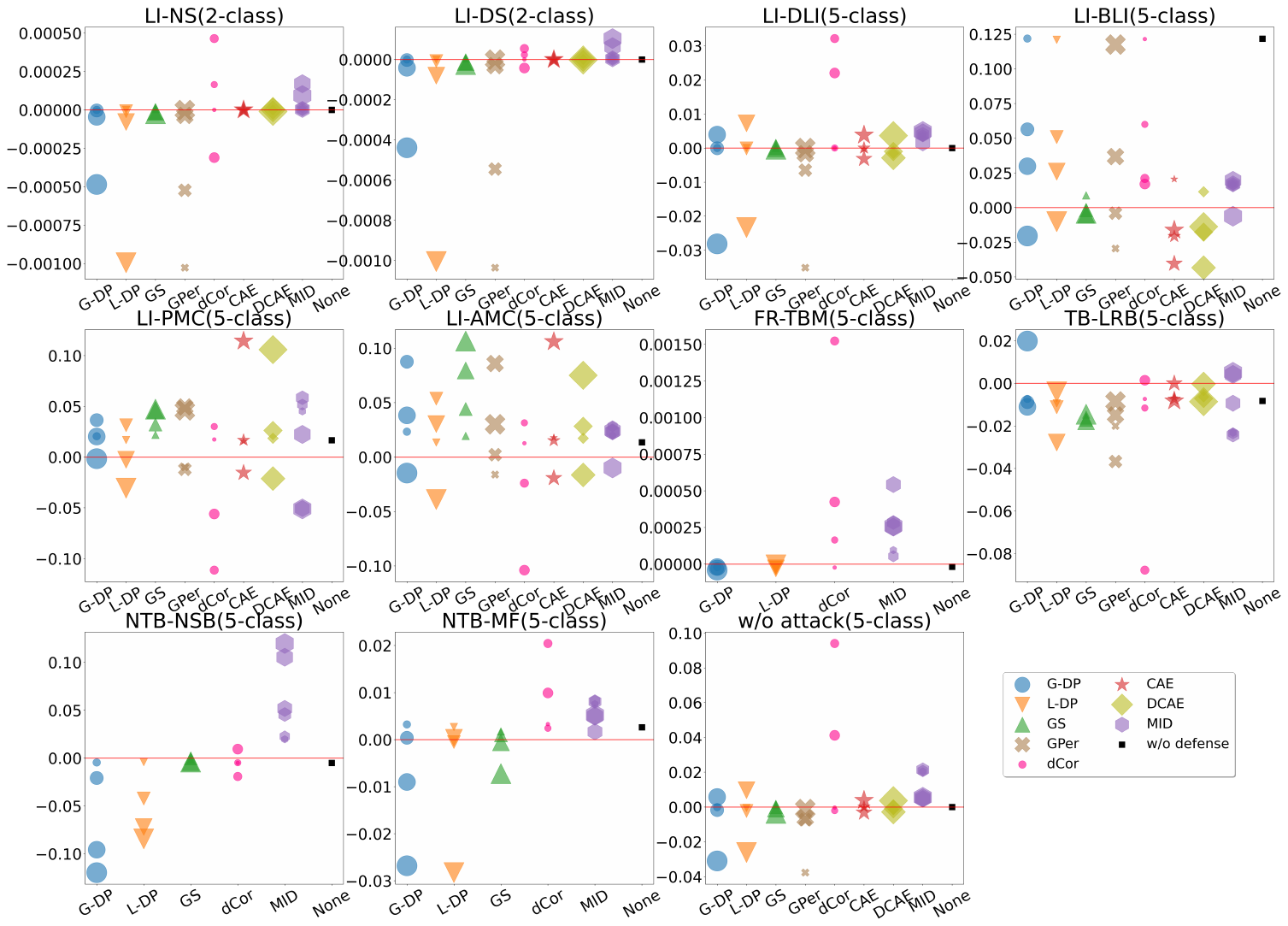}
  \caption{DCS gap for each attack-defense point [NUSWIDE dataset, aggVFL, FedBCD/FedSGD]}
  \label{fig:nuswide_FedBCD_DCS_gap}
\end{figure}

The conclusion given in
% \cref{subsec:FedBCD_comparison} 
\cref{subsec:attack_defense_performance}
that \textit{a system is less vulnerable to attacks under FedBCD setting when no defense method is applied} is also evident from \cref{fig:mnist_FedBCD_DCS_gap}, with all the black square points, except the one for MF attack, appear on or above the red horizontal line at a value of $0.0$. Also, the vulnerability of a system under FedBCD setting to attacks differs between different attacks. Specifically, as shown in \cref{fig:mnist_FedBCD_DCS_gap,fig:mnist_FedBCD_DCS_hist}, evaluating with MNIST dataset, VFL trained with FedBCD is much less vulnerable to BLI attack, which exploits batch-level gradient to recover labels. This is because gradients from earlier epochs of an un-trained model shared under FedSGD reveal more information about labels compared to FedBCD which only shares gradients every $Q>1$ iterations. Similar results can be seen from \cref{fig:nuswide_FedBCD_hist,fig:nuswide_FedBCD_DCS_gap} that compares the DCS between FedSGD and FedBCD under aggVFL setting with NUSWIDE dataset.

The DCS ranking using FedBCD communication protocol under aggVFL setting using MNIST dataset is also included in \cref{tab:mnist_fedbcd_dcs_ranking}, which is quite similar to that in \cref{tab:mnist_dcs_ranking} using FedSGD communication protocol under aggVFL setting using MNIST dataset. This also indicates the robustness of our DCS evaluation metrics as well as the inner consistency of defense abilities under different settings.

\subsubsection{Consistency of C-DCS ranking.}
\begin{figure}[!htb]
  \centering
    \includegraphics[width=0.79\linewidth]{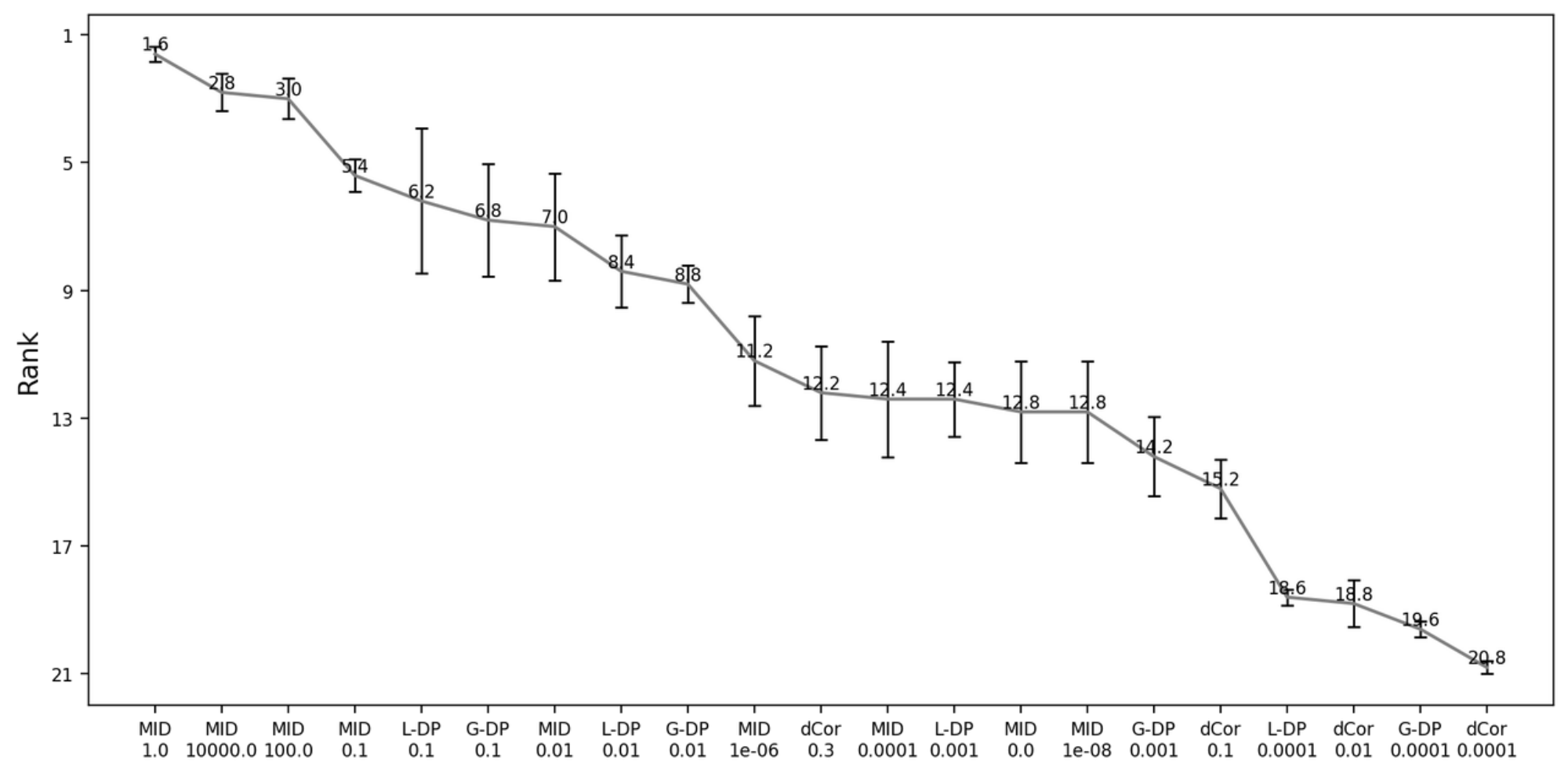}
  \caption{C-DCS ranking comparison across $3$ datasets, $2$ communication protocols and $2$ dataset partition strategies presented in \cref{tab:mnist_dcs_ranking,tab:cifar10_dcs_ranking,tab:nuswide_dcs_ranking,tab:mnist_split_dcs_ranking,tab:mnist_fedbcd_dcs_ranking}.}
  \label{fig:dcs_ranking_mean_std}
\end{figure}

We conduct a comparative analysis of the C-DCS rankings that are presented in \cref{tab:mnist_dcs_ranking,tab:cifar10_dcs_ranking,tab:nuswide_dcs_ranking,tab:mnist_split_dcs_ranking,tab:mnist_fedbcd_dcs_ranking} across multiple datasets, communication protocols, and model partition strategies. The mean and standard deviation of the $5$ rankings for each defense are calculated and visualized in \cref{fig:dcs_ranking_mean_std}. Remarkably, despite the diversity in datasets, communication methods, and model partitioning, the variations in the rankings remain at a low level. This suggests that the C-DCS rankings are generally consistent across various datasets, communication protocol and model partition settings, and that the relative performance of different defense methods is relatively stable under different datasets and settings.

\section{Social Impact of the work}
% \textbf{Societal Impacts.} 
Our work introduces an extensible and lightweight VFL platform for research which will for sure facilitate the research considering VFL. Moreover, our work encourages not only the development of stronger defense methods, but also the development of new attacks for practical VFL scenarios. These attacks can be used for either benign or malicious purpose. Developing stronger defense or exploring related regulations and laws are possible approaches for alleviating the potential negative impacts.

\end{document}